\documentclass{elsarticle}

\usepackage{lineno,hyperref}
\modulolinenumbers[5]

\journal{Artificial Intelligence}

\biboptions{authoryear}

\renewcommand{\cite}[1]{\citep{#1}}

\usepackage{graphicx}
\usepackage{amsmath}
\usepackage{xcolor}
\usepackage{url}
\usepackage{mathtools}

\usepackage{algpseudocode,algorithm,algorithmicx}

\usepackage{booktabs}
\usepackage{gensymb}
\usepackage{amssymb}
\usepackage{subcaption}

\usepackage{psgo}

\usepackage{chessfss}
\usepackage{tikz}
\usetikzlibrary{shapes,arrows,backgrounds,fit,positioning,calc,angles,quotes,arrows.meta}

\newcommand{\refalgorithm}[1]{Algorithm~\ref{#1}}
\newcommand{\refappendix}[1]{\ref{#1}}
\newcommand{\refequation}[1]{Equation~\ref{#1}}
\newcommand{\reffigure}[1]{\figurename~\ref{#1}}
\newcommand{\refsection}[1]{Section~\ref{#1}}
\newcommand{\refsubsection}[1]{Subsection~\ref{#1}}
\newcommand{\refsubsubsection}[1]{Subsubsection~\ref{#1}}
\newcommand{\reftable}[1]{Table~\ref{#1}}    

\newcommand{\transpose}[1]{#1^\top}
\newcommand{\argmax}{\operatornamewithlimits{argmax}}   
  
\newcommand{\rowvector}[1]{\begin{bmatrix} #1 \end{bmatrix}}

\makeatletter
\newcommand{\algorithmicbreak}{\textbf{break}}
\newcommand{\Break}{\State \algorithmicbreak}
\newcommand{\algorithmiccontinue}{\textbf{continue}}
\newcommand{\Continue}{\State \algorithmiccontinue}
\makeatother

\usepackage[T1]{fontenc}

\begin{document}

\begin{frontmatter}

\title{Spatial State-Action Features for General Games}


\author[umdacs]{Dennis J. N. J. Soemers\corref{mycorrespondingauthor}}
\cortext[mycorrespondingauthor]{Corresponding author}
\ead{dennis.soemers@maastrichtuniversity.nl}

\author[leiden,umdacs]{{\'E}ric Piette}

\author[flinders]{Matthew Stephenson}

\author[umdacs]{Cameron Browne}

\address[umdacs]{Maastricht University, Department of Advanced Computing Sciences, Paul-Henri Spaaklaan 1, 6229 EN, Maastricht, the Netherlands}
\address[leiden]{Leiden University, Leiden Institute of Advanced Computer Science, Snellius Building, Niels Bohrweg 1, 2333 CA, Leiden, the Netherlands}
\address[flinders]{Flinders University, College of Science and Engineering, Tonsley Campus, 1284 South Roach, Clovelly Park, 5042, Adelaide, Australia}

\begin{abstract}
In many board games and other abstract games, patterns have been used as features that can guide automated game-playing agents. Such patterns or features often represent particular configurations of pieces, empty positions, etc., which may be relevant for a game's strategies. Their use has been particularly prevalent in the game of Go, but also many other games used as benchmarks for AI research. 
In this paper, we formulate a design and efficient implementation of spatial state-action features for general games. These are patterns that can be trained to incentivise or disincentivise actions based on whether or not they match variables of the state in a local area around action variables. We provide extensive details on several design and implementation choices, with a primary focus on achieving a high degree of generality to support a wide variety of different games using different board geometries or other graphs. Secondly, we propose an efficient approach for evaluating active features for any given set of features. In this approach, we take inspiration from heuristics used in problems such as SAT to optimise the order in which parts of patterns are matched and prune unnecessary evaluations. 
This approach is defined for a highly general and abstract description of the problem---phrased as optimising the order in which propositions of formulas in disjunctive normal form are evaluated---and may therefore also be of interest to other types of problems than board games.
An empirical evaluation on $33$ distinct games in the Ludii general game system demonstrates the efficiency of this approach in comparison to a naive baseline, as well as a baseline based on prefix trees, and demonstrates that the additional efficiency significantly improves the playing strength of agents using the features to guide search.
\end{abstract}


\end{frontmatter}


\section{Introduction} \label{Sec:Introduction}

In research on machine learning for automated game-playing agents and Artificial intelligence (AI), the focus is often on achieving state-of-the-art or superhuman performance in one or a handful of games. In recent years, this is often based on combinations of Monte-Carlo Tree Search (MCTS) \cite{Kocsis_2006_Bandit,Coulom_2007_MCTS,Browne_2012_MCTS} and deep neural networks (DNNs) \cite{LeCun_2015_DeepLearning}. In principle this combination of techniques can be successfully applied \cite{Silver_2016_AlphaGo,Anthony_2017_ExIt,Silver_2017_AlphaGoZero,Lorentz_2017_Breakthrough,Silver_2018_AlphaZero,Tian_2019_ELF,Morandin_2019_SAI,Wu_2019_Accelerating,Cazenave_2020_Polygames,Cazenave_2022_Mobile} to a wide variety of games. However, in practice the high computational requirements make it infeasible to scale this up to large-scale studies that involve training agents for hundreds or thousands of distinct games \cite{Stephenson_2020_Database}, in addition to possibly many more variants of games generated automatically as possible reconstructions of games with incomplete rules \cite{Browne_2018_Modern,Browne_2019_DAL}.



Another common approach for improving the playing strength of search algorithms is to use higher-level features---as opposed to raw game state inputs---which may be expected to be capable of providing useful guidance to search algorithms without first being processed into useful representations through multiple layers of computation as in DNNs. Such features can be used by simpler, more efficient function approximators (e.g., linear functions, shallow neural networks, etc.) \cite{Enderton_1991_GolemGoProgram,Levinson_1991_Adaptive,Buro_1999_Features,vanderWerf_2003_Local,Coulom_2007_EloRatingsPatterns,Gelly_2007_Combining,Sturtevant_2007_Hearts,Huang_2014_MoHex,Lorentz_2017_Breakthrough}, or used for other purposes such as move ordering, directly playing winning moves, directly pruning losing moves, (de)prioritising other special types of moves, etc. \cite{Stoutamire_1991_Go,Muller_1995_PhD,Cazenave_1996_Automatic,Bouzy_2005_Associating,Bouzy_2005_Bayesian,Stern_2006_Bayesian,Gelly_2006_Modification,Araki_2007_Move,Raiko_2008_UCT,Skowronski_2009_Automated}---without otherwise using them as inputs for parameterised functions such as policy, value, or heuristic score functions. AlphaGo also used such high-level patterns for a computationally efficient rollout policy \cite{Silver_2016_AlphaGo}. While there are some exceptions, the majority of these approaches use local, \textit{spatial patterns} as high-level features. These are features that describe a specific arrangement of elements (e.g. pieces, off-board indicators, empty-position indicators, etc.) that must be present (or absent) in a specific area of a game board (e.g., a small $3$$\times$$3$ patch of a larger Go board) for the feature to be considered active in that area. The general concept of such spatial features appears to be useful in various different games, but in each of the publications listed above they are designed, implemented, and evaluated for a single, specific game (most commonly Go)---often leveraging game-specific domain knowledge for efficient designs and implementations. In addition to possibly improving the playing strength in many games, such spatial features or patterns are attractive because they may contribute towards \textit{human-like} game-playing \cite{Levinson_1991_Adaptive,Mandziuk_2011_Plausible}, may facilitate generalisation and transfer thanks to the game-independent spatial semantics, and may be easy to visualise, interpret, and explain \cite{Browne_2019_Strategic}.

\begin{figure}
\centering

\begin{subfigure}{.3\textwidth}
\centering
\resizebox{\linewidth}{!}{
\begin{tikzpicture} [hexa/.style= {shape=regular polygon,regular polygon sides=6,minimum size=1cm,very thin,blue!50!white,draw,inner sep=0,anchor=south,fill=blue!16!white}]
\foreach \j in {0,...,2}{%
     \ifodd\j 
         \foreach \i in {0,...,2}{
            \node[hexa] (h\j;\i) at ({\j/2+\j/4},{(\i+1/2)*sin(60)}) {\ifthenelse{\i=1}{\textcolor{green!50!black}{\textbf{\textasteriskcentered}}}{}};
          }
    \else
         \foreach \i in {1,...,2}{
            \node[hexa] (h\j;\i) at ({\j/2+\j/4},{\i*sin(60)}) {};
          }
    \fi}  
\node[circle,draw=black,fill=white,inner sep=0pt,minimum size=0.6cm] (a) at ({1/2+1/4},{3*sin(60)}){};
\node[circle,draw=black,fill=white,inner sep=0pt,minimum size=0.6cm] (b) at ({0},{1.5*sin(60)}){};
\node[circle,draw=black,fill=black!75,inner sep=0pt,minimum size=0.6cm] (c) at ({0},{2.5*sin(60)}){};
\end{tikzpicture}
}
\end{subfigure}
\begin{subfigure}{.3\textwidth}
\centering
\resizebox{\linewidth}{!}{
\begin{tikzpicture}

\fill[blue!16!white] (0.1,0.1) rectangle (3.9,5.9);
\draw[step=1cm,blue!50!white,very thin] (0.1,0.1) grid (3.9,5.9);

\node[circle,draw=black,fill=black!75,inner sep=0pt,minimum size=0.6cm] (a) at ({2},{4}){};
\node[circle,draw=black,fill=black!75,inner sep=0pt,minimum size=0.6cm] (b) at ({2},{5}){};
\node[circle,draw=black,fill=black!75,inner sep=0pt,minimum size=0.6cm] (c) at ({2},{2}){};
\node[circle,draw=black,fill=black!75,inner sep=0pt,minimum size=0.6cm] (d) at ({2},{1}){};

\node[draw=black,fill=blue!4!white,inner sep=0pt,minimum size=0.5cm] (e) at ({2},{3}) {\textcolor{green!50!black}{\textbf{\textasteriskcentered}}};

\end{tikzpicture}
}
\end{subfigure}
\begin{subfigure}{.3\textwidth}
\centering
\resizebox{\linewidth}{!}{
\begin{tikzpicture}

\fill[blue!8!white] (0.1,0.1) rectangle (2.9,2.9);
\draw[step=1cm,blue!50!white,very thin] (0.1,0.1) grid (2.9,2.9);

\node at ({1.5},{1.5}) {\BlackPawnOnWhite};
\node at ({2.5},{1.5}) {\WhitePawnOnWhite};
\node at ({0.5},{0.5}) {\WhitePawnOnWhite};

\draw[very thick,->,green!50!black] (0.5,0.5) -- (0.5,1.5);

\end{tikzpicture}
}
\end{subfigure}
\caption{Three examples of local spatial features that may be useful in various games. The leftmost feature matches actions that either complete or break a ``bridge'' of white pieces, depending on which player is the player to move. The middle feature matches actions that either complete or break a line of five black pieces. The rightmost feature matches actions that move the bottom-left white pawn in such a way that the black pawn becomes squashed between two white pawns. }
\label{Fig:ExampleFeatures}
\end{figure}
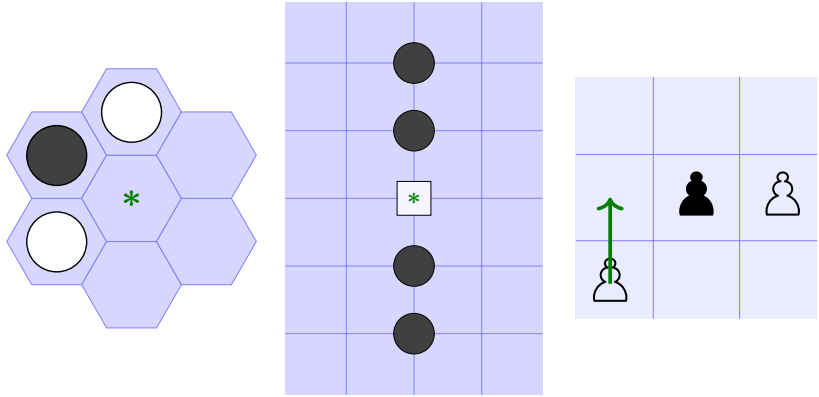

In this paper, we focus on the design and efficient implementation of spatial features for \textit{general games} implemented in the Ludii general game system \cite{Browne_2020_Practical,Piette_2020_Ludii}. \reffigure{Fig:ExampleFeatures} depicts several examples of what such features may look like. The aim is not to achieve superhuman levels of performance as may be expected from DNN-based approaches, but rather to learn at least meaningful strategies that can substantially improve the playing strength of standard (unbiased) MCTS agents across many games, with a level of efficiency such that scaling up to many hundreds of games is feasible. To this end, we have two core contributions:
\begin{enumerate}
    \item We provide a significant extension of previous work in which we proposed a design of such spatial features \cite{Browne_2019_Strategic}.
    \item We explore how to implement them in a way such that the activity of features can be efficiently evaluated and used in, for instance, rollouts as used by MCTS. The algorithm we propose for this is defined for a highly general and abstract formulation of the problem, where we simply aim to optimise the order in which to evaluate propositions that appear in formulas in disjunctive normal form. This may also make it applicable to other problem domains than games (which we discuss in \refsection{Sec:Discussion}).
\end{enumerate}
Previous work with earlier, less extensive and optimised designs of such features has already demonstrated the ability to train effective policies for a wide variety of games using such features \cite{Soemers_2019_Biasing,Soemers_2020_Manipulating}.

\refsection{Sec:Background} provides background information on Ludii and general game playing, as well as a brief summary of MCTS. Related work is discussed in \refsection{Sec:RelatedWork}. We formalise the design and format of our spatial features in \refsection{Sec:Formalisation}, which also includes discussions of how relevant (spatial) aspects of states and actions are represented in the Ludii general game system. In \refsection{Sec:MainSection} we propose our approach for efficiently evaluating which spatial features out of a feature set are active for any given state-action pair in the Ludii general game system. \refsection{Sec:Experiments} describes and discusses the setup and results of our experiments. A discussion of potential applications beyond games is provided by \refsection{Sec:Discussion}. Finally, \refsection{Sec:Conclusion} concludes the paper and explores ideas for future work.

\section{Background} \label{Sec:Background}

In this section, we provide background information on some of the basic concepts that we build on in the remainder of the paper. We start with a discussion of the Ludii general game system \cite{Piette_2020_Ludii,Browne_2020_Practical}, and how it relates to the research field of General Game Playing. We use the vast library of games available in Ludii for experiments, and also rely on some domain knowledge of its game, state, and move representations \cite{Piette_2021_LudiiGameLogicGuide} to implement spatial features. Afterwards, we briefly describe Monte-Carlo Tree Search.

\subsection{Ludii and General Game Playing}

General Game Playing (GGP) is a field of research in which the goal is to develop agents that can play \textit{general games} without human intervention \cite{Pitrat_1968_GGP,Swiechowski_2015_RecentAdvancesGGP}, i.e. agents that can play any arbitrary game without requiring any game-specific domain knowledge. Such agents typically expect any game they are tasked with playing to have been implemented in a specific Game Description Language (GDL)---like the Stanford GDL \cite{Love_2008_GDL} or the game description language of Ludii \cite{Piette_2020_Ludii,Browne_2020_Practical}---such that the agents can interact with any game through a single, common API.

Game descriptions in the Stanford GDL describe the rules of games in low-level logic, which means that many commonly-used high-level game concepts (such as \textit{square boards}, \textit{hexagonal boards}, \textit{lines of pieces}, \textit{slide moves}, \textit{step moves}, etc.) need to be expressed from scratch in low-level logic in any new game description that requires them. Game descriptions in Ludii's GDL are not written in low-level logic, but instead use a significantly larger library of \textit{ludemes} (i.e., keywords), most of which summarise such commonly-used, high-level concepts in a single word or phrase. This design is intended to make it significantly easier---for humans, but possibly also evolutionary algorithms \cite{Browne_2009_PhD}---to write, read, and understand new game descriptions---in particular for games that are similar to ``real-world'' board games and primarily use rules and equipment that are already implemented as first-class citizen, and encapsulated by appropriately-named keywords, in Ludii's GDL. Ludii's object-oriented game and state representations \cite{Piette_2020_Ludii} similarly make commonly-used concepts such as \textit{game boards}, \textit{cells}, \textit{vertices}, \textit{edges}, \textit{orthogonal} and \textit{diagonal connections}, \textit{neighbours}, etc. readily available for any game to any agent, which allows for reliable implementations of spatial features, which rely on knowledge of such concepts.

Note that a significant amount of GGP research has been inspired by the International General Game Playing (IGGP) competition \cite{Genesereth_2013_GGP}. In this competition, agents only gain access to the game descriptions of any games they play when the competition starts, leaving relatively little time for any offline training. In this paper, we consider perhaps a less strict idea of GGP. In principle, we do not mind ``telling'' an agent in advance which games it will be expected to play, and making use of more offline training time. However, the sheer scale of working with many hundreds or thousands of games \cite{Browne_2018_Modern,Stephenson_2020_Database} does put practical limitations on how much training time and hardware can be used for any individual game, and makes it infeasible to program game-specific agents. This means that we cannot, for instance, handcraft strong heuristic evaluation functions or features for every game---not necessarily because we do not know which games we wish to play, but because we simply wish to play too many different games for the programming effort to be feasible. On the other hand, we can for instance leverage the knowledge that the vast majority of games we are interested in are ``real-world'' board games. Most of these involve spatial semantics; a game board with defined positions and connectivity relations between those positions, pieces placed on such positions, movement rules or victory conditions based on the connectivity structure, etc. This knowledge informs our choice to focus on \textit{spatial} features.

\subsection{Monte-Carlo Tree Search}

\begin{figure}
\centering
\includegraphics[width=\linewidth]{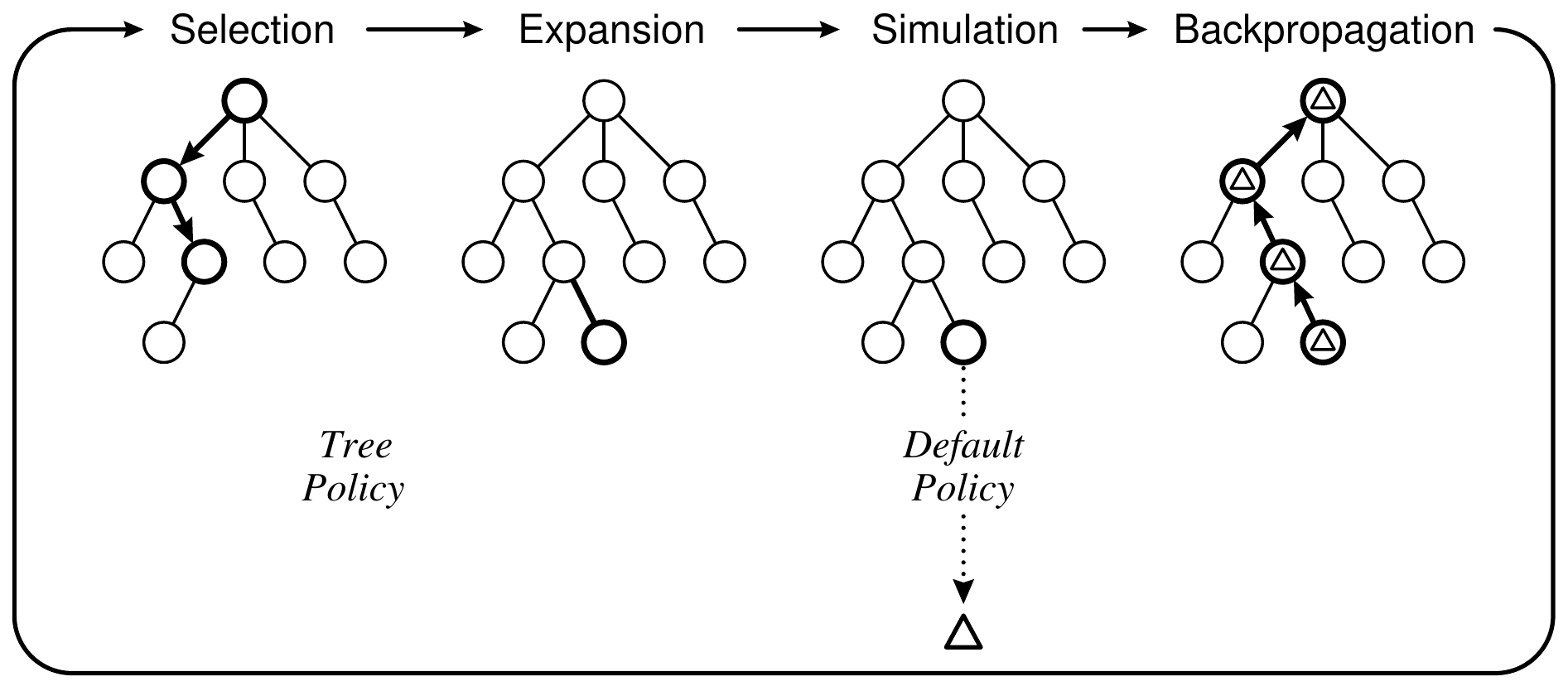}
\caption{The four phases of an MCTS iteration. Source of image: \cite{Browne_2012_MCTS}.}
\label{Fig:MCTSSteps}
\end{figure}

Monte-Carlo Tree Search (MCTS) \cite{Kocsis_2006_Bandit,Coulom_2007_MCTS,Browne_2012_MCTS} is a commonly-used tree search algorithm in GGP \cite{Finnsson_2010_Learning,Swiechowski_2015_RecentAdvancesGGP}, as well as several recent state-of-the-art game-specific agents \cite{Silver_2018_AlphaZero,Cazenave_2020_Polygames}. It gradually builds up a search tree in an asymmetric manner, such that it focuses a greater amount of search effort on parts of the search tree that appear to be promising so far---based on intermediate estimates during the search---and less effort on less promising parts of the tree. This works by iterating through a sequence of four phases, referred to as \textit{Selection}, \textit{Expansion}, \textit{Simulation} (or \textit{Play-out}), and \textit{Backpropagation}, for as many iterations as allowed by some search budget. This is depicted in \reffigure{Fig:MCTSSteps}.
\begin{enumerate}
    \item \textit{Selection}: In the Selection phase, the algorithm traverses from the root of the tree to a part of the search tree that warrants more search effort. This is typically based on a policy that provides a balance between \textit{exploitation} of parts of the search tree that appear promising so far, and \textit{exploration} of parts of the search tree that have received relatively little attention, such as the UCB1 policy \cite{Auer_2002_Finite}.
    \item \textit{Expansion}: In the Expansion phase, the algorithm can choose to grow the search tree. Typically this is done by adding a single new node to represent a successor---which was previously not represented by any node---of the final node that the Selection phase ended up in, but it is also possible to add multiple different new nodes at once, or add no new nodes (for instance when the Selection phase already traversed all the way to a node that represents a terminal game state without successors).
    \item \textit{Simulation}: The goal of the simulation phase is to obtain an estimate of the value of a node (or trajectory of actions) chosen by the Selection (plus Expansion) phase. The most straightforward approach for this is to run one or more random play-outs, where actions are selected uniformly at random, until a terminal game state or other limit on the play-out duration is reached, using the expanded node as starting point. This typically results in a very noisy value estimate, but it is easy to implement, efficient to run, and does not require any domain knowledge or offline learning. Alternatives include running non-uniformly random play-outs based on online \cite{Finnsson_2010_Learning} or offline \cite{Silver_2016_AlphaGo} learning, or replacing play-outs altogether by a trained value function estimator \cite{Silver_2017_AlphaGoZero}.
    \item \textit{Backpropagation}: The Backpropagation phase propagates the obtained value estimate back through the path of the tree that was traversed in this iteration. Typically, this results in the value of a node being estimated as the average of all value estimates that have been backpropagated through that node, over all iterations that traversed that node.
\end{enumerate}

\section{Related Work} \label{Sec:RelatedWork}

Previous work in automated discovery of useful features or heuristics for GGP \cite{Kuhlmann_2006_Automatic,Clune_2007_Heuristic,Schiffel_2007_Fluxplayer,Finnsson_2010_Learning,Michulke_2011_Heuristic,Kirci_2011_GGP,Waledzik_2011_Multigame,Michulke_2012_Distance,Michulke_2013_Admissible,Waledzik_2014_Automatically} primarily focuses on approaches that are specific to games described in the Stanford GDL \cite{Love_2008_GDL}, with its logic-based state representation. While some of these approaches may end up learning features that could be interpreted as being similar to the spatial features we consider, actually recognising and interpreting them as such first requires a human to manually analyse and interpret how the abstract, logic-based expressions in a specific game's description file relate to human-understandable concepts. Furthermore, in Ludii, many such approaches are unnecessary because spatial patterns and features can be directly implemented based on the data that is explicitly available in its game, state, and move representations \cite{Piette_2021_LudiiGameLogicGuide}. Higher-level, game-wide features have also been proposed for Ludii \cite{Piette_2021_Concepts}, but in this paper we focus on features that can provide information about individual game states (or individual actions within game states).

Outside of GGP, spatial patterns have frequently been used in game-specific programs for games such as Chess \cite{Bratko_1978_Pattern,Beal_1980_Construction,Levinson_1991_Adaptive}, Othello \cite{Buro_1999_Features}, Breakthrough \cite{Skowronski_2009_Automated,Saffidine_2012_Solving,Lorentz_2017_Breakthrough}, Hex \cite{Huang_2014_MoHex}, and perhaps most commonly Go \cite{Muller_1995_PhD,Cazenave_1996_Automatic,Muller_2002_Go,vanderWerf_2003_Local,Bouzy_2005_Associating,Bouzy_2005_Bayesian,Stern_2006_Bayesian,Gelly_2006_Modification,Coulom_2007_EloRatingsPatterns,Silver_2007_LocalShape,Gelly_2007_Combining,Araki_2007_Move}. In all of this related work, patterns are formalised and implemented for a specific game, which means that the same formalisations are not directly applicable to general games. For example, in the game of Go, there are only two players (black and white), each of which only has a single piece type (a stone), and stones are only ever placed on the intersections of a square tiling of square cells. Therefore, in related work on patterns in Go, it is customary to only have patterns that test for particular arrangements of empty, white, and black positions---sometimes extended with board edge detectors and wildcards (positions for which it does not matter whether or not, and by what, they are occupied). For applicability to general games, our formalisation of patterns requires support for different board structures, more diverse sets of piece types and numbers of players, etc.

\begin{figure}
\centering
\includegraphics[width=.45\linewidth]{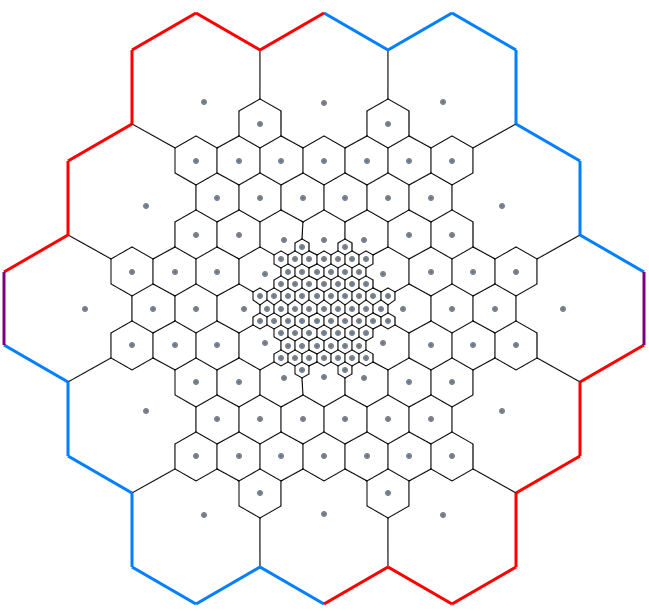}
\caption{The game of Fractal in Ludii. The dots mark playable sites; these are cells of a variety of shapes.}
\label{Fig:Fractal}
\end{figure}

In the majority of research based on deep learning for board games in more recent years, \textit{convolutional neural networks} (CNNs) \cite{LeCun_1989_Backpropagation} are used \cite{Silver_2016_AlphaGo,Silver_2017_AlphaGoZero,Anthony_2017_ExIt,Lorentz_2017_Breakthrough,Silver_2018_AlphaZero,Tian_2019_ELF,Morandin_2019_SAI,Wu_2019_Accelerating,CohenSolal_2020_Learning,Cazenave_2020_Polygames,CohenSolal_2021_Minimax,Soemers_2021_Transfer,Soemers_2022_DeepLearning}. These typically operate on raw, low-level state inputs, as opposed to the higher-level features considered in this paper. However, due to the way in which CNNs implement translation invariance and weight sharing by ``sliding'' learned filters over the spatial dimensions of a state input, such learned filters may intuitively be understood as encoding ``fuzzy'' versions of spatial patterns. Note that most of this related work only uses convolutional layers at the ``start'' of a neural network, but follows these up with at least one fully-connected layer. This means that there is no translation invariance or other use of spatial semantics in the action-based policy outputs. This is in contrast to the spatial state-action features considered in this paper, which specifically look at a local area around an action in any given state. Fully-convolutional architectures \cite{Shelhamer_2017_Fully}, which have been demonstrated to facilitate transfer learning between games with different board sizes \cite{Soemers_2021_Transfer}, bear a closer resemblance in this respect. It should also be noted that CNNs normally work best on regular tilings, such as grids of pixels or square cells, or regular tilings containing only hexagonal cells, etc. In this paper, we aim to formalise features independent of any single particular regular tiling, and even aim for them to be applicable to game boards that do not have a regular tiling (see \reffigure{Fig:Fractal}). Such levels of generalisation can be present in more general (graph-based) architectures than CNNs \cite{Bronstein_2021_Geometric}, which have for instance been applied to the game of Risk \cite{Carr_2020_Risk}.

\section{Formalisation of Spatial State-Action Features} \label{Sec:Formalisation}

In this section, we formalise our design of spatial state-action features.
These are binary features $\phi(s, a)$ of game states $s$ and actions\footnote{Throughout this paper, we follow standard reinforcement learning terminology \cite{Sutton_2018_RL}, referring to the decisions of agents (or players) as actions. Within the Ludii general game system, there is a distinction between actions and moves, with actions describing primitive modifications of game states, and moves---formed by sequences of one or more actions---describing decisions made by players. In this paper, there is no need for any such distinction.} $a$. The basic idea is that a feature tests for one or several conditions in the neighbourhood around the positions involved in an action, and is either active ($\phi(s, a) = 1$) or inactive ($\phi(s, a) = 0$) if the conditions are satisfied or not satisfied, respectively, for $s$ and $a$. Simple examples of conditions include testing whether there is a friendly piece, or enemy piece, or empty position, etc., in some position relative to (e.g., adjacent to) the position that is the destination of an action $a$ in a state $s$. Before describing the design of these features in further detail, we discuss some preliminaries concerning the game, state, and action representations in Ludii \cite{Piette_2021_LudiiGameLogicGuide}.

\subsection{Ludii Game, State, and Action Representations}

\begin{figure}
\centering
\begin{subfigure}{.24\textwidth}
  \centering
  \includegraphics[width=\linewidth]{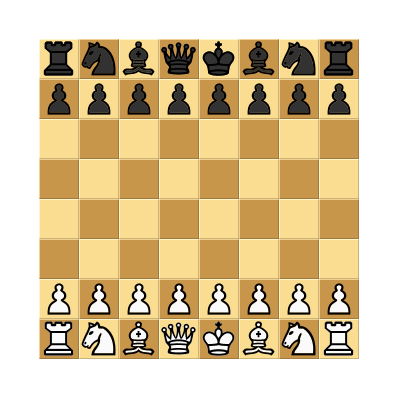}
  \caption{Chess.}
  \label{Fig:Chess}
\end{subfigure}
\begin{subfigure}{.24\textwidth}
  \centering
  \includegraphics[width=\linewidth]{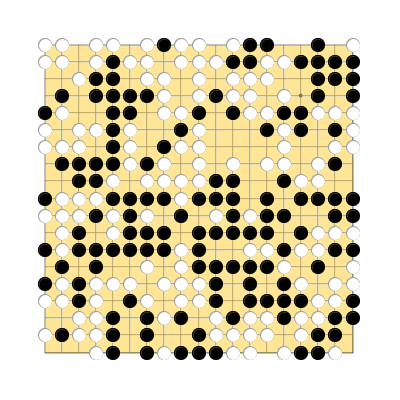}
  \caption{Go.}
  \label{Fig:Go}
\end{subfigure}
\begin{subfigure}{.24\textwidth}
  \centering
  \includegraphics[width=\linewidth]{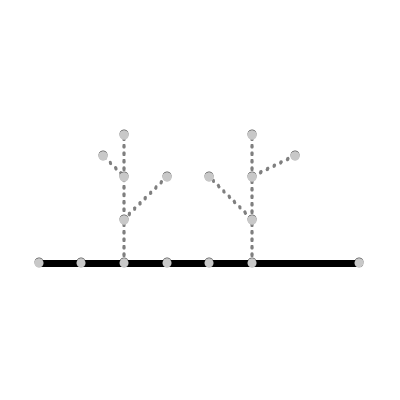}
  \caption{Hackenbush.}
  \label{Fig:Hackenbush}
\end{subfigure}
\begin{subfigure}{.24\textwidth}
  \centering
  \includegraphics[width=\linewidth]{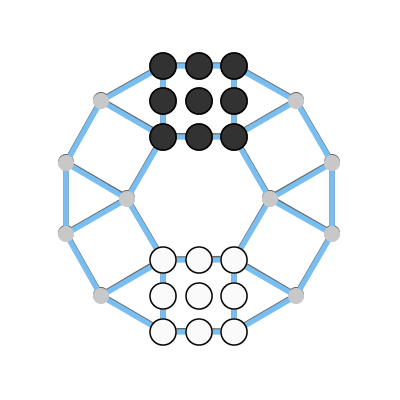}
  \caption{Triple Tangle.}
  \label{Fig:TripleTangle}
\end{subfigure}
\caption{Four different games in Ludii that use different graph element types in their rules. Chess only involves the cells of its board. Go only involves the vertices of its board. Hackenbush only involves the edges of its graph. In Triple Tangle, pieces are placed on cells, vertices, and edges.}
\label{Fig:SiteTypeExampleGames}
\end{figure}

Every game's representation in Ludii includes one or more \textit{containers}, each of which defines a ``playable area'' as a graph, which has vertices, edges, and---in many cases---cells (or faces). Typically there is one primary container (often simply representing a game's board), and sometimes one or more additional containers to represent supplementary areas. For instance, the game of Shogi as modelled in Ludii has additional containers to represent ``players' hands'', which hold captured pieces. Among all games (over 1,000) implemented in Ludii at the time of this writing, there is not a single game with a meaningful connectivity structure or spatial semantics within any container other than a game's primary board. Hence, in this paper, we only consider using spatial features within any game's main board. While the graphs used for many games define vertices, edges, and cells, the vast majority of games only use one of those three types in their rules. For example, chess only uses cells, Go only uses vertices, and the implementation of Hackenbush in Ludii only uses edges (see \reffigure{Fig:SiteTypeExampleGames}). A few games use a combination of multiple types, such as Triple Tangle, which uses all three types.

\subsubsection{Ludii Board Geometry} \label{Subsubsec:LudiiBoardGeometry}
\begin{figure}
\centering

\begin{subfigure}{.32\textwidth}
\centering
\resizebox{\linewidth}{!}{
\begin{tikzpicture}

\fill[blue!8!white] (0.1,0.1) rectangle (4.9,4.9);
\draw[step=1cm,blue!25!white,very thin] (0.1,0.1) grid (4.9,4.9);

\foreach \j in {0,...,4}{%
    \foreach \i in {1,...,4}{%
        \draw[<->,blue,thick] (\j+0.5,\i+0.5) -- (\j+0.5,\i-0.5);
    }
}

\foreach \j in {0,...,3}{%
    \foreach \i in {0,...,4}{%
        \draw[<->,blue,thick] (\j+0.5,\i+0.5) -- (\j+1.5,\i+0.5);
    }
}
\end{tikzpicture}
}
\caption{Orthogonal connections between cells.}
\label{Fig:SquareOrthogonalConnections}
\end{subfigure}
\begin{subfigure}{.32\textwidth}
\centering
\resizebox{\linewidth}{!}{
\begin{tikzpicture}

\fill[blue!8!white] (0.1,0.1) rectangle (4.9,4.9);
\draw[step=1cm,blue!25!white,very thin] (0.1,0.1) grid (4.9,4.9);

\foreach \j in {1,...,4}{%
    \foreach \i in {1,...,4}{%
        \draw[<->,blue,thick,dashed] (\j+0.5,\i+0.5) -- (\j-0.5,\i-0.5);
    }
}

\foreach \j in {0,...,3}{%
    \foreach \i in {1,...,4}{%
        \draw[<->,blue,thick,dashed] (\j+0.5,\i+0.5) -- (\j+1.5,\i-0.5);
    }
}

\end{tikzpicture}
}
\caption{Diagonal connections between cells.}
\label{Fig:SquareDiagonalConnections}
\end{subfigure}
\begin{subfigure}{.32\textwidth}
\centering
\resizebox{\linewidth}{!}{
\begin{tikzpicture}

\fill[blue!8!white] (0,0) rectangle (5,5);
\draw[step=1cm,blue!25!white,very thin] (0,0) grid (5,5);

\foreach \j in {0,...,5}{%
    \foreach \i in {1,...,5}{%
        \draw[<->,blue,thick] (\j,\i) -- (\j,\i-1);
    }
}

\foreach \j in {0,...,4}{%
    \foreach \i in {0,...,5}{%
        \draw[<->,blue,thick] (\j,\i) -- (\j+1,\i);
    }
}

\foreach \j in {1,...,5}{%
    \foreach \i in {1,...,5}{%
        \draw[<->,blue,thick,dashed] (\j,\i) -- (\j-1,\i-1);
    }
}

\foreach \j in {0,...,4}{%
    \foreach \i in {1,...,5}{%
        \draw[<->,blue,thick,dashed] (\j,\i) -- (\j+1,\i-1);
    }
}
\end{tikzpicture}
}
\caption{Orthogonal and diagonal connections between vertices.}
\label{Fig:SquareVertexConnections}
\end{subfigure}
\caption{Various types of connections on square grids.}
\label{Fig:SquareConnections}
\end{figure}
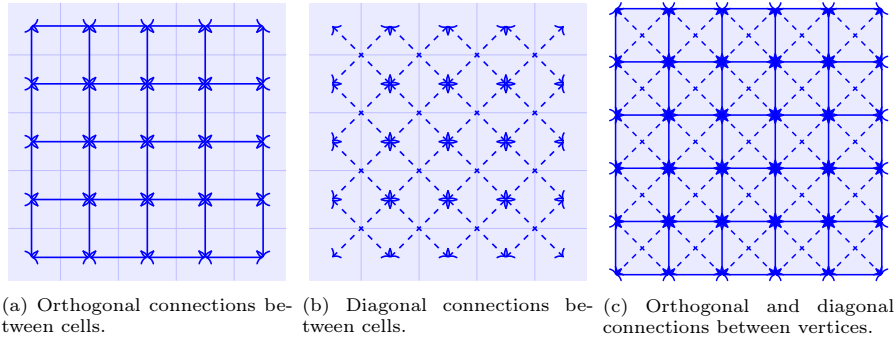

Given any graph as described above, the \textit{connectivity} relations between certain elements of the graphs are typically crucial for the game's rules and its strategies. For example, movement rules of pieces in Chess are typically defined in terms of the connectivity relations between cells of the board, win conditions in line-completion games such as Tic-Tac-Toe or connection games such as Hex use similar relations to determine whether or not specific collections of cells count as lines or connections, etc. \reffigure{Fig:SquareConnections} provides several examples of orthogonal and diagonal connections between cells, and between vertices, on square grids.

For the features discussed in this paper, we only focus on orthogonal connections. Whether or not any pair of cells or vertices are considered to be orthogonally connected in Ludii generally closely matches human intuition, in particular on graphs constructed from regular tilings (see \reffigure{Fig:SquareConnections}). We provide more formal definitions as follows:
\begin{itemize}
    \item Two cells $c_1$ and $c_2$ ($c_1 \neq c_2$) are orthogonally connected if and only if they share at least one edge.
    \item Two vertices $v_1$ and $v_2$ ($v_1 \neq v_2$) are orthogonally connected if and only if they share at least one edge.
\end{itemize}
We do not build in explicit support for other types of connections \cite{Browne_2022_Geometry}, such as diagonal connections, in our features. This is primarily to avoid the large increase in the space of features that could be represented. Note that in many cases, such as square tilings, diagonal connections may still be expressed indirectly as sequences of two orthogonal connections. Since we are, in most cases, only interested in a single type of graph element (only cells in Chess, only vertices in Go, etc.), we will frequently use the term ``\textit{site}'' to refer to one of these elements---regardless of its actual type. 

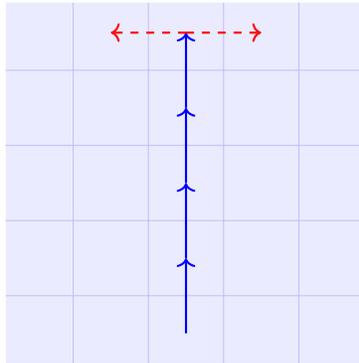
\begin{figure}
\centering

\begin{tikzpicture}

\fill[blue!8!white] (0.1,0.1) rectangle (4.9,4.9);
\draw[step=1cm,blue!25!white,very thin] (0.1,0.1) grid (4.9,4.9);

\draw[thick,->,blue] (2.5,0.5) -- (2.5,1.5);
\draw[thick,->,blue] (2.5,1.5) -- (2.5,2.5);
\draw[thick,->,blue] (2.5,2.5) -- (2.5,3.5);
\draw[thick,->,blue] (2.5,3.5) -- (2.5,4.5);

\draw[thick,->,red,dashed] (2.5,4.5) -- (3.5,4.5);
\draw[thick,->,red,dashed] (2.5,4.5) -- (1.5,4.5);

\end{tikzpicture}
\caption{Example of a radial. From the bottom to the top, a radial can be built up of consecutive orthogonal steps with absolute changes in angle less than $0.25\text{rad}$ (in fact, angles of $0\text{rad}$ in these cases). When the top row is reached, and there no longer exists any step in the same direction, there is a tie for the lowest absolute change in angle between two steps---one westwards and one eastwards. The absolute value of the change in angle ($\frac{1}{2}\pi$) exceeds $0.25\text{rad}$, so these are not valid continuations and the radial ends.}
\label{Fig:ExampleRadial}
\end{figure}

In many games, rules are not only based on the connectivity structures between sites, but also on a sense of ``continuity'' of connections. For instance, a queen in Chess may follow a sequence of many connections in a single move, but only if they all continue in the same ``direction.'' Ludii supports this by automatically computing \textit{radials} \cite{Browne_2022_Geometry}, which may intuitively be understood as sequences of consecutive steps such that they continue in the same direction. More formally, let $s_1$ and $s_2$ ($s_1 \neq s_2$) denote a pair of sites, such that there is an orthogonal step from $s_1$ to $s_2$. Let $s_3$ ($s_3 \neq s_2$) denote whichever connected site of $s_2$ has the smallest absolute change in angle between an arrow pointing from $s_1$ to $s_2$, and an arrow pointing from $s_2$ to $s_3$. Let $\theta$ denote this absolute change in angles. If $\theta < 0.25\text{rad}$, we include the step from $s_2$ to $s_3$ after the step from $s_1$ to $s_2$ in a single radial (this process may be repeated many times for a longer radial). If $\theta \geq 0.25\text{rad}$, the radial does not continue. Note that, for the computation of these angles, we assume that every site has $x$- and $y$-coordinates that define its position. This is enforced in Ludii, in part also to facilitate visualising games and game states in a graphical user interface, but---especially for arbitrary graphs---these coordinates are not necessarily guaranteed to be meaningful for gameplay. In practice, since we focus on ``real-world'' games also played by humans, such coordinates often tend to be meaningful. \reffigure{Fig:ExampleRadial} depicts an example of a full radial.

\subsubsection{Ludii State Representation}
While Ludii's state representation contains many variables \cite{Piette_2021_LudiiGameLogicGuide}, the core variables that are important to understand for the purposes of this paper are three bit arrays referred to as \texttt{empty}, \texttt{who}, and \texttt{what}. For simplicity, we assume a game that uses only a single graph element type (only cells, only edges, or only vertices), with $P \geq 1$ different players, $M \geq 1$ different piece types, and $N \geq 1$ different sites. For any give game state $s$, the three bit arrays encode the following data:
\begin{itemize}
    \item \texttt{empty}: A bit array of length $N$ with, for every site $0 \leq i < N$, a value of $1$ at index $i$ if and only if site $i$ is empty in $s$.
    \item \texttt{who}: A bit array of length $NB$, representing $N$ consecutive chunks of $B$ bits each. Here, $B = \left\lfloor 2^{\left\lceil \log_2 (X) \right\rceil} \right\rfloor$, and $X = \left\lfloor \log_2 (P+1) + 1 \right\rfloor$. $X$ is the number of bits required to encode positive integers up to and including $P+1$, and $B$ is the lowest power of $2$ that is greater than or equal to $X$. Chunks are preferred to have powers of 2 as size, because that means that a single chunk will never be split up over different 64-bit \texttt{long} values in Java when the bit array is implemented as an array of \texttt{long} values, which in turn is important for the implementation of efficient bitwise operators to extract or modify chunk values. For every site $0 \leq i < N$, the bits in the $N^{th}$ chunk are set such that they form the binary representation of the integer indicating the player that occupies site $i$, where a value of $0$ indicates a neutral piece or no occupier (empty site), and a value of $P+1$ indicates occupied by a ``shared'' piece.
    \item \texttt{what}: A bit array of length $NB$, with $X = \left\lfloor \log_2 (M) + 1 \right\rfloor$ and $B$ defined as above, such that it has $N$ consecutive chunks of $B$ bits each, where the chunks can encode positive integers up to and including $M$. For every site $0 \leq i < N$, the bits in the $N^{th}$ chunk are set such that they form the binary representation of the integer indicating the piece type that occupies site $i$, where a value of $0$ indicates no piece (i.e., an empty site).
\end{itemize}

\subsubsection{Ludii Action Representation}
Like the state representations, action representations in Ludii can contain many variables \cite{Piette_2021_LudiiGameLogicGuide}. The two variables that are most important for the purposes considered in this paper are \texttt{from} and \texttt{to}. These are typically sites (i.e., integers $0 \leq i < N$ in games with $N$ sites), which may intuitively be understood as the ``source'' (position that a piece moves away from) and ``destination'' (position that a piece moves towards or is placed on), respectively. Sometimes \texttt{from} and \texttt{to} have the same value (for instance in games such as Go, Hex, or Tic-Tac-Toe), and sometimes they have values of $-1$ (for instance for moves where players opt to pass). It is relatively uncommon, but there can be games where multiple distinct legal actions in a single state have identical \texttt{from} and \texttt{to} values. Such actions are \textit{aliased} (i.e., impossible to distinguish) when using a feature space that only accounts for these two variables.

\subsection{Representing Relative Positions in Spatial Features} \label{Subsec:Walks}

The basic premise of the spatial features $\phi(s,a)$ that we propose is that they should encode patterns, or arrangements, where certain elements (empty positions, stones, pawns, queens, etc.) are present or absent in locations relative to some particular point of interest---such as a \texttt{from} or \texttt{to} position of an action $a$. When dealing with a specific game such as Chess or Go, as in much of the related work discussed in \refsection{Sec:RelatedWork}, with a fixed and regular tiling of sites on a board, it is straightforward to define relative positions based on offsets that are applicable to any position on such a board.

\subsubsection{Describing Relative Positions as Walks}

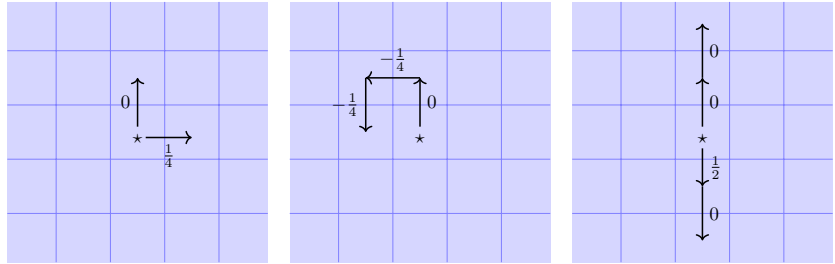
\begin{figure}
\centering

\begin{subfigure}{.3\textwidth}
\centering
\resizebox{\linewidth}{!}{
\begin{tikzpicture}

\fill[blue!16!white] (0.1,0.1) rectangle (4.9,4.9);
\draw[step=1cm,blue!50!white,very thin] (0.1,0.1) grid (4.9,4.9);

\draw[thick,->] (2.5,2.6) node[anchor=north] {$\star$} -- (2.5,3.5) node[midway,left] {$0$};
\draw[thick,->] (2.65,2.4) -- (3.5,2.4) node[midway,below] {$\frac{1}{4}$};

\end{tikzpicture}
}
\caption{Two example walks, each consisting of a single step.}
\label{Fig:ExampleWalk1}
\end{subfigure}
\begin{subfigure}{.3\textwidth}
\centering
\resizebox{\linewidth}{!}{
\begin{tikzpicture}

\fill[blue!16!white] (0.1,0.1) rectangle (4.9,4.9);
\draw[step=1cm,blue!50!white,very thin] (0.1,0.1) grid (4.9,4.9);

\draw[thick,->] (2.5,2.6) node[anchor=north] {$\star$} -- (2.5,3.5) node[midway,right] {$0$};
\draw[thick,->] (2.5,3.5) -- (1.5,3.5) node[midway,above] {$-\frac{1}{4}$};
\draw[thick,->] (1.5,3.5) -- (1.5,2.5) node[midway,left] {$-\frac{1}{4}$};

\end{tikzpicture}
}
\caption{A single example walk, consisting of three steps.}
\label{Fig:ExampleWalk2}
\end{subfigure}
\begin{subfigure}{.3\textwidth}
\centering
\resizebox{\linewidth}{!}{
\begin{tikzpicture}

\fill[blue!16!white] (0.1,0.1) rectangle (4.9,4.9);
\draw[step=1cm,blue!50!white,very thin] (0.1,0.1) grid (4.9,4.9);

\draw[thick,->] (2.5,2.6) node[anchor=north] {$\star$} -- (2.5,3.5) node[midway,right] {$0$};
\draw[thick,->] (2.5,3.5) -- (2.5,4.5) node[midway,right] {$0$};
\draw[thick,->] (2.5,2.2) -- (2.5,1.5) node[midway,right] {$\frac{1}{2}$};
\draw[thick,->] (2.5,1.5) -- (2.5,0.5) node[midway,right] {$0$};

\end{tikzpicture}
}
\caption{Two example walks, each consisting of two steps.}
\label{Fig:ExampleWalk3}
\end{subfigure}
\caption{Several example walks, describing sites relative to some starting point marked by $\star$, on regular tilings of sites with four orthogonal connections each (i.e., square cells).  (\subref{Fig:ExampleWalk1}) A step with $\rho=0$ travels northwards, whereas a step with $\rho=\frac{1}{4}$ travels eastwards, due to a $360\degree \times \frac{1}{4} = 90\degree$ rotation clockwise from the ``default'' direction. (\subref{Fig:ExampleWalk2}) Rotations are relative to the ``current'' direction. Hence, after the second step of this walk with a rotation $\rho=-\frac{1}{4}$ already resulted in a westwards facing, an additional rotation of $-\frac{1}{4}$ results in a southwards step. (\subref{Fig:ExampleWalk3}) A two-step $\{\frac{1}{2}, 0\}$ walk may be viewed as a vertically reflected, or a rotated (by $180\degree$) version of a two-step $\{0 ,0\}$ walk.}
\label{Fig:ExampleWalks}
\end{figure}

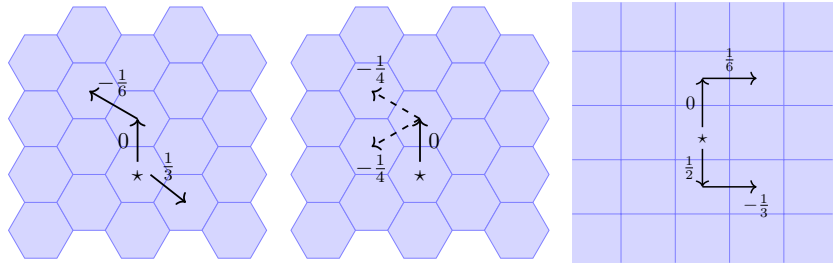
\begin{figure}
\centering

\begin{subfigure}{.3\textwidth}
\centering
\resizebox{\linewidth}{!}{
\begin{tikzpicture} [hexa/.style= {shape=regular polygon,regular polygon sides=6,minimum size=1cm,very thin,blue!50!white,draw,inner sep=0,anchor=south,fill=blue!16!white}]
\foreach \j in {0,...,4}{%
     \ifodd\j 
         \foreach \i in {0,...,3}{\node[hexa] (h\j;\i) at ({\j/2+\j/4},{(\i+1/2)*sin(60)}) {};}
    \else
         \foreach \i in {0,...,3}{\node[hexa] (h\j;\i) at ({\j/2+\j/4},{\i*sin(60)}) {};}
    \fi}  
    
\draw[thick,->] (1.5,{0.2+(3/2)*sin(60)}) node[anchor=north] {$\star$} -- (1.5,{2.5*sin(60)}) node[midway,left] {$0$};
\draw[thick,->] (1.5,{2.5*sin(60)}) node[anchor=north] {} -- (0.75,{3*sin(60)}) node[midway,above] {$-\frac{1}{6}$};
\draw[thick,->] (1.7,{(3/2)*sin(60)}) node[anchor=north] {} -- (2.25,{sin(60)}) node[midway,above] {$\frac{1}{3}$};
\end{tikzpicture}
}
\caption{Two example walks on a grid of hexagonal cells.}
\label{Fig:ExampleWalksHex1}
\end{subfigure}
\begin{subfigure}{.3\textwidth}
\centering
\resizebox{\linewidth}{!}{
\begin{tikzpicture} [hexa/.style= {shape=regular polygon,regular polygon sides=6,minimum size=1cm,very thin,blue!50!white,draw,inner sep=0,anchor=south,fill=blue!16!white}]
\foreach \j in {0,...,4}{%
     \ifodd\j 
         \foreach \i in {0,...,3}{\node[hexa] (h\j;\i) at ({\j/2+\j/4},{(\i+1/2)*sin(60)}) {};}
    \else
         \foreach \i in {0,...,3}{\node[hexa] (h\j;\i) at ({\j/2+\j/4},{\i*sin(60)}) {};}
    \fi}  
    
\draw[thick,->] (1.5,{0.2+(3/2)*sin(60)}) node[anchor=north] {$\star$} -- (1.5,{2.5*sin(60)}) node[midway,right] {$0$};
\draw[thick,dashed,->] (1.5,{2.5*sin(60)}) node[anchor=north] {} -- (0.75,{3*sin(60)}) node[above] {$-\frac{1}{4}$};
\draw[thick,dashed,->] (1.5,{2.5*sin(60)}) node[anchor=north] {} -- (0.75,{2*sin(60)}) node[below] {$-\frac{1}{4}$};
\end{tikzpicture}
}
\caption{Example walk with two possible destinations from $\star$.}
\label{Fig:ExampleWalksHex2}
\end{subfigure}
\begin{subfigure}{.3\textwidth}
\centering
\resizebox{\linewidth}{!}{
\begin{tikzpicture}

\fill[blue!16!white] (0.1,0.1) rectangle (4.9,4.9);
\draw[step=1cm,blue!50!white,very thin] (0.1,0.1) grid (4.9,4.9);

\draw[thick,->] (2.5,2.6) node[anchor=north] {$\star$} -- (2.5,3.5) node[midway,left] {$0$};
\draw[thick,->] (2.5,3.5) -- (3.5,3.5) node[midway,above] {$\frac{1}{6}$};
\draw[thick,->] (2.5,2.2) -- (2.5,1.5) node[midway,left] {$\frac{1}{2}$};
\draw[thick,,->] (2.5,1.5) -- (3.5,1.5) node[below] {$-\frac{1}{3}$};

\end{tikzpicture}
}
\caption{Two example walks resolved on square cells.}
\label{Fig:ExampleHexWalksSquareBoard}
\end{subfigure}
\caption{Several more complex example walks. (\subref{Fig:ExampleWalksHex1}) Because hexagonal cells have six orthogonal connections each, rotations are most naturally expressed as multiples of $\frac{1}{6}$. (\subref{Fig:ExampleWalksHex2}) If a walk has a step with a rotation of $-\frac{1}{4}$ on a grid of hexagonal cells, this can be interpreted as either $-\frac{1}{6}$ or $-\frac{1}{3}$, because it is perfectly in the middle of those two values. Hence, a single walk can ``split'' into two different walks (depicted with dashed arrows), and it can represent either of the two destinations as position relative to $\star$. (\subref{Fig:ExampleHexWalksSquareBoard}) On a grid of square cells, rotations of $(-)\frac{1}{6}$ and $(-)\frac{1}{3}$ both resolve to the same closest ``natural'' rotation of $(-)\frac{1}{4}$.}
\label{Fig:ExampleWalksNotSquare}
\end{figure}

For general games, we propose to define relative positions as \textit{walks} of $0$ or more steps along orthogonal connections, where every step is represented by a real number $-1 \leq \rho \leq 1$ that describes the ratio of a $360\degree$ clockwise turn that should be taken, relative to a ``current'' direction, before taking the next step. To make this easier to understand, \reffigure{Fig:ExampleWalks} provides several examples of different walks on a grid of square cells. 
Cases where we know in advance that we are dealing with a regular tiling of sites, which all have the same number of connections, are still the easiest cases to work with.
For example, all example walks in \reffigure{Fig:ExampleWalks} use rotations that are multiples of $\frac{1}{4}$, since these are the most natural rotations to use given four-sided cells that have four orthogonal connections each. Other rotations may be more natural on other boards, such as multiples of $\frac{1}{6}$ for tilings of hexagonal cells (see \reffigure{Fig:ExampleWalksHex1}). Note that constraining rotations $\rho$ to a smaller range of values, such as $\rho \in [0, 1)$, would still be sufficient and equally representative. However, we sometimes find the larger $[-1, 1]$ range to be more convenient and easier to use when manually reading or writing patterns for testing or development purposes.

Rotations other than these ``most natural'' rotations can still be used, simply by rounding them to the closest natural rotation for any given site. If a rotation is exactly\footnote{In practice we use a tolerance value of $\epsilon=0.02$.} in the middle of two consecutive natural rotations for a given site, we ``split up'' the walk into two different walks; one for each of the rotations we can round to. This behaviour is illustrated in \reffigure{Fig:ExampleWalksNotSquare}. This functionality can be useful for transfer between games with different board geometries, and also for games played on boards that are not regular tilings (such as Fractal; see \reffigure{Fig:Fractal}). This representation is intended to preserve spatial semantics as closely as possible even when resolving walks in situations where some of the described rotations do not match the natural rotations of the sites involved. For example, rotations of $0$ and $\frac{1}{2}$ will always be ``opposites'' of each other, and a rotation of $\frac{1}{4}$ (relative to a default northwards facing) will always face approximately eastwards---regardless of whether it is evaluated on a site that has $4$ orthogonal connections, or $400$ or any other number of orthogonal connections.

\subsubsection{Representing Off-board Space} \label{Subsubsec:OffBoard}

Many game's rules do not solely involve positions and connectivity relations defined as an arbitrary graph, but also incorporate a sense of ``direction.'' As described in \refsubsubsection{Subsubsec:LudiiBoardGeometry}, Ludii automatically computes \textit{radials}, which represent sequences of steps that follow along a single direction. However, as depicted in \reffigure{Fig:ExampleRadial}, these radials stop when the border of a game board is reached. This also leads to the issue that, for example, all the sites along the border of a square tiling only have three orthogonal connections (or two for the corners), whereas inner sites have four connections. Arguably a more natural representation would introduce additional connections leading to ``off-board'' positions, such that we can detect at what point of a walk a step in a certain direction would cause us to wander ``off the board.'' While it may be considered obvious where such additional connections would have to be added for certain specific cases, such as regular square tilings, this is not always true for more unconventional boards, or arbitrary graphs, which may also be used in Ludii. To support more general cases, we propose a series of steps to automatically compute such off-board connections, based on ``extending'' existing radials past their stopping point. While we cannot guarantee that this produces the expected result in any arbitrary case (or even just formally define what the expected result would be in any arbitrary case), we can demonstrate the results for a variety of common cases.

\begin{figure}
\centering

\newcommand*\rows{5}
\begin{tikzpicture}[scale=1.75]
    \filldraw[draw=blue!25!white,very thin,fill=blue!8!white] (1,0) -- ($(\rows-1,0)+0*(-0.5, {0.5*sqrt(3)})$) -- ($(\rows/2,{\rows/2*sqrt(3)})+{\rows-1}*(0.5,{-0.5*sqrt(3)})$) -- ($(0.5,{-0.5*sqrt(3)})+{\rows}*(0.5,{0.5*sqrt(3)})$) -- ($({-0.5+\rows/2},{(-0.5*sqrt(3))+(\rows/2*sqrt(3))})$) -- ($(0.5,{0.5*sqrt(3)})$) -- cycle;

    \draw[blue!25!white,very thin] (1,0) -- ($(\rows-1,0)+0*(-0.5, {0.5*sqrt(3)})$);
    \foreach \row in {1, 2, ...,\rows} {
        \draw[blue!25!white,very thin] ($\row*(0.5, {0.5*sqrt(3)})$) -- ($(\rows,0)+\row*(-0.5, {0.5*sqrt(3)})$);
    }
    
    \draw[blue!25!white,very thin] ($(0.5,{0.5*sqrt(3)})$) -- ($({-0.5+\rows/2},{(-0.5*sqrt(3))+(\rows/2*sqrt(3))})$);
    \foreach \row in {1, 2, ...,\rows} {
        \draw[blue!25!white,very thin] ($\row*(1, 0)$) -- ($(\rows/2,{\rows/2*sqrt(3)})+\row*(0.5,{-0.5*sqrt(3)})$);
    }

    \draw[blue!25!white,very thin] ($({\rows-0.5}, {0.5*sqrt(3)})$) -- ($(0.5,{-0.5*sqrt(3)})+{\rows}*(0.5,{0.5*sqrt(3)})$);
    \foreach \row in {1, ...,\rows} {
        \draw[blue!25!white,very thin] (${(\row-1)}*(1, 0)$) -- ($(0,0)+{(\row-1)}*(0.5,{0.5*sqrt(3)})$);
    }
    
    \coordinate (origin) at ($(1.5,{(0.5*sqrt(3))/3})$);
    \coordinate (leftpoint) at ($(1,{sqrt(3)/3})$);
    \coordinate (rightpoint) at ($(2,{sqrt(3)/3})$);
    \coordinate (dummy) at ($(1.5,{(-1*sqrt(3))/3})$);
    
    \draw[thick,->,blue] (origin) -- (rightpoint);
    \draw[thick,->,blue] (rightpoint) -- ($(3,{(2*sqrt(3))/3})$);
    \draw[thick,->,blue] ($(3,{(2*sqrt(3))/3})$) -- ($(3.5,{(2.5*sqrt(3))/3})$);
    
    \draw[thick,->,blue] (origin) -- (leftpoint);
    
    \pic [draw, ->, "$\frac{2\pi}{3}$", angle eccentricity=1.5] {angle = rightpoint--origin--leftpoint};
    
    \draw[thick,->,green!40!gray,dashed] (origin) -- (dummy);
    
    \pic [draw, ->, "$\frac{2\pi}{3}$", angle eccentricity=1.5] {angle = leftpoint--origin--dummy};
    
    \pic [draw, ->, "$\frac{2\pi}{3}$", angle eccentricity=1.5] {angle = dummy--origin--rightpoint};
\end{tikzpicture}
\caption{The solid blue arrows represent two radials that each start with an orthogonal step, originating from the same cell. There is an angle of $\frac{2\pi}{3}\text{rad}$ between them. We insert an artificial ``off-board'' connection, represented by the green dashed arrow, to complete the circle.}
\label{Fig:CompleteTriangle}
\end{figure}
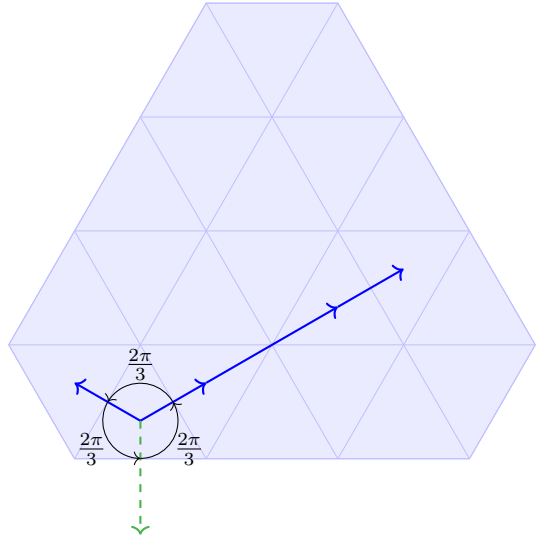

\paragraph{1. Completing triangles}
Whenever a site has only two outgoing radials that start with an orthogonal step, with an angle of $\frac{2\pi}{3}$ between them, we insert an additional off-board step at another $\frac{2\pi}{3}$ angle. This ``completes'' triangular cells along a board edge, as depicted in \reffigure{Fig:CompleteTriangle}.

\paragraph{2. Continuing opposite radials} 

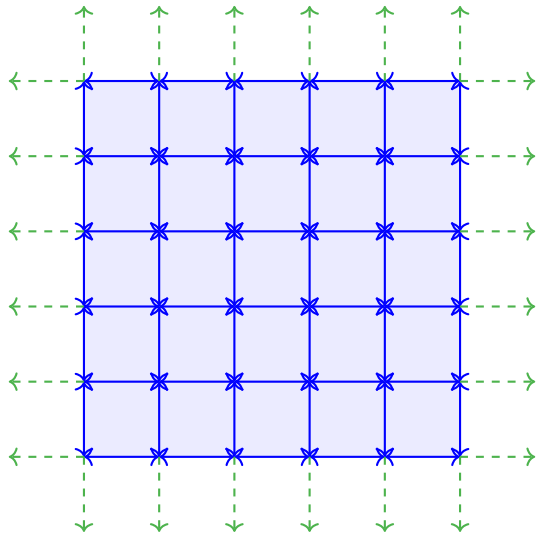
\begin{figure}
\centering
\begin{tikzpicture}
\fill[blue!8!white] (0,0) rectangle (5,5);
\draw[step=1cm,blue!25!white,very thin] (0,0) grid (5,5);

\foreach \j in {0,...,5}{%
    \foreach \i in {1,...,5}{%
        \draw[<->,blue,thick] (\j,\i) -- (\j,\i-1);
    }
}

\foreach \j in {0,...,4}{%
    \foreach \i in {0,...,5}{%
        \draw[<->,blue,thick] (\j,\i) -- (\j+1,\i);
    }
}

\foreach \j in {0,...,5}{%
    \draw[->,green!40!gray,dashed,thick] (\j,5) -- (\j,6);
    \draw[->,green!40!gray,dashed,thick] (\j,0) -- (\j,-1);
    \draw[->,green!40!gray,dashed,thick] (5,\j) -- (6,\j);
    \draw[->,green!40!gray,dashed,thick] (0,\j) -- (-1,\j);
}
\end{tikzpicture}
\caption{The solid blue arrows represent radials for orthogonal steps on a $6$$\times$$6$ Go board (assuming play on the intersections). The dashed green arrows represent off-board continuations of these radials.}
\label{Fig:ContinueOppositeRadials}
\end{figure}

Let $s_1$ denote a site with an orthogonal step to another site $s_2$, such that $s_2$ has a radial back to $s_1$ that does not extend beyond $s_1$. In such a case, we insert an artificial off-board connection from $s_1$, in the same direction that the step from $s_2$ to $s_1$ points to. This handles common cases such as sites along the edge of a chessboard or Go board, as depicted in \reffigure{Fig:ContinueOppositeRadials}. This step is skipped if the previous step (for completing triangles) already introduced new connections.

\begin{figure}
\centering
\begin{tikzpicture}[hexa/.style= {shape=regular polygon,regular polygon sides=6,minimum size=1cm,very thin,blue!25!white,draw,inner sep=0,anchor=south,fill=blue!8!white}]
\foreach \i in {0,...,3}{\node[hexa] (h0;\i) at ({0/2+0/4},{\i*sin(60)}) {};}

\foreach \i in {0,...,2}{\node[hexa] (h1;\i) at ({1/2+1/4},{(\i+1/2)*sin(60)}) {};}

\foreach \i in {1,...,2}{\node[hexa] (h2;\i) at ({2/2+2/4},{\i*sin(60)}) {};}

\node[hexa] (h3;1) at ({3/2+3/4},{(1+1/2)*sin(60)}) {};

\draw[->,blue,thick] (0,{0.5*sin(60)}) -- (0.75,{sin(60)});
\draw[->,blue,thick] (0.75,{sin(60)}) -- (1.5,{(3/2)*sin(60)});
\draw[->,blue,thick] (1.5,{(3/2)*sin(60)}) -- (2.25,{2*sin(60)});
\draw[->,blue,thick] (2.25,{2*sin(60)}) -- (3,{2.5*sin(60)});

\draw[->,blue,thick] (0,{3.5*sin(60)}) -- (0.75,{3*sin(60)});
\draw[->,blue,thick] (0.75,{3*sin(60)}) -- (1.5,{2.5*sin(60)});
\draw[->,blue,thick] (1.5,{2.5*sin(60)}) -- (2.25,{2*sin(60)});
\draw[->,blue,thick] (2.25,{2*sin(60)}) -- (3,{1.5*sin(60)});

\draw[->,green!40!gray,dashed,thick] (2.25,{2*sin(60)}) -- (2.25,{3*sin(60)});
\draw[->,green!40!gray,dashed,thick] (2.25,{2*sin(60)}) -- (2.25,{sin(60)});

\coordinate (M) at (2.25,{2*sin(60)});
\coordinate (A) at (2.25,{3*sin(60)});
\coordinate (B) at (3,{2.5*sin(60)});
\coordinate (C) at (3,{1.5*sin(60)});
\coordinate (D) at (2.25,{sin(60)});
\coordinate (E) at (1.5,{(3/2)*sin(60)});
\coordinate (F) at (1.5,{2.5*sin(60)});

\pic [draw, ->, "$\frac{\pi}{3}$", angle eccentricity=1.5] {angle = B--M--A};
\pic [draw, ->, "$\frac{\pi}{3}$", angle eccentricity=1.5] {angle = A--M--F};
\pic [draw, ->, "$\frac{\pi}{3}$", angle eccentricity=1.5] {angle = F--M--E};
\pic [draw, ->, "$\frac{\pi}{3}$", angle eccentricity=1.5] {angle = E--M--D};
\pic [draw, ->, "$\frac{\pi}{3}$", angle eccentricity=1.5] {angle = D--M--C};
\pic [draw, ->, "$\frac{\pi}{3}$", angle eccentricity=1.5] {angle = C--M--B};
\end{tikzpicture}
\caption{The solid blue arrows represent radials, plus two off-board continuation steps, for the rightmost cell. The dashed green arrows represent two additional off-board connections added to ensure uniform angles between adjacent connections.}
\label{Fig:UniformAngles}
\end{figure}
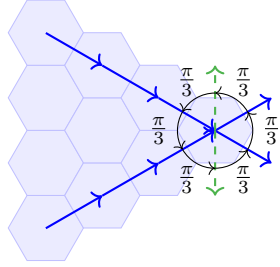

\paragraph{3. Encouraging uniform angles} While it is possible to play games on arbitrary graphs with arbitrary structures, the vast majority of games use regular or semiregular tilings with regular polygons. Hence, as a heuristic, we prefer uniform angles between all outgoing radials starting with orthogonal steps for any single given site. For any site to which some artificial connections were already added in the previous step, we introduce additional ones if this can ensure that the angles between any pair of adjacent connections becomes equal (not allowing the total number of connections to be more than doubled). \reffigure{Fig:UniformAngles} depicts, as an example, how this affects two of the corners of a rhombus-shaped board of hexagonal cells, as used in the game of Hex.

\paragraph{Walks into off-board space}
Using the additional off-board connections as described above, we can detect when a walk wanders off the board. However, beyond those steps, the ``off-board'' space is not further modelled with additional sites and connections. This means that, once a walk wanders off the board, it can never return; the final destination of such a walk is always assumed to be an ``off-board'' position.

\subsection{Representing Actions and State Elements in Spatial Features} \label{Subsec:SpatialFeatureFormat}

A spatial state-action feature $\phi(s, a)$ must encode relevant (spatial) aspects of the action $a$, as well as some elements that must be present or absent in the game state $s$, typically in a local area around the sites affected by $a$, for the feature to be considered active. For every such feature, we leave one variable site referred to as the \textit{anchor}, denoted $\star$. Any other aspects are defined relative to $\star$ using walks, as explained in \refsubsection{Subsec:Walks}. Firstly, we define the following four action-related properties, for which any feature $\phi(s, a)$ can specify that they must be present in positions $i$ relative to an anchor $\star$ for $\phi(s, a)$ to be considered active:
\begin{itemize}
    \item \texttt{to}: Requires that the \texttt{to} position of action $a$ coincides with the position $i$ relative to $\star$.
    \item \texttt{from}: Requires that the \texttt{from} position of action $a$ coincides with the position $i$ relative to $\star$.
    \item \texttt{last to}: Requires that the \texttt{to} position of the last action $a'$, which led to $s$, coincides with the position $i$ relative to $\star$.
    \item \texttt{last from}: Requires that the \texttt{from} position of the last action $a'$, which led to $s$, coincides with the position $i$ relative to $\star$.
\end{itemize}
Any state-action feature $\phi(s, a)$ must have at least a specifier for either \texttt{to} or \texttt{from}---because otherwise it would simply be a state feature $\phi(s)$ that could not distinguish between any actions---and it can have specifiers for both. If a feature has specifiers for either \texttt{last to} or \texttt{last from} (or both), we refer to it as a \textit{reactive} feature. Reactive features are generally more efficient to use because they only need to be evaluated in a local area around the last action, which they ``react'' to.

Secondly, we define several state properties, for which any feature $\phi(s, a)$ can specify that they must be present or absent in positions $i$ relative to an anchor $\star$ for $\phi(s, a)$ to be considered active. Note that for each of these, we consider affirmative as well as negated variants:
\begin{itemize}
    \item \texttt{empty}: Requires that the position $i$ relative to $\star$ is (not) empty, i.e. the $i^{th}$ bit in the \texttt{empty} bit array of $s$ must (not) be set to $1$.
    \item \texttt{friend}: Requires that the position $i$ relative to $\star$ is (not) occupied by a friendly piece, i.e. the bits in the $i^{th}$ chunk of the \texttt{who} bit array of $s$ must (not) represent the integer value $p$, where $p$ is the player to move.
    \item \texttt{enemy}: Requires that the position $i$ relative to $\star$ is (not) occupied by an enemy piece, i.e. the bits in the $i^{th}$ chunk of the \texttt{who} bit array of $s$ must (not) represent an integer value other than $0$ or $p$, where $p$ is the player to move.
    \item \texttt{off}: Requires that the walk specifying $i$, relative to $\star$, leads to a position that is (not) ``off the board'' (see \refsubsubsection{Subsubsec:OffBoard}).
    \item \texttt{item}: Requires that the position $i$ relative to $\star$ is (not) occupied by a specific piece type with a specified index $k$, i.e. the bits in the $i^{th}$ chunk of the \texttt{what} bit array of $s$ must (not) represent the integer value $k$.
    \item \texttt{connectivity}: Requires that the position $i$ relative to $\star$ is (not) a site that has a specified number $k$ of orthogonal connections.
    \item \texttt{region proximity}: Requires that the position $i$ relative to $\star$ is (not) closer than $\star$ to the \textit{region} defined in the game with a specified index $k$. Regions are predefined sets of sites that typically have a specific, important meaning in a game's rules (e.g., a region of sites that a player must reach or to win).
\end{itemize}

Note that these state and action properties are not necessarily sufficient to perfectly distinguish all unique states and actions in all games; in some games, there may be pairs of distinct states $s$ and $s'$, or distinct actions $a$ and $a'$ that will always be indistinguishable based on only these properties. The choice to include these properties in the representation and no others is a subjective choice. This is primarily based on our intuition of which properties permit efficient evaluations of features, are relevant and important to include in some games, and also generally applicable to the extent that they are not only relevant in a small, highly specific selection of games. For most of these properties, analogous properties are also frequently included in related work on game-specific patterns (see \refsection{Sec:RelatedWork}). Similarly, approaches based on deep learning with convolutional neural networks frequently use channels encoding similar data, such as binary channels indicating presence or absence per piece type \cite{Silver_2018_AlphaZero,Cazenave_2020_Polygames,Soemers_2022_DeepLearning}. An example of a more advanced property that we considered is a ``line-of-sight'' property, which tests whether or not a position is in line of sight---without other obstructions in between---of a particular player or piece type. Including features with such properties was found to require an excessively large amount of memory in preliminary testing.

\subsection{Exploiting Symmetries in Features}

A common approach to improve sample efficiency and training speed in AI for games, as well as machine learning more generally, is to exploit various symmetries through weight sharing. For example, a significant amount of related work on patterns in games such as Go uses translational weight sharing; the same weight is associated with a pattern, regardless of where it appears on the game board. Other forms of symmetry that are frequently leveraged include reflection, rotation, and player colour inversion \cite{Stern_2006_Bayesian,Gelly_2006_Modification,Gelly_2007_Combining,Silver_2016_AlphaGo,Silver_2017_AlphaGoZero,Lorentz_2017_Breakthrough}. Outside of game AI applications, CNNs \cite{LeCun_1989_Backpropagation}---which are ubiquitous in deep learning approaches for machine learning problems with image-based inputs---implement weight-sharing through translation equivariance. Other architectures have been developed to exploit additional symmetries in various domains \cite{Koriche_2017_Symmetry,Bronstein_2021_Geometric,Cohen_2021_Equivariant}.

Many of these symmetries are not necessarily perfectly applicable to all games. Consider, for example, a simple feature that tests for moves that capture a queen in chess. Capturing the opponent's queen may be more or less valuable depending on which position it is located in, which suggests that such a feature have different weights for different anchor positions. On the other hand, it is likely that---even if the exact position may be relevant---capturing the opponent's queen will in general be likely to be a valuable move, which suggests that training can be sped up significantly by sharing the same weight for such a feature regardless of the exact position of the anchor when it is active.

\begin{figure}
\centering
\begin{subfigure}{.32\textwidth}
  \centering
  \includegraphics[width=\linewidth]{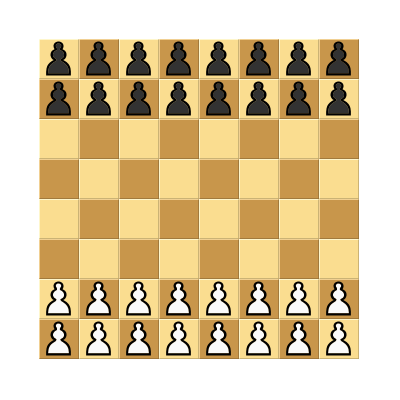}
  \caption{Breakthrough.}
  \label{Fig:Breakthrough}
\end{subfigure}
\begin{subfigure}{.32\textwidth}
  \centering
  \includegraphics[width=\linewidth]{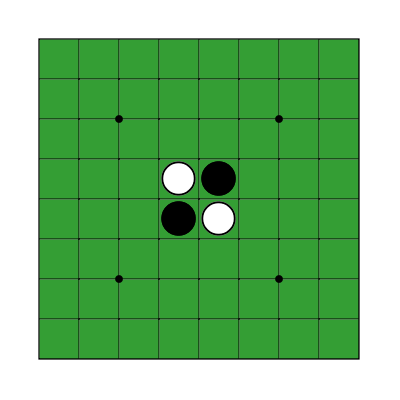}
  \caption{Reversi.}
  \label{Fig:Reversi}
\end{subfigure}
\begin{subfigure}{.32\textwidth}
  \centering
  \includegraphics[width=\linewidth]{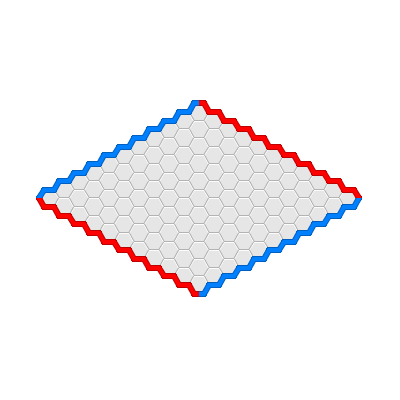}
  \caption{Hex.}
  \label{Fig:Hex}
\end{subfigure}
\caption{Three examples games where certain symmetries may not be applicable.}
\label{Fig:ExampleGamesSymmetries}
\end{figure}

In this work, we share the same weight across all translations (distinct anchor positions), rotations, and reflections per feature. This improves generalisation and computational efficiency, possibly at a cost in representational capacity. We discuss three examples of well-known games where we know some of these symmetries to be ``incorrect'', and ways in which negative effects on the representational capacity of policies based on our features are mitigated in these games:
\begin{enumerate}
    \item In the game of Breakthrough (see \reffigure{Fig:Breakthrough}), players are only allowed to moves their pawns ``forwards'' (diagonally or orthogonally), towards the board edge opposite their starting positions. This suggests that patterns that are representative of strong or weak situations likely no longer represent the same after rotation or vertical reflection---except if the player colours also were to be inverted. However, our state-action features $\phi(s, a)$ also encode a representation of the action $a$, and they need only be evaluated for legal actions $a$. Because such rotations or vertical reflections would only be applicable to illegal actions, this form of weight-sharing is harmless (as well as useless) in this situation.
    \item In the game of Reversi (see \reffigure{Fig:Reversi}), it is well known that the corners of the board are particularly important to control. This suggests that some patterns may be more or less important depending on whether or not they are close to a corner, and translational weight sharing may not be appropriate. However, we include the ability to detect off-board positions relative to the anchor, and such elements can be used by features $\phi(s,a)$ to ensure that $\phi(s,a) = 1$ only if the anchor is in a particular offset relative to any corner---still allowing for generalisation between different corners through combinations of translation and rotation or reflection.
    \item In the game of Hex (see \reffigure{Fig:Hex}), one player aims to connect the northeast and southwest sides of the board, whereas the other player aims to connect the northwest and southeast sides of the board. This means that for each player, there is generally one axis along which it is significantly more valuable to extend a connection than the other axis, which means that weight sharing across some rotations may be less sensible than others \cite{Huang_2014_MoHex}. However, each of the pairs of sides per player is defined as a region in the game's description in Ludii. Therefore, the inclusion of elements that test for proximity to either of these regions in certain positions relative to the anchor can ensure that a feature can or cannot be active under certain rotations.
\end{enumerate}

\begin{figure}
\centering
\begin{subfigure}{.32\textwidth}
  \centering
  \includegraphics[width=\linewidth]{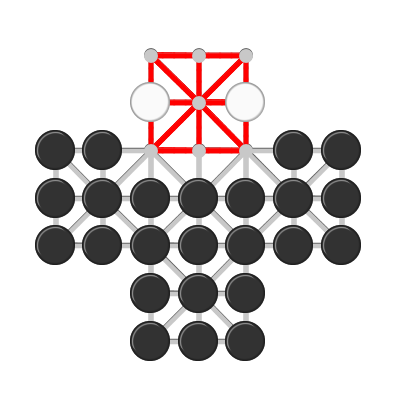}
  \caption{Asalto.}
  \label{Fig:Asalto}
\end{subfigure}
\begin{subfigure}{.32\textwidth}
  \centering
  \includegraphics[width=\linewidth]{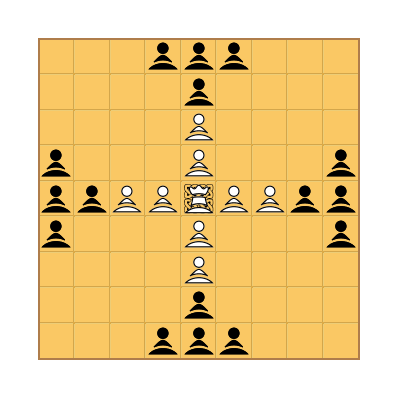}
  \caption{Tablut.}
  \label{Fig:Tablut}
\end{subfigure}
\begin{subfigure}{.32\textwidth}
  \centering
  \includegraphics[width=\linewidth]{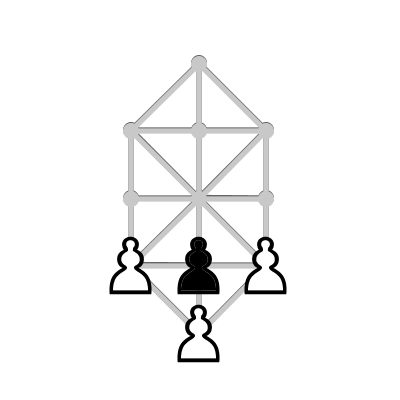}
  \caption{Jeu Militaire.}
  \label{Fig:FrenchMilitaryGame}
\end{subfigure}
\caption{Three examples of highly asymmetric games.}
\label{Fig:AsymmetricGames}
\end{figure}

In games such as Go, it is also common to leverage symmetry under colour inversion; for example, training datasets can be augmented by generating additional sample states where all black stones are changed to white stones, white stones changed to black stones, and training labels (such as the outcomes of games) similarly inverted. We do not consider any generalisation of this form in this paper, instead opting to train sets of features and weights separately for every player's perspective. One reason for this is that we also aim to train policies for games with $n > 2$ players, for which it is not immediately clear how colour inversion should work. Another reason is that, even among two-player games, there is a significant number with a high degree of asymmetry, such as many hunt games \cite{Murray_1951_History}, Tablut \cite{Linnaeus_1732_Iter}, and Jeu Militaire \cite{Lucas_1887_Recreations}; see \reffigure{Fig:AsymmetricGames}. In these games, two opposing players can have different piece types, different move rules, different initial setups, different victory conditions, etc.

\subsection{Examples of Complete Features}

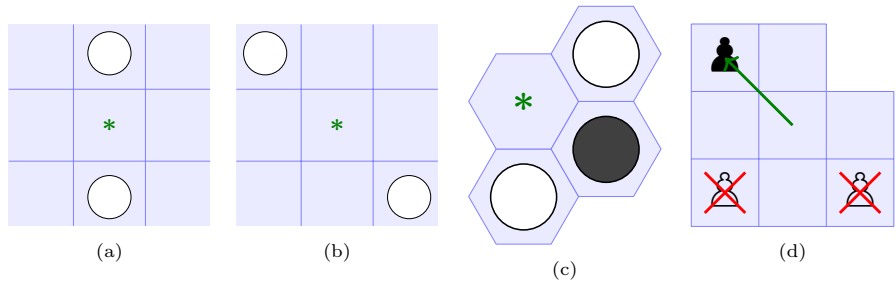
\begin{figure}
\centering

\begin{subfigure}{.24\textwidth}
\centering
\resizebox{\linewidth}{!}{
\begin{tikzpicture}
    \fill[blue!8!white] (0.1,0.1) rectangle (2.9,2.9);
    \draw[step=1cm,blue!50!white,very thin] (0.1,0.1) grid (2.9,2.9);
    \node[circle,draw=black,fill=white,inner sep=0pt,minimum size=0.6cm] at ({1.5},{2.5}) {};
    \node at ({1.5},{1.5}){\textcolor{green!50!black}{\textbf{\textasteriskcentered}}};
    \node[circle,draw=black,fill=white,inner sep=0pt,minimum size=0.6cm] at ({1.5},{0.5}) {};
\end{tikzpicture}
}
\caption{}
\label{Fig:ExampleFeature1}
\end{subfigure}
\begin{subfigure}{.24\textwidth}
\centering
\resizebox{\linewidth}{!}{
\begin{tikzpicture}
        \fill[blue!8!white] (0.1,0.1) rectangle (2.9,2.9);
        \draw[step=1cm,blue!50!white,very thin] (0.1,0.1) grid (2.9,2.9);
        \node[circle,draw=black,fill=white,inner sep=0pt,minimum size=0.6cm] at ({0.5},{2.5}) {};
        \node at ({1.5},{1.5}){\textcolor{green!50!black}{\textbf{\textasteriskcentered}}};
        \node[circle,draw=black,fill=white,inner sep=0pt,minimum size=0.6cm] at ({2.5},{0.5}) {};
\end{tikzpicture}
}
\caption{}
\label{Fig:ExampleFeature2}
\end{subfigure}
\begin{subfigure}{.24\textwidth}
\centering
\resizebox{\linewidth}{!}{
\begin{tikzpicture} [hexa/.style= {shape=regular polygon,regular polygon sides=6,minimum size=1cm,very thin,blue!50!white,draw,inner sep=0,anchor=south,fill=blue!8!white}]
\foreach \j in {1,...,2}{%
     \ifodd\j 
         \foreach \i in {0,...,1}{
            \node[hexa] (h\j;\i) at ({\j/2+\j/4},{(\i+1/2)*sin(60)}) {\ifthenelse{\i=1}{\textcolor{green!50!black}{\textbf{\textasteriskcentered}}}{}};
          }
    \else
         \foreach \i in {1,...,2}{
            \node[hexa] (h\j;\i) at ({\j/2+\j/4},{\i*sin(60)}) {}; 
          }
    \fi}  
\node[circle,draw=black,fill=white,inner sep=0pt,minimum size=0.6cm] at ({1.5},{2.5*sin(60)}){};
\node[circle,draw=black,fill=black!75!white,inner sep=0pt,minimum size=0.6cm] at ({1.5},{1.5*sin(60)}){};
\node[circle,draw=black,fill=white,inner sep=0pt,minimum size=0.6cm] at ({0.75},{1*sin(60)}){};
\end{tikzpicture}
}
\caption{}
\label{Fig:ExampleFeature3}
\end{subfigure}
\begin{subfigure}{.24\textwidth}
    \centering
    \resizebox{\linewidth}{!}{
    \begin{tikzpicture}
            \fill[blue!8!white] (0.0,0.0) rectangle (3.0,2.0);
            \draw[step=1cm,blue!50!white,very thin] (0.0,0.0) grid (3.0,2.0);
            \fill[blue!8!white] (0.0,2.0) rectangle (2.0,3.0);
            \draw[step=1cm,blue!50!white,very thin] (0.0,2.0) grid (2.0,3.0);
            \node at ({2.5},{0.5}) {\WhitePawnOnWhite};
            \draw[very thick,red] (2.2,0.2) -- (2.8,0.8);
            \draw[very thick,red] (2.2,0.8) -- (2.8,0.2);
            \node at ({0.5},{0.5}) {\WhitePawnOnWhite};
            \draw[very thick,red] (0.2,0.2) -- (0.8,0.8);
            \draw[very thick,red] (0.2,0.8) -- (0.8,0.2);
            \node at ({0.5},{2.5}) {\BlackPawnOnWhite};
            \draw[very thick,->,green!50!black] (1.5,1.5) -- (0.5,2.5);
    \end{tikzpicture}
    }
    \caption{}
    \label{Fig:ExampleFeature4}
\end{subfigure}
\caption{Four visualisations of examples of complete state-action features. }
\label{Fig:ExamplesFullFeatures}
\end{figure}

\reffigure{Fig:ExamplesFullFeatures} depicts visualisations of four examples of complete features that could be modelled using the design described in this section. In the first three subfigures, a green star is used to represent both the anchor and the \texttt{to} position. In the final subfigure, a green arrow represents the \texttt{from} and \texttt{to} positions, of which one is customarily used as anchor. Internally, positions of all other elements are represented by walks relative to anchors, as described previously, but explicit visualisations of these walks are omitted to improve visual clarity. \reffigure{Fig:ExampleFeature1} is a feature that matches actions placing a piece in between two existing white pieces along an orthogonal axis. This can be used to recommend players to make a winning move (if they are the white player), or block a winning move for the opponent (if they are the black player) in Tic-Tac-Toe. \reffigure{Fig:ExampleFeature2} shows a similar feature, except for diagonal rather than orthogonal lines of pieces. Depending on the perspective, \reffigure{Fig:ExampleFeature3} can be used to incentivise either completing a bridge of white pieces, or breaking one. This is an important tactic in connection games such as Hex \cite{Browne_2000_Hex,Browne_2005_Connection}. Finally, \reffigure{Fig:ExampleFeature4} recommends a diagonal movement to capture a black pawn, starting from a position that does not have a white pawn to protect it on either of the diagonal cells below it. Such a pattern is particularly useful for games such as Breakthrough (in which pawns that capture diagonally are the only piece type), but may also have some value in games with more varied piece types like Chess.

\section{Efficiently Evaluating Spatial State-Action Features} \label{Sec:MainSection}

When using state-action features $\phi(s, a)$ to guide an MCTS process, it is important that the activity level of such features for any relevant state-action pair $(s, a)$ can be computed efficiently. If this is not efficient, the computational overhead may reduce the number of iterations that can be run within any given time budget too much, which may in turn reduce the playing strength relative to an unguided, more efficient search. Efficiency is particularly important for the play-out phase, where---for the sake of efficiency---it is common to use fewer, smaller, or otherwise more efficient features or policies \cite{Gelly_2006_Modification,Coulom_2007_EloRatingsPatterns,Silver_2016_AlphaGo} than in the selection phase. Nevertheless, also in the selection phase it is naturally advantageous if features can be computed efficiently.

One common approach in computer Go is to directly index into a table, using variants of Zobrist hashing \cite{Zobrist_1970_Hashing}, to retrieve active patterns for any given state $s$ and position corresponding to an action $a$. Such an approach only works when, for any considered pattern size, only all \textit{complete} patterns---without any ``wildcard'' or ``do-not-care'' elements or positions---are used \cite{Fotland1993Knowledge,Stern_2006_Bayesian,Silver_2007_LocalShape,Gelly_2007_Combining,Araki_2007_Move}. For example, in Go this approach can be used if all patterns of a given size, such as $3$$\times$$3$, fully specify every position within that area as being either empty, black, white, or off-board. With our feature formalisation, applicable to arbitrarily-shaped graphs, this is impossible to guarantee. Furthermore, for games with many more piece types than just the two of Go, such as Chess with twelve piece types (six per player), the number of unique, fully-specified patterns that could be enumerated---even for a small area such as $3$$\times$$3$---would be excessively large.

\begin{figure}
\centering
\begin{subfigure}{.32\textwidth}
  \centering
\includegraphics[width=.9\linewidth]{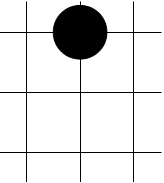}
  \caption{This pattern generalises \subref{Fig:GoPattern2} and \subref{Fig:GoPattern3}.}
  \label{Fig:GoPattern1}
\end{subfigure}
\begin{subfigure}{.32\textwidth}
  \centering
\includegraphics[width=.9\linewidth]{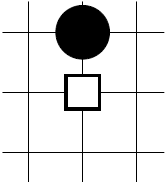}
  \caption{This pattern is generalised by \subref{Fig:GoPattern1}, but generalises \subref{Fig:GoPattern3}.}
  \label{Fig:GoPattern2}
\end{subfigure}
\begin{subfigure}{.32\textwidth}
  \centering
\includegraphics[width=.9\linewidth]{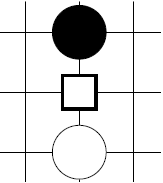}
  \caption{This pattern is generalised by \subref{Fig:GoPattern1} and \subref{Fig:GoPattern2}.}
  \label{Fig:GoPattern3}
\end{subfigure}
\caption{Three example patterns for Go in $3$$\times$$3$ windows. Small squares denote positions that must be empty for the pattern to match, whereas any other empty positions are wildcard positions, the contents of which are irrelevant to pattern matching.}
\label{Fig:GoPatterns}
\end{figure}

The second common approach for evaluating patterns in computer Go is to store patterns to be evaluated in data structures that account for generalisation relations between patterns, and avoid matching patterns that can already be inferred to not be a match based on evaluations of other patterns. Consider, for example, the three patterns for Go depicted in \reffigure{Fig:GoPatterns}. All three patterns share the same requirement for a black stone above the centre, and the last two also share the same requirement for the centre to be empty; the first pattern \textit{generalises} the last two, and the second also generalises the last. In any situation where the first pattern does not match, the last two need not be evaluated because they will for sure also not match. Similarly, the third pattern can be skipped if the second pattern does not match. This is typically implemented by storing patterns in tree structures \cite{Levinson_1991_Adaptive,Muller_1991_Pattern,Muller_1995_PhD}---such as prefix trees, where more general patterns are ``prefixes'' of more specific patterns---or deterministic finite state automata \cite{Urvoy_2002_Pattern}. The approach we propose in this section bears some resemblance to these ideas, but includes additional optimisations.

\subsection{Instantiating Features} \label{Subsec:InstantiatingFeatures}

Let $\Phi = \{ \phi_0, \phi_1, \dots, \phi_n \}$ denote a feature set of $n$ spatial features $\phi_i$, as formalised in \refsection{Sec:Formalisation}, for some arbitrary game. We aim to efficiently compute binary feature vectors $\boldsymbol{\phi}(s, a) = \rowvector{\phi_0(s,a) & \phi_1(s, a) & \dots & \phi_n(s, a)}$ for any state-action pair $(s, a)$. For every unique possible anchor position $\star$ in the given game, we \textit{instantiate} every feature $\phi_i \in \Phi$ by resolving all walks in $\phi_i$ from $\star$. After resolving the walks, we know the specific positions for which $\phi_i$ has requirements relative to the anchor $\star$. This process of instantiating the feature is repeated for all rotations and reflections, where $\star$ is assumed to represent the origin, and the number of orthogonal connections from $\star$ determines the number of rotations to consider. 
Note that feature instantiations are specific to a single player's perspective, because the meaning of ``friend'' or ``enemy'' is different for different players. For notational brevity, we leave the dependence on a specific player $p$ implicit; there is no ambiguity because the player for whom features $\phi(s,a)$ are evaluated is always the player that can select the action $a$ in the state $s$.

Recall that every feature $\phi_i$ must specify a walk for at least an action's \texttt{from} or \texttt{to} position, and it can specify walks for both. When we wish to compute which features are active for a state-action pair $(s, a)$, the anchors for which any instantiations were generated are irrelevant; rather, we require the ability to find any instantiations for which any specified \texttt{from} or \texttt{to} walks match the action $a$. Hence, we store instantiations in maps that can be accessed quickly with hash keys generated from the player index $p$, as well as any mix of \texttt{to}, \texttt{from}, \texttt{last to}, and \texttt{last from} indices. Keys including either (or both) of the last two are used to allow for fast retrieval of relevant instantiations of reactive features for any state-action pair $(s, a)$.

Let $\mathcal{F}(s, a) = \{ f_0, f_1, \dots, f_k \}$ denote such a set of feature instantiations that can be relevant to the state-action pair $(s, a)$ according to the action-related properties. Any requirements that features may have for \texttt{off}, \texttt{connectivity}, or \texttt{region proximity} elements can already be evaluated at instantiation time; instantiations with conditions that are violated can already be removed, and for the remaining instantiations these conditions need no longer be checked at runtime. This leaves only \texttt{empty}, \texttt{friend}, \texttt{enemy}, and \texttt{item} conditions to be evaluated at runtime; these can all be implemented as simple tests involving the \texttt{empty}, \texttt{who}, and \texttt{what} bit arrays of a game state $s$. \refalgorithm{Alg:NaiveFeatureEvaluation} provides pseudocode for a simple, naive algorithm to evaluate which features are active for any given state-action pair $(s, a)$ from such a set of instantiations. It simply evaluates all the conditions of every feature instantiation in sequence, which is straightforward to implement, but slow because it does not leverage any overlap in conditions or other relationships between multiple instantiations.

\begin{algorithm}
\begin{algorithmic}[1]
\Require Set of relevant feature instantiations $\mathcal{F}(s, a) = \{ f_0, f_1, \dots, f_k \}$.
\State $\boldsymbol{\phi} \gets \boldsymbol{0}$ \Comment{Init. to zero vector.}
\For{each feature instantiation $f_i \in \mathcal{F}(s, a)$} \Comment{Test whether $f_i$ is a match.}
    \State \texttt{match} $\gets$ \texttt{true}
    \For{each condition $c$ of $f_i$}
        \If{$(s, a)$ violates $c$}
            \State \texttt{match} $\gets$ \texttt{false}
            \Break
        \EndIf
    \EndFor
    \If{\texttt{match}}
        \State $j \gets$ feature index of which $f_i$ is an instantiation
        \State $\boldsymbol{\phi}\left[ j \right] \gets 1$
    \EndIf
\EndFor
\State \Return feature vector $\boldsymbol{\phi}$
\end{algorithmic}
\caption{Naive evaluation of active features.} \label{Alg:NaiveFeatureEvaluation}
\end{algorithm}

\subsection{Features as Disjunctions of Conjunctions} \label{Subsec:FeaturesAsDisjunctions}

As a first step towards a more efficient approach for computing feature vectors from a set of relevant instantiations $\mathcal{F}(s, a)$, we discuss how every feature (represented by one or more feature instantiations) may be represented as a disjunction of conjunctions of propositions (i.e., a logical formula in disjunctive normal form).

Let $f \in \mathcal{F}(s, a)$ denote any arbitrary feature instantiation in the set. Let $\phi(f)$ denote the original feature of which $f$ is an instantiation. Let $\Phi(s,a) = \{ \phi(f) \mid f \in \mathcal{F}(s,a) \}$ denote the set of all features for which there exists at least one instantiation in $\mathcal{F}(s, a)$. Let $\mathcal{C}(f) = \{ c_1, c_2, \dots, c_k \}$ denote a set of $k \geq 0$ conditions, or propositions, that must hold in the state-action pair $(s,a)$ for the feature instantiation $f$ to be considered active. We may view every such condition $c_i = \langle \texttt{site}, \texttt{bit array}, \texttt{value}, \texttt{negated} \rangle$ as a four-tuple specifying:
\begin{enumerate}
\item The \texttt{site} for which we have a condition.
\item The \texttt{bit array} (\texttt{empty}, \texttt{who}, or \texttt{what}) in which we look for a specific value.
\item The \texttt{value} that we check for in the specific site, mapping to a chunk, of the specific bit array.
\item Whether or not the condition should be \texttt{negated}; if true, the condition is satisfied if and only if the specified value is \textit{not} found in the bit array.
\end{enumerate}
A feature $\phi$ is considered active if and only if there exists at least one active instantiation: $\phi(s, a) = 1 \Leftrightarrow \exists_{f \in \mathcal{F}(s, a)} \left( \phi = \phi(f) \land \forall_{c \in \mathcal{C}(f)} \left[ (s,a) \text{ satisfies } c \right] \right)$.

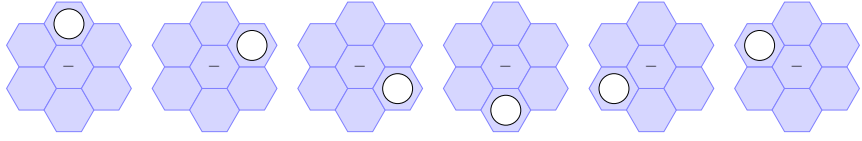
\begin{figure}
\centering

\begin{subfigure}{.15\textwidth}
\centering
\resizebox{\linewidth}{!}{
\begin{tikzpicture} [hexa/.style= {shape=regular polygon,regular polygon sides=6,minimum size=1cm,very thin,blue!50!white,draw,inner sep=0,anchor=south,fill=blue!16!white}]
\foreach \j in {0,...,2}{%
     \ifodd\j 
         \foreach \i in {0,...,2}{
            \node[hexa] (h\j;\i) at ({\j/2+\j/4},{(\i+1/2)*sin(60)}) {\ifthenelse{\i=1}{\textcolor{black}{\textbf{--}}}{}};
          }
    \else
         \foreach \i in {1,...,2}{
            \node[hexa] (h\j;\i) at ({\j/2+\j/4},{\i*sin(60)}) {};
          }
    \fi}  
\node[circle,draw=black,fill=white,inner sep=0pt,minimum size=0.6cm] (a) at ({1/2+1/4},{3*sin(60)}){};
\end{tikzpicture}
}
\end{subfigure}
\begin{subfigure}{.15\textwidth}
\centering
\resizebox{\linewidth}{!}{
\begin{tikzpicture} [hexa/.style= {shape=regular polygon,regular polygon sides=6,minimum size=1cm,very thin,blue!50!white,draw,inner sep=0,anchor=south,fill=blue!16!white}]
\foreach \j in {0,...,2}{%
     \ifodd\j 
         \foreach \i in {0,...,2}{
            \node[hexa] (h\j;\i) at ({\j/2+\j/4},{(\i+1/2)*sin(60)}) {\ifthenelse{\i=1}{\textcolor{black}{\textbf{--}}}{}};
          }
    \else
         \foreach \i in {1,...,2}{
            \node[hexa] (h\j;\i) at ({\j/2+\j/4},{\i*sin(60)}) {};
          }
    \fi}  
\node[circle,draw=black,fill=white,inner sep=0pt,minimum size=0.6cm] (a) at ({3/2},{2.5*sin(60)}){};
\end{tikzpicture}
}
\end{subfigure}
\begin{subfigure}{.15\textwidth}
\centering
\resizebox{\linewidth}{!}{
\begin{tikzpicture} [hexa/.style= {shape=regular polygon,regular polygon sides=6,minimum size=1cm,very thin,blue!50!white,draw,inner sep=0,anchor=south,fill=blue!16!white}]
\foreach \j in {0,...,2}{%
     \ifodd\j 
         \foreach \i in {0,...,2}{
            \node[hexa] (h\j;\i) at ({\j/2+\j/4},{(\i+1/2)*sin(60)}) {\ifthenelse{\i=1}{\textcolor{black}{\textbf{--}}}{}};
          }
    \else
         \foreach \i in {1,...,2}{
            \node[hexa] (h\j;\i) at ({\j/2+\j/4},{\i*sin(60)}) {};
          }
    \fi}  
\node[circle,draw=black,fill=white,inner sep=0pt,minimum size=0.6cm] (a) at ({3/2},{1.5*sin(60)}){};
\end{tikzpicture}
}
\end{subfigure}
\begin{subfigure}{.15\textwidth}
\centering
\resizebox{\linewidth}{!}{
\begin{tikzpicture} [hexa/.style= {shape=regular polygon,regular polygon sides=6,minimum size=1cm,very thin,blue!50!white,draw,inner sep=0,anchor=south,fill=blue!16!white}]
\foreach \j in {0,...,2}{%
     \ifodd\j 
         \foreach \i in {0,...,2}{
            \node[hexa] (h\j;\i) at ({\j/2+\j/4},{(\i+1/2)*sin(60)}) {\ifthenelse{\i=1}{\textcolor{black}{\textbf{--}}}{}};
          }
    \else
         \foreach \i in {1,...,2}{
            \node[hexa] (h\j;\i) at ({\j/2+\j/4},{\i*sin(60)}) {};
          }
    \fi}  
\node[circle,draw=black,fill=white,inner sep=0pt,minimum size=0.6cm] (a) at ({1/2+1/4},{sin(60)}){};
\end{tikzpicture}
}
\end{subfigure}
\begin{subfigure}{.15\textwidth}
\centering
\resizebox{\linewidth}{!}{
\begin{tikzpicture} [hexa/.style= {shape=regular polygon,regular polygon sides=6,minimum size=1cm,very thin,blue!50!white,draw,inner sep=0,anchor=south,fill=blue!16!white}]
\foreach \j in {0,...,2}{%
     \ifodd\j 
         \foreach \i in {0,...,2}{
            \node[hexa] (h\j;\i) at ({\j/2+\j/4},{(\i+1/2)*sin(60)}) {\ifthenelse{\i=1}{\textcolor{black}{\textbf{--}}}{}};
          }
    \else
         \foreach \i in {1,...,2}{
            \node[hexa] (h\j;\i) at ({\j/2+\j/4},{\i*sin(60)}) {};
          }
    \fi}  
\node[circle,draw=black,fill=white,inner sep=0pt,minimum size=0.6cm] (a) at ({0},{1.5*sin(60)}){};
\end{tikzpicture}
}
\end{subfigure}
\begin{subfigure}{.15\textwidth}
\centering
\resizebox{\linewidth}{!}{
\begin{tikzpicture} [hexa/.style= {shape=regular polygon,regular polygon sides=6,minimum size=1cm,very thin,blue!50!white,draw,inner sep=0,anchor=south,fill=blue!16!white}]
\foreach \j in {0,...,2}{%
     \ifodd\j 
         \foreach \i in {0,...,2}{
            \node[hexa] (h\j;\i) at ({\j/2+\j/4},{(\i+1/2)*sin(60)}) {\ifthenelse{\i=1}{\textcolor{black}{\textbf{--}}}{}};
          }
    \else
         \foreach \i in {1,...,2}{
            \node[hexa] (h\j;\i) at ({\j/2+\j/4},{\i*sin(60)}) {};
          }
    \fi}  
\node[circle,draw=black,fill=white,inner sep=0pt,minimum size=0.6cm] (a) at ({0},{2.5*sin(60)}){};
\end{tikzpicture}
}
\end{subfigure}
\caption{Six different instantiations, each with the same position (centre of the depicted area) as anchor, of a single feature that requires the anchor to be empty (indicated by ``\textbf{--}''), and an adjacent position to contain a white stone. This means that every instantiation is a conjunction of two conditions, where the condition for the empty position is identical for each of the instantiations. The feature is a disjunction of these six conjunctions.}
\label{Fig:HexFeatureInstantiations}
\end{figure}

Since a feature instantiation is considered active if and only if all of its conditions are satisfied, it can be represented simply as a conjunction of its conditions: $C = c_1 \land c_2 \land \dots \land c_k$. Since a feature only requires any one of its instantiations to be active for the feature to be considered active, the feature can be represented as a disjunction of several such conjunctions: $D = \left( c_i \land \dots \ c_j \right) \lor \dots \lor \left( c_k \land \dots \land c_l \right)$. See \reffigure{Fig:HexFeatureInstantiations} for an example.

We say that a conjunction (or feature instantiation) $C_i$ \textit{generalises} another conjunction $C_j$ if and only if $C_j$ contains all (and possibly more) conditions that $C_i$ contains. In such a case, we have $\lnot C_i \Rightarrow \lnot C_j$, i.e. $C_j$ will for sure not be satisfied if $C_i$ is not satisfied. We say that a disjunction (or feature) $D_i$ \textit{generalises} another disjunction $D_j$ if and only if $D_j$ has at least one non-empty conjunction, and for every conjunction $C_j$ in $D_j$, there exists a conjunction $C_i$ in $D_i$ such that $C_i$ generalises $C_j$. In such a case, we have $\lnot D_i \Rightarrow \lnot D_j$, i.e. $D_j$ will for sure not be satisfied if $D_i$ is not satisfied. These relationships are one aspect that we will take into consideration when determining the order in which to evaluate propositions.

\subsection{Implications Between Propositions Based on Domain Knowledge} \label{Subsec:ImplicationsPropositions}

In addition to the domain-independent knowledge about generalisation relationships discussed above, we can leverage domain knowledge about implications between different propositions to further optimise the evaluation of features. In this subsection, we simply present the domain knowledge that is available. The way in which this is used is discussed afterwards. Note that in this case, ``domain knowledge'' refers to domain knowledge about the game state representations across all of Ludii \cite{Piette_2021_LudiiGameLogicGuide}; this domain knowledge still generalises across many hundreds of games as they are modelled in Ludii.

The left column of \reftable{Table:Implications} lists various propositions that feature instantiations may test for, and the right column lists other propositions that are automatically proven to be true whenever the matching propositions from the left column evaluate to true. For example, whenever a proposition that requires a site $x$ to be empty evaluates to true, we can directly infer that that same position $x$ is not owned by player $1$ (or player $2$ or any other player $p > 0$), that it is not occupied by a piece of type $1$ (or type $2$ or any other piece type $i > 0$), etc. Similarly, the negations of all the propositions in the right column can immediately be \textit{disproven} whenever the matching propositions from the left column evaluate to true. Whenever a proposition $a$ from the left column evaluates to false, it proves and disproves the propositions that correspond to the negation of $a$.

\begin{table}
\centering
\begin{tabular}{@{}ll@{}}
\toprule
Proposition $a$ & Propositions proven by $a$ \\
\midrule
$x$ is empty & $x$ is empty \\
& $x$ is not owned by player $p$ (for any player $p > 0$) \\
& $x$ is not piece $i$ (for any $i > 0$) \\

$x$ is not empty & $x$ is not empty \\

$x$ is owned by player $p$ & $x$ is owned by player $p$ \\
& $x$ is not piece $i$ (for any $i$ not owned by $p$) \\
& $x$ is piece $i$ (if $i$ is the sole type owned by $p$) \\
& $x$ is not empty \\

$x$ is not owned by player $p$ & $x$ is not owned by player $p$ \\
& $x$ is not piece $i$ (for any $i$ owned by $p$) \\

$x$ is piece $i$ & $x$ is piece $i$ \\
& $x$ is not empty \\
& $x$ is not piece $j$ (for any $j \neq i$) \\
& $x$ is owned by player $p$ (where $p$ is the owner of $i$) \\
& $x$ is not owned by player $p$ (for any $p$ that does not own $i$) \\

$x$ is not piece $i$ & $x$ is not piece $i$ \\
& $x$ is not owned by player $p$ (if $i$ is the sole type owned by $p$) \\
\bottomrule
\end{tabular}
\caption{The left column lists propositions $a$ that may be conditions of feature instantiations, where $x$ always refers to a specific site. The right column lists propositions that can be proven by $a$; these can be automatically inferred to be true, without evaluating, whenever $a$ has been evaluated to true. }
\label{Table:Implications}
\end{table}

\subsection{Organising Instantiations and Propositions in Spatial Pattern Networks}

Let $\mathcal{F}(s, a)$ denote a set of relevant feature instantiations for any arbitrary state-action pair $(s, a)$, $\Phi(s, a) = \{ \phi(f) \mid f \in \mathcal{F}(s,a) \}$ a set of features with at least one relevant instantiation, and $\mathcal{P}(s, a) = \bigcup \{ \mathcal{C}(f) \mid f \in \mathcal{F}(s, a) \}$ a set containing all the propositions (or conditions) of all feature instantiations. The naive approach presented in \refalgorithm{Alg:NaiveFeatureEvaluation} would evaluate every proposition in $\mathcal{P}(s,a)$ at least once---and possibly many of them multiple times---to compute a feature vector $\boldsymbol{\phi}(s, a)$. We aim to construct a more efficient algorithm that:
\begin{enumerate}
    \item Never evaluates the same proposition more than once.
    \item Does not evaluate a feature instantiation $f$ if its feature $\phi(f)$ is already known to be active (due to a different instantiation $f'$ with $\phi(f') = \phi(f)$ already having been proven to be active).
    \item Does not evaluate a proposition $c$ for a feature instantiation $f$ if there exists another instantiation $f'$ such that $f$ is generalised by $f'$ (see \refsubsection{Subsec:FeaturesAsDisjunctions}), $f'$ still needs to be evaluated (taking into account the previous point), and $f'$ does not have $c$ as a condition.
    \item Leverages implications between propositions based on domain knowledge (see \refsubsection{Subsec:ImplicationsPropositions}) to keep track of propositions that can be proven or disproven, without actually evaluating them.
\end{enumerate}

To this end, we propose the algorithm described in \refalgorithm{Alg:EvalSPatterNet}, where it is assumed that a total ordering of all propositions in $\mathcal{P}(s,a)$, as well as total ordering of all feature instantiations in $\mathcal{F}(s, a)$, has already been computed. The way in which we compute these orderings is addressed in the next subsection. Furthermore, we assume that a vector $\boldsymbol{\phi}_{init}$ has already been precomputed, such that it contains entries of $1$ for any features that can be guaranteed to be active irrespective of the game state $s$; these are features with at least one instantiation for which all conditions can already be satisfied at instantiation time. We assume that any feature instantiations for such features have also already been removed from $\mathcal{F}(s, a)$, since they are no longer necessary.

\begin{algorithm}
\begin{algorithmic}[1]
\Require Feature instantiations $\mathcal{F}(s, a)$ represented as totally ordered list.
\Require Propositions $\mathcal{P}(s, a)$ represented as totally ordered list.
\State $\boldsymbol{\phi} \gets \boldsymbol{\phi}_{init}$
\State \texttt{active\_props} $\gets \boldsymbol{1}^{\left| \mathcal{P}(s,a) \right|}$ \Comment{Bit array filled with $1$ entries for every proposition.}
\State \texttt{active\_inst} $\gets \boldsymbol{1}^{\left| \mathcal{F}(s,a) \right|}$ \Comment{Bit array filled with $1$ entries for every instantiation.}
\For{each set bit $i$ in \texttt{active\_inst}} \Comment{Evaluate $i^{th}$ instantiation.}
    \For{each proposition $c$ of the $i^{th}$ feature instantiation}
        \If{\texttt{active\_props[$c$]} $= 0$} 
            \Continue \Comment{$c$ has already been evaluated to true, so move on.}
        \EndIf
        \State \texttt{active\_props[$c$] $\gets 0$}
        \If{$s$ violates $c$} \Comment{Condition not satisfied.}
            \State \textsc{Deduce}(\texttt{active\_props}, \texttt{active\_inst}, $\lnot c$)
            \State \textbf{continue} outer loop through \texttt{active\_inst}
        \Else \Comment{$s$ satisfies $c$}
            \State \textsc{Deduce}(\texttt{active\_props}, \texttt{active\_inst}, $c$)
        \EndIf
    \EndFor
    \State $j \gets$ feature index corresponding to the $i^{th}$ feature instantiation
    \State $\boldsymbol{\phi}\left[ j \right] \gets 1$ \Comment{This feature has been proven to be active.}
    \For{each other instantiation $f$ of the same feature with index $j$}
        \State \texttt{active\_inst[$f$]} $\gets 0$ \Comment{Feature already active, so skip other instantiations.}
    \EndFor
\EndFor
\State \Return feature vector $\boldsymbol{\phi}$
\end{algorithmic}
\caption{Proposed algorithm for evaluation of active features.} \label{Alg:EvalSPatterNet}
\end{algorithm}

\begin{algorithm}
\begin{algorithmic}[1]
\Require Bit array \texttt{active\_props} of propositions that we may deactivate.
\Require Bit array \texttt{active\_inst} of instantiations that we may deactivate.
\Require Proposition $c$ that has been evaluated to true.
\For{each proposition $c'$ such that $c \Rightarrow c'$}
    \State \texttt{active\_props[$c'$] $\gets 0$} \Comment{$c'$ already proven to be true.}
\EndFor
\For{each proposition $c'$ such that $c \Rightarrow \lnot c'$}
    \For{each instantiation $f$ such that $f$ requires $c'$}
        \State \texttt{active\_inst[$f$]} $\gets 0$ \Comment{$c'$, and hence $f$, already disproven.}
    \EndFor
\EndFor
\end{algorithmic}
\caption{\textsc{Deduce()} function that makes deductions after evaluating a proposition.} \label{Alg:Deduce}
\end{algorithm}

The basic premise of \refalgorithm{Alg:EvalSPatterNet} is that the \texttt{active\_inst} bit array tracks feature instantiations that should still be (partially) evaluated, and \texttt{active\_props} tracks which propositions have not yet been evaluated. At first, we assume that all instantiations should be evaluated, and do so in the order in which they have been sorted. Evaluating a feature instantiation is done by evaluating all of its propositions. As propositions and feature instantiations are evaluated, others may also already be deactivated (by assigning values of $0$ in their respective bit arrays) and hence pruned. This is done by the \textsc{Deduce()} function described in \refalgorithm{Alg:Deduce}. Note that each of the loops that sets entries in bit arrays to $0$ in this algorithm can be implemented to run at least partially in parallel by implementing them as bitwise \textsc{AndNot} operations.

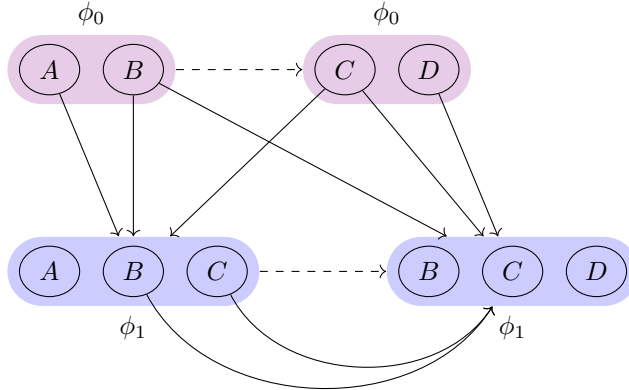
\begin{figure}
\centering
\begin{tikzpicture}[node distance = 2 cm,auto, ]
\tikzset{prop/.style={ellipse,draw,minimum height=0.5cm,minimum width=0.8cm,align=center}}
\tikzset{inst/.style={rounded rectangle, inner xsep=0pt, fill=#1!20}}

\node[prop] (A0) {$A$};
\node[prop,right = .3cm of A0] (B0) {$B$};

\node[prop,right = 2cm of B0] (C0) {$C$};
\node[prop,right = .3cm of C0] (D0) {$D$};

\node[prop,below = 2cm of A0] (A1) {$A$};
\node[prop,right = .3cm of A1] (B1) {$B$};
\node[prop,right = .3cm of B1] (C1) {$C$};

\node[prop,right = 2cm of C1] (B2) {$B$};
\node[prop,right = .3cm of B2] (C2) {$C$};
\node[prop,right = .3cm of C2] (D1) {$D$};

\begin{pgfonlayer}{background}
  \node[fit=(A0)(B0), inst=violet, label=above:{$\phi_0$}] (phi_0_1) {};
  \node[fit=(C0)(D0), inst=violet, label=above:{$\phi_0$}] (phi_0_2) {};
  
  \node[fit=(A1)(C1), inst=blue, label=below:{$\phi_1$}] (phi_1_1) {};
  \node[fit=(B2)(D1), inst=blue, label=below:{$\phi_1$}] (phi_1_2) {};
\end{pgfonlayer}

\path[->,draw]
    (A0) edge (phi_1_1)
    (B0) edge (phi_1_1)
    (B0) edge (phi_1_2)
    (C0) edge (phi_1_1)
    (C0) edge (phi_1_2)
    (D0) edge (phi_1_2)
    ;
    
\path[->,draw,bend right=60]
    (B1) edge (phi_1_2)
    (C1) edge (phi_1_2)
    ;
    
\path[->,draw,dashed]
    (phi_0_1) edge (phi_0_2)
    (phi_1_1) edge (phi_1_2)
    ;
\end{tikzpicture}
\caption{A network representation of features, feature instantiations, and propositions. Circles represent propositions, labelled with letters $A$, $B$, $C$, $D$. Coloured boxes represent feature instantiations---conjunctions of the propositions they surround. Instantiations are labelled and coloured according to the feature they represent. In this example, there are two features---$\phi_0$ and $\phi_1$---with two instantiations each. A solid arrow from a proposition to an instantiation indicates that, if the proposition evaluates to false, the instantiation it points to is also disproven (in addition to the instantiation that the proposition is a part of itself). A dashed arrow from an instantiation $f$ to another instantiation $f'$ indicates that, if $f$ is active, $f'$ no longer needs to be evaluated.}
\label{Fig:ExampleSPatterNet}
\end{figure}

Intuitively, this approach may be viewed as organising the feature instantiations and propositions into a network, with a variety of relationships between propositions and instantiations. See \reffigure{Fig:ExampleSPatterNet} for an example. Feature instantiations are evaluated by traversing the network in a fixed order; in the example figure, going from top-left to top-right to bottom-left to bottom-right. When a proposition evaluates to false, it can immediately deactivate any other complete feature instantiations that include the same proposition. When a feature instantiation evaluates to true, it can immediately deactivate any other complete feature instantiations that represent the same feature. Finally, when a proposition evaluates to true or false, it can prove or disprove other propositions or instantiations according to \reftable{Table:Implications} (not included in the example figure). We refer to such a network as a \textit{Spatial Pattern Network} (SPatterNet). We use a representation based on bit arrays and other primitive tables, rather than an explicit network representation, for improved performance. This is similar to how efficient implementations of Propositional Networks for GDL-based GGP \cite{Cox_2009_Factoring,Schkufza_2008_Propositional} use table-based internal state representations \cite{Draper_2014_Propnets}.

\subsection{Ordering Propositions and Instantiations} \label{Subsec:OrderingPropsInstantiations}

Given a set of relevant feature instantiations $\mathcal{F}(s, a)$, we aim to order feature instantiations and propositions in such a way that, in expectation, the number of evaluations of propositions required by \refalgorithm{Alg:EvalSPatterNet} for any arbitrary state-action pair $(s, a)$ is minimised. The effectiveness of different pruning strategies (based on generalisation relationships between instantiations, different instantiations representing the same feature, and propositions proving or disproving other propositions) can be affected by these orderings in different ways. Hence, we start by considering only generalisation relationships between feature instantiations, and afterwards build on the resulting approach for ordering by incrementally taking into account the other strategies.

\subsubsection{Ordering Based on Generalisation Relationships}

Let $f \in \mathcal{F}(s, a)$ and $f' \in \mathcal{F}(s, a)$ denote two different feature instantiations such that $f'$ is generalised by $f$ (i.e., $f'$ is a conjunction of all propositions of $f$, plus at least one more). In such a case, $f$ should always be fully evaluated before $f'$. This creates a trivial partial ordering of instantiations, where more general instantiations are always evaluated before more specific instantiations. An arbitrary ordering can be used between any pair of instantiations for which there exists no generalisation relationship.

\subsubsection{Ordering Instantiations Based on Shared Features} \label{Subsubsec:OrderingInstantiations}

Let $f \in \mathcal{F}(s, a)$ and $f' \in \mathcal{F}(s, a)$ denote two different feature instantiations such that both instantiations represent the same feature, i.e., $\phi(f) = \phi(f')$. When either one of these instantiations evaluates to true, evaluation of the other can be skipped entirely, because the feature is already known to be active. Intuitively, this suggests that ``shorter'' instantiations (conjunctions of fewer propositions) should be ordered and evaluated before ``longer'' instantiations of the same feature. This intuition can easily be combined with the rule described above that more general instantiations should be evaluated before more specific instantiations. There are no conflicts between these two ideas, since a more general instantiation is also always shorter than a more specific instantiation. However, this intuition does not account for the observation that when a single proposition evaluates to false, this can immediately disprove any complete instantiation that it is a part of---including potentially many large or highly-generalised instantiations that would otherwise be deprioritised. This insight suggests that it may be useful to order individual propositions, rather than complete instantiations (conjunctions), and to prioritise propositions that appear in short conjunctions as well as propositions that appear in many conjunctions. These two criteria may conflict.

A similar conflict between two such heuristics commonly appears in boolean satisfiability (SAT) problems, in which the goal is to determine for any given propositional formula in conjunctive normal form, whether there exists an assignment of truth values to all variables such that the formula is true. Typical approaches for SAT problems use a backtracking search through the space of all possible assignments of truth values to variables \cite{Davis_1960_Computing,Davis_1962_Machine}. In such a backtracking search, it is typically desirable to prioritise assigning truth values to the \textit{most constrained} variables, because those are often the most likely to lead to unsatisfiable clauses---which allows for early backtracking. A popular set of heuristics are the \textit{Maximum Occurrences in clauses of Minimum Size} (MOMS) heuristics \cite{Pretolani_1993_Efficiency}, which prioritise variables that occur frequently, as well as variables that occur in small clauses (disjunctions). We take inspiration from these heuristics in our approach.

We propose an approach that orders propositions by iteratively picking propositions, sorting them in the order in which they are picked, keeping the following principles in mind:
\begin{enumerate}
    \item If a disjunction (feature; disjunction of conjunctions of propositions) is not generalised by any other disjunctions, it must be evaluated.
    \item If a disjunction must be evaluated, at least one of its conjunctions must be evaluated (repeated until either one conjunction evaluates to true, or all conjunctions evaluate to false).
    \item If a conjunction must be evaluated, at least one of its propositions must be evaluated (repeated until either all its propositions evaluate to true, or one proposition evaluates to false).
\end{enumerate}
Based on the first principle, we start out by splitting the set of all disjunctions (features) in two sets: a set $\mathcal{U}_0$ of \textit{ungeneralised} disjunctions, and a set $\mathcal{G}_0$ of \textit{generalised} disjunctions. For every ungeneralised disjunction $D \in \mathcal{U}_0$, there exists no other disjunction $D' \in \mathcal{U}_0 \cup \mathcal{G}_0$ such that $D'$ generalises $D$. For every generalised disjunction $D \in \mathcal{G}_0$, there exists at least one ungeneralised disjunction $D' \in \mathcal{U}_0$ such that $D'$ generalises $D$. For ease of reference, we list these and several other definitions of symbols in \reftable{Table:SymbolsDefinitions}.

The second and third principles suggest that we should start out by picking a set of propositions such that at least one proposition from at least one conjunction from every ungeneralised disjunction $D \in \mathcal{U}_0$ is picked.\footnote{Picking a \textit{minimal} set of such propositions would be equivalent to the hitting set problem, or set cover problem, which is NP-complete \cite{Karp_1972_Reducibility}. A common and simple heuristic in greedy algorithms to generate approximate solutions \cite{Chvatal_1979_Greedy} would translate to our setting as a heuristic that would prioritise propositions that appear in a maximal number of disjunctions.} We partition the ungeneralised disjunctions $\mathcal{U}_0$ into subsets $\mathcal{U}_0^1, \mathcal{U}_0^2, \dots$, such that $\mathcal{U}_0^i$ is the subset of all disjunctions that contain $i$ conjunctions. As a special case, we start by immediately picking all propositions for conjunctions of length $1$ in disjunctions of length $1$ (i.e., disjunctions in $\mathcal{U}_0^1$). Afterwards, we loop through all $\mathcal{U}_0^i$ in increasing order of $i$, every time looping through the remaining uncovered disjunctions (in an arbitrary order) and picking a single proposition to cover that disjunction. Let $D \in \mathcal{U}_0^i$ denote such a disjunction that we need to pick a proposition from. Let $C(D)$ denote its set of conjunctions, $C \in C(D)$ one of the conjunctions, and $c \in C$ one of the propositions of such a conjunction. With some abuse of notation, we use $c \in C(D)$---with a lowercase $c$---to denote a proposition from any one of the conjunctions in $C(D)$. Let $P$ denote the set of propositions that have already been picked. Let $\vert C \vert$ denote the length (i.e., number of propositions) of a conjunction $C$. Let $\Omega$ denote a set of all conjunctions $C'$ such that $C'$ is a part of some disjunction $D' \in \mathcal{U}_0^i$ where $D'$ is not covered by any of the propositions picked so far in $P$. Let $\Omega_c = \{ C \in \Omega \mid c \in C \}$ denote the subset of conjunctions in $\Omega$ that contain $c$ as one of their propositions. Then, we pick the proposition $c^* \in C(D)$ given by \refequation{Eq:JeroslowWang1}, which is based on the Jeroslow-Wang heuristic for SAT \cite{Jeroslow_1990_Solving}:
\begin{equation} \label{Eq:JeroslowWang1}
    c^* = \argmax_{c \in C(D)} \sum_{C' \in \Omega_c} 2^{- \vert C' \vert}
\end{equation}

Intuitively, this heuristic simply increases the score of a proposition for every conjunction that includes that proposition, while ignoring conjunctions from disjunctions that are already covered by any of the previously-picked propositions $P$. The exponent in the $2^{-\vert C' \vert}$ term favours propositions that appear in short conjunctions $C'$ over propositions that appear in long conjunctions $C'$. Ties in the heuristic score are broken by computing a tie-breaker score in the same way over disjunctions in the set of generalised disjunctions $\mathcal{G}_0$ (instead of $\mathcal{U}_0$ in the definition of $\Omega$), and any further ties are broken randomly.

After going through the procedure described above once, we end up with an initial list of propositions that cover at least one conjunction of every ungeneralised (and hence also every generalised) disjunction. Which propositions should be prioritised after this initial ordering can typically be very different depending on, for each of the propositions that have already been picked, whether they evaluate to true or false for any given state-action pair $(s, a)$. This is because, whenever a proposition evaluates to false, any conjunction it is a part of becomes irrelevant. In contrast, whenever a proposition evaluates to true, any conjunction it is a part of becomes easier to prove, and hence arguably \textit{more} important. This suggests that it may be desirable to construct a (binary) tree, rather than a single list, such that the order in which propositions are evaluated can be conditioned on the values that earlier propositions evaluate to. We explore this topic in more detail in \refsubsection{Subsection:SPatterNetDiscussion}, but in the remainder of this paper assume that we prefer only a single ordered list. 

We make the simple assumption that all propositions picked so far would evaluate to true (even if that is typically not possible due to mutually exclusive propositions), and continue consecutive rounds of picking propositions based on that assumption. This is implemented by removing picked propositions from all conjunctions that contain them, deleting conjunctions that have a length of $0$ after such removals from their disjunctions, and removing disjunctions that no longer have any conjunctions. After these changes, we re-compute which disjunctions are generalised or ungeneralised. At this stage, disjunctions are not split up in only two sets $\mathcal{U}_0$ and $\mathcal{G}_0$, but potentially many sets $\mathcal{U}_i$ and $\mathcal{G}_i$ for $i \geq 0$. A set with subscript $i$ contains any disjunctions from which $i$ full conjunctions have already been removed due to being fully covered. Where we previously described that we would loop through $\mathcal{U}_0$ to pick new propositions, we now loop through $\mathcal{U}_i$ for the minimal $i$ such that $\mathcal{U}_i \neq \varnothing$. When applying the heuristic score of \refequation{Eq:JeroslowWang1}, all $\mathcal{U}_i$ in increasing order of $i$, followed by all $\mathcal{G}_i$, are used as tie-breakers in the definition of $\Omega$. Intuitively, we prioritise disjunctions with fewer fully covered conjunctions over disjunctions with more fully covered conjunctions, and place more importance on ungeneralised disjunctions than generalised disjunctions. This complete process is repeated until all propositions have been picked. \refalgorithm{Alg:OrderPropositions} provides pseudocode of the entire algorithm to order propositions.

\begin{table}
\centering
\begin{tabular}{@{}lp{0.8\linewidth}@{}}
\toprule
Symbol(s) & Definition \\
\midrule
$f$, $f'$ & Feature instantiations. \\
$\phi(f)$ & Feature of which $f$ is an instantiation. \\
$D$, $D'$ & Disjunction (of conjunctions of propositions). \\
$\vert D \vert$ & Cardinality (number of conjunctions) of a disjunction $D$. \\
$C(D)$ & Set of conjunctions in the disjunction $D$. \\
$C$, $C'$ & Conjunction of propositions. \\
$\vert C \vert$ & Cardinality (number of propositions) of a conjunction $C$. \\
$c \in C$ & One of the propositions of a conjunction $C$. \\
$c \in C(D)$ & One of the propositions of any one of the conjunctions in $C(D)$. \\
$\mathcal{U}_0$ & Set of ungeneralised disjunctions from which no conjunctions have been fully removed yet. \\
$\mathcal{G}_0$ & Set of generalised disjunctions from which no conjunctions have been fully removed yet. \\
$\mathcal{U}_i$ & Set of ungeneralised disjunctions from which $i$ conjunctions have been fully removed already. \\
$\mathcal{G}_i$ & Set of generalised disjunctions from which $i$ conjunctions have been fully removed already. \\
$\mathcal{U}_j^i$ & Subset of $\mathcal{U}_j$ containing only disjunctions that contain $i$ conjunctions. \\
$P$ & Set of propositions that have already been picked. \\
$\Omega$ & Set of all conjunctions $C'$ such that $C'$ is a part of some disjunction $D'$ that, in turn, is part of a set $\mathcal{U}_i^j$ with the minimum $i$ such that $\mathcal{U}_i \neq \varnothing$, such that $D'$ is not covered by any propositions picked in $P$. \\
$\Omega_c$ & Subset of conjunctions $C$ in $\Omega$ that contain $c$ as one of their propositions. \\
\bottomrule
\end{tabular}
\caption{Listing of symbols and definitions used throughout \refsubsubsection{Subsubsec:OrderingInstantiations}.}
\label{Table:SymbolsDefinitions}
\end{table}

\begin{algorithm}
\begin{algorithmic}[1]
\Require Initial set of ungeneralised disjunctions $\mathcal{U}_0$
\Require Initial set of generalised disjunctions $\mathcal{G}_0$
\Statex
\Function{OrderPropositions}{}
    \State $\mathcal{U}_i \gets \varnothing$ for all $i > 0$
    \State $\mathcal{G}_i \gets \varnothing$ for all $i > 0$
    \State \texttt{ordered\_propositions $\gets$ []} \Comment{Init as empty list. $P$ is set representation of list.}
    \While{at least one $\mathcal{U}_i \neq \varnothing$}
        \State $i \gets \min_i$ such that $\mathcal{U}_i \neq \varnothing$
        \State Partition $\mathcal{U}_i$ into subsets $\mathcal{U}_i^j$ of disjunctions with $j \geq 1$ conjunctions
        \For{$j = 1, 2, 3, \dots$}
            \For{each disjunction $D \in \mathcal{U}_i^j$}
                \If{$D$ not already covered by $P$}
                    \State $c \gets$ \Call{PickProposition}{$D$, $i$, $j$}
                    \State Append $c$ to \texttt{ordered\_propositions}
                \EndIf
            \EndFor
        \EndFor
        \For{all $i'$}
            \For{all disjunctions $D$ in $\mathcal{U}_{i'}$ or $\mathcal{G}_{i'}$}
                \For{all conjunctions $C \in C(D)$}
                    \State Remove all picked propositions $P$ from $C$
                    \If{$\vert C \vert = 0$}
                        \State Remove $C$ from $C(D)$
                    \EndIf
                \EndFor
                \If{$\vert D \vert = 0$}
                    \State Remove $D$ from $\mathcal{U}_{i'}$ or $\mathcal{G}_{i'}$
                \EndIf
            \EndFor
        \EndFor
        \State Re-compute all $\mathcal{U}_{i'}$ and $\mathcal{G}_{i'}$ \Comment{See \reftable{Table:SymbolsDefinitions}.}
    \EndWhile
    \State \Return \texttt{ordered\_propositions}
\EndFunction

\Statex

\Function{PickProposition}{$D$, $i$, $j$}
    \State Compute set of conjunctions $\Omega$ from $\mathcal{U}_i^j$. \Comment{See \reftable{Table:SymbolsDefinitions}.}
    \State Pick $c \in C(D)$ that maximises heuristic. \Comment \refequation{Eq:JeroslowWang1} or \refequation{Eq:JeroslowWang2}
    \State Break ties using $\mathcal{U}_i^{j+1}, \mathcal{U}_i^{j+2}, \dots, \mathcal{U}_{i+1}^1, \mathcal{U}_{i+1}^2, \dots, \mathcal{G}_0^1, \mathcal{G}_0^2, \dots, \mathcal{G}_1^1, \dots$ 
    \State \Return $c$
\EndFunction
\end{algorithmic}
\caption{Order propositions of feature set (see \reftable{Table:SymbolsDefinitions} for definitions of symbols).} \label{Alg:OrderPropositions}
\end{algorithm}

After obtaining a total ordering of the propositions from this process, we order the feature instantiations such that no instantiation $f$ is ordered before another instantiation $f'$ if $f$ requires a proposition that is ordered after all propositions required by $f'$. In other words, the last-ordered proposition of an instantiation determines the ordering of that instantiation. This lets us use \refalgorithm{Alg:EvalSPatterNet}---which is based on looping through an ordered list of instantiations---while still ensuring that the propositions are evaluated in approximately the order that they were sorted in.

\subsubsection{Ordering Based on Implications Between Propositions}

As a final step, we consider taking into account the implication relationships between various propositions, as described in \refsubsection{Subsec:ImplicationsPropositions}, when ordering feature instantiations. Whenever a proposition $c$ is evaluated, it will evaluate to either \texttt{true} or \texttt{false}, and in either case there can be several other propositions $c'$ that will be proven ($c \Rightarrow c'$ or $\lnot c \Rightarrow c'$) or disproven ($c \Rightarrow \lnot c'$ or $\lnot c \Rightarrow \lnot c'$). We avoid making any additional assumptions concerning the marginal likelihoods for any given proposition to be true or false (which would require game-specific domain knowledge), instead simply assuming that any proposition is equally likely to evaluate to either true or false. This means that, when a procedure like the one described above picks a proposition $c$ to be inserted in the ordered list, we expect there to be a probability of $0.5$ to automatically gain information on any propositions that are proven or disproven by $c$ being true, and a probability of $0.5$ to automatically gain information on any propositions that are proven or disproven by $c$ being false.

We incorporate these ideas in the ordering procedure by adapting the heuristic of \refequation{Eq:JeroslowWang1} by adding $50\%$ of the heuristic score of any proposition $c'$ that may be proven or disproven by the evaluation of another proposition $c$, to the heuristic score of $c$. More formally, let $\top(c)$ denote the set of propositions that are either proven or disproven when $c$ evaluates to true, and let $\bot(c)$ denote the set of propositions that are either proven or disproven when $c$ evaluates to false. Then we replace the original heuristic of \refequation{Eq:JeroslowWang1} by the updated version in \refequation{Eq:JeroslowWang2}:
\begin{equation} \label{Eq:JeroslowWang2}
    c^* = \argmax_{c \in C(D)} \sum_{C' \in \Omega_c} 2^{-\vert C' \vert} + \frac{1}{2} \sum_{c' \in \top(c)} \left[ \sum_{C' \in \Omega_{c'}} 2^{-\vert C' \vert} \right] + \frac{1}{2} \sum_{c' \in \bot(c)} \left[ \sum_{C' \in \Omega_{c'}} 2^{-\vert C' \vert} \right]
\end{equation}

\subsection{Discussion} \label{Subsection:SPatterNetDiscussion}

In \refsubsubsection{Subsubsec:OrderingInstantiations}, we remarked that in theory, it may be preferably to order propositions in a binary tree rather than a list, such that the order in which later propositions are evaluated can be conditional on the truth values that earlier propositions evaluate to. If an earlier proposition evaluates to false, any conjunction it is a part of is immediately disproven and therefore becomes irrelevant, which should in turn deprioritise any other propositions in the same conjunctions. Conversely, if an earlier proposition evaluates to true, any conjunctions it is a part of are closer to being proven, which should raise the priority level of other propositions in those conjunctions.

Intuitively, this also happens when MOMS heuristics are used to select variables to focus on next in a backtracking search for SAT problems. During a backtracking search, truth values are assigned to selected variables, and the heuristics are used ``online'' (during the search) to select a single next variable for that specific part of the search tree---which corresponds to a specific assignment of truth values to previously-selected variables.

The core reason not to consider such an approach further in this paper is that it would have excessive memory requirements. In contrast to SAT problems where the selection of variables happens online during a backtracking search, we deal with an offline preprocessing step of which the result (ordering of propositions) remains stored in memory for future use (to speed up online feature evaluations during gameplay). Let $K \geq 0$ denote the number of propositions for a full feature set. A single ordering of these propositions in a list only contains $K$ entries, but an ordering in a binary tree---allowing for later propositions to be conditioned on truth values of earlier propositions---would have $2^K - 1$ entries. 

\section{Experiments} \label{Sec:Experiments}

In this section, we describe several experiments used to evaluate the performance of the proposed approach for evaluating active features using SPatterNets. In our experiments, we consider the following four approaches for evaluating active features:
\begin{enumerate}
    \item \textbf{Naive}: a straightforward, naive approach for evaluating active features, as described by \refalgorithm{Alg:NaiveFeatureEvaluation}. This represents the simplest baseline.
    \item \textbf{Tree}: an approach that organises feature instantiations in a tree, such that more specific instantiations are located below more general instantiations. When an instantiation evaluates to false, any (more specific) instantiations below it are skipped. This is similar to the approaches used, for example, by \citet{Levinson_1991_Adaptive} and \citet{Muller_1991_Pattern,Muller_1995_PhD}, although likely less efficient due to our requirements for a general system that is not specific to a single game or board geometry. It also bears resemblance to the DAG-based approach described by \citet{Buro_1999_Features}. More precisely, this approach incrementally builds up a tree by inserting feature instantiations one by one, always inserting any given new instance as a child of whichever potential parent (generaliser) already has the deepest current position in the tree. This represents a more advanced baseline.
    \item \textbf{SPatterNet}: the main approach proposed in this paper, which organises propositions and instantiations as described in \refsubsection{Subsec:OrderingPropsInstantiations}, and evaluates active features using \refalgorithm{Alg:EvalSPatterNet}.
    \item \textbf{SPatterNet (JIT)}: A variant of the SPatterNet approach, in which features are instantiated in a just-in-time (JIT) manner when needed, rather than instantiating all features for all possible move keys (as described in \refsubsection{Subsec:InstantiatingFeatures}) in advance. This can significantly reduce the initialisation time required before a game can start being played. Not generating instantiations for moves that are never legal in practical play can also reduce memory usage and the sizes of hash maps, which in turn may be a benefit in terms of processing speed. Because in some cases it will be necessary to instantiate features at runtime (while a game is being played), there may also be a reduction in processing speed.
\end{enumerate}
All experiments are run using a set of 33 highly varied, distinct games, which we provide additional details on in \refappendix{Appendix:Games}. All experiments are run using the Ludii general game system. The source code of Ludii, as well as all the code for spatial features, is available from \url{https://github.com/Ludeme/Ludii}. Version 1.3.1 of Ludii was used for the experiments described in this paper.

The primary measure of performance we focus on is the number of playouts per second we can run. Playouts are run from the initial game state until terminal game states are reached, by sampling actions uniformly at random (unless stated otherwise), while computing the active features for every legal action in every state. For every game, every feature set, and each of the evaluated implementations, we run a separate process with access to two cores of a 2.6GHz Intel Xeon E5-2690 v3 CPU, and 4096MB allocated to the Java Virtual Machine (JVM). Playouts are run sequentially, but access to a second core may be used, for instance, by Java's garbage collector. Every process uses 60 seconds of warming up time for the JVM (during which the JIT variant of SPatterNet can also instantiate features), after which the number of playouts per second is measured over a duration of 600 seconds.

\subsection{Atomic Feature Sets}

For cases where it is infeasible or undesirable to exhaustively generate all patterns or features up to a given size (as commonly done in Go programs), several approaches have been proposed that start with a smaller set of \textit{atomic} features, and gradually grow this by generating more complex features that are composed of two or more other (atomic, or previously-generated composite) features \cite{Buro_1999_Features,Sturtevant_2007_Hearts,Skowronski_2009_Automated,Soemers_2019_Biasing}. For our first experiment, we evaluate the performance on such atomic feature sets.

Using the spatial feature format as described in \refsubsection{Subsec:SpatialFeatureFormat}, we define atomic features to be features that have exactly one requirement for the state (i.e., one walk with one element such as ``must be empty'' or ``must not be enemy''), in addition to any specifiers related to actions. We consider several different atomic feature sets, described as Atomic-$M$-$N$ for different integer values $M \geq 1$ and $N \geq M$. In a set labelled Atomic-$M$-$N$, we generate all such features for any walk restricted to a length of up to $M$ steps, allowing walks of up to $N \geq M$ steps for ``straight'' walks (which only have steps with rotation values $\rho = 0$, i.e. no turns). Furthermore, the following rules apply for generating atomic feature sets:
\begin{itemize}
    \item Let $K$ denote the largest number of orthogonal connections (including artificial off-board connections) for any site in a given game. In the vast majority of games, this number is the same for \textit{all} sites (due to the prevalence of boards based on regular tilings). In all generated walks, we only consider rotation values that are fractions of $K$ (i.e. $\rho = \frac{1}{K}$, $\rho = \frac{2}{K}$, $\dots$, $\rho = 1$).
    \item Every feature must have at least a \texttt{from} or a \texttt{to} position specified (or both). In games for which the rules can never generate any moves with a distinct \texttt{from} position, no features with \texttt{from} or \texttt{last\_from} specifiers are generated.
    \item For any generated feature, either the \texttt{from} or the \texttt{to} position (or both) must have a walk of length $0$, causing it to overlap with the anchor.
    \item In $2$-player games with only one piece type per player (Hex, Go, Tic-Tac-Toe, etc.), no features are generated with elements testing for piece types---only testing for friend or enemy is sufficient in these games.
\end{itemize}

\subsubsection{Results}

The raw number of playouts per second varies significantly between different games, which means that the effect of evaluating features on those playout rates cannot be directly compared or averaged across games. To present aggregate results, we take the number of playouts per second for the smallest feature set---Atomic-$1$-$1$---with the simplest feature evaluation approach---Naive---as a baseline $b$ per game. For every other pair of feature set and feature evaluation approach, for every game, we compute the \textit{slowdown} by dividing the per-game baseline $b$ by that pair's raw number of playouts per second. Larger values indicate a greater slowdown (relative to Naive with the smallest Atomic-$1$-$1$ feature set) as a result of using a specific combination of a feature set and feature evaluation approach. Values below $1.0$ are speedups.

\begin{figure}
\centering
\includegraphics[width=\linewidth]{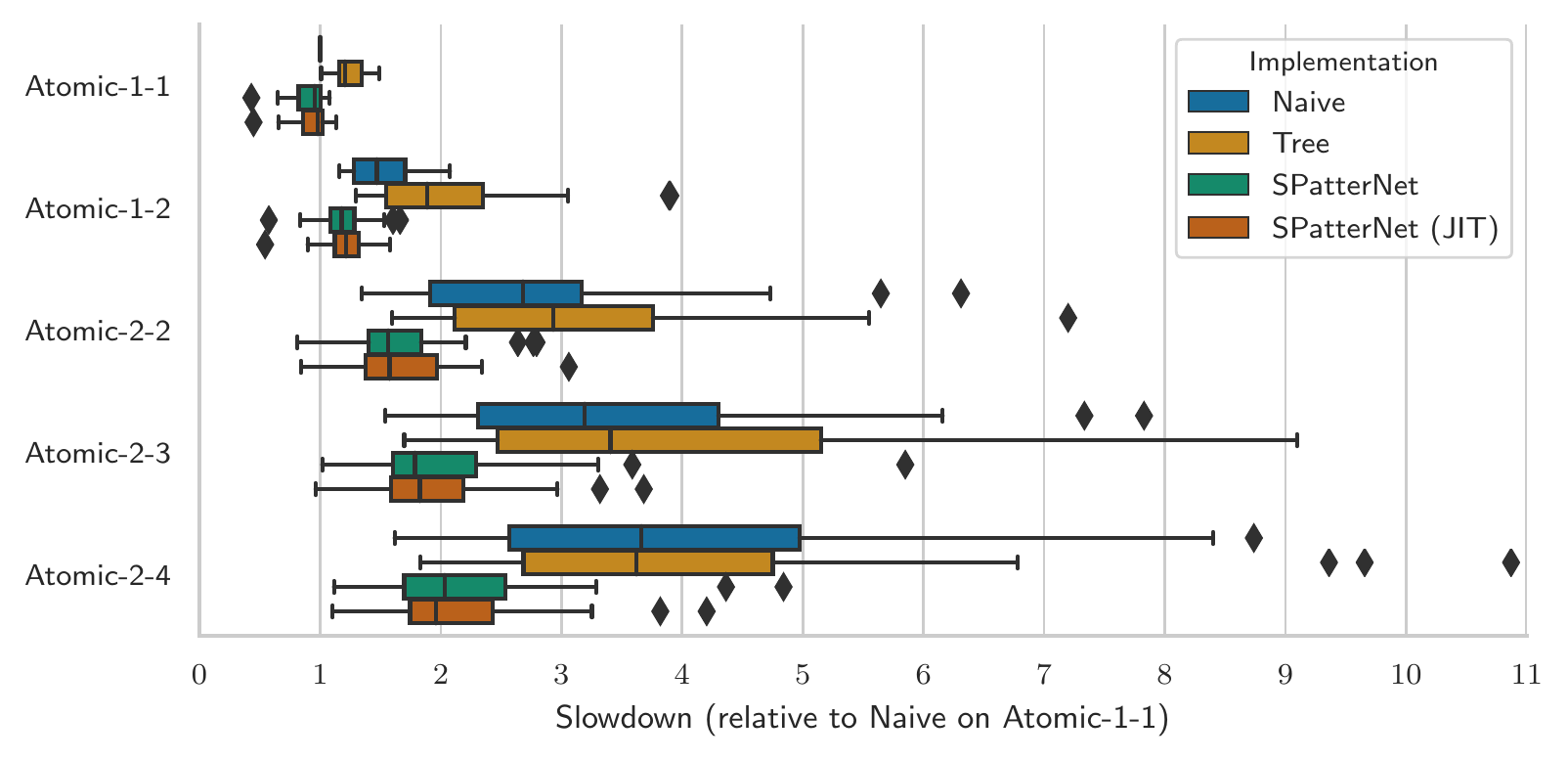}
\caption{Boxplots summarising slowdowns of several feature evaluation implementations on several different sets of atomic features. Every boxplot summarises $33$ data points, from $33$ different games. Lower slowdown values are better (with values $< 1$ being speedups). The size (number of features) per feature set increases as we go down along the $y$-axis.}
\label{Fig:BenchmarkAtomicFeatureSets}
\end{figure}

The boxplots in \reffigure{Fig:BenchmarkAtomicFeatureSets} summarise these results, with every boxplot---one for every possible pair of an atomic feature set and a feature evaluation approach---representing $33$ data points for $33$ different games. Every atomic feature set is a strict subset of all the ones below it, which means that feature sets that appear lower in the figure will always require at least as much work to evaluate active features as those that are higher in the figure---and therefore have greater slowdowns. In every boxplot, the vertical bar represents the median result (over 33 games), the coloured box covers the interquartile range, and the whiskers cover any remaining data up to a distance of $1.5$ times the interquartile range below or above the $25^{th}$ or $75^{th}$ percentile, respectively. Data points outside of the whiskers are visualised individually as diamonds.

\begin{table}
\centering
\begin{tabular}{@{}lrrrr@{}}
\toprule
& \multicolumn{4}{c}{Frequency per Rank} \\
\cmidrule(lr){2-5}
& Rank $1$ & Rank $2$ & Rank $3$ & Rank $4$ \\
\midrule
\textbf{Atomic-$1$-$1$} & & & & \\
\hspace{3mm}Naive & 8 & 6 & 19 & 0 \\
\hspace{3mm}Tree & 0 & 0 & 0 & 33 \\
\hspace{3mm}SPatterNet & 21 & 6 & 6 & 0 \\
\hspace{3mm}SPatterNet (JIT) & 4 & 21 & 8 & 0 \\
\textbf{Atomic-$1$-$2$} & & & & \\
\hspace{3mm}Naive & 1 & 3 & 28 & 1 \\
\hspace{3mm}Tree & 0 & 0 & 1 & 32 \\
\hspace{3mm}SPatterNet & 19 & 12 & 2 & 0 \\
\hspace{3mm}SPatterNet (JIT) & 13 & 18 & 2 & 0 \\
\textbf{Atomic-$2$-$2$} & & & & \\
\hspace{3mm}Naive & 0 & 3 & 23 & 7 \\
\hspace{3mm}Tree & 0 & 0 & 7 & 26 \\
\hspace{3mm}SPatterNet & 19 & 12 & 2 & 0 \\
\hspace{3mm}SPatterNet (JIT) & 14 & 18 & 1 & 0 \\
\textbf{Atomic-$2$-$3$} & & & & \\
\hspace{3mm}Naive & 0 & 0 & 27 & 6 \\
\hspace{3mm}Tree & 0 & 0 & 6 & 27 \\
\hspace{3mm}SPatterNet & 18 & 15 & 0 & 0 \\
\hspace{3mm}SPatterNet (JIT) & 15 & 18 & 0 & 0 \\
\textbf{Atomic-$2$-$4$} & & & & \\
\hspace{3mm}Naive & 0 & 0 & 24 & 9 \\
\hspace{3mm}Tree & 0 & 0 & 9 & 24 \\
\hspace{3mm}SPatterNet & 13 & 20 & 0 & 0 \\
\hspace{3mm}SPatterNet (JIT) & 20 & 13 & 0 & 0 \\
\bottomrule
\end{tabular}
\caption{For each of the $33$ games, and for every feature set, we rank the four feature evaluation approaches based on their performance. This table lists, for every feature set, how often each rank was obtained by every feature evaluation approach, where a rank of 1 means best performance and a rank of 4 means worst performance in a given game.}
\label{Table:AtomicFeaturesRankCounts}
\end{table}

For each of the $33$ games, and for each of the atomic feature sets, we rank the four feature evaluation approaches based on their performance for that specific game and atomic feature set. This means that, for each atomic feature set, we distribute ranks ranging from 1 (best) to 4 (worst) among the four evaluated approaches $33$ times. For each approach, on each feature set, the frequency per rank (number of times---out of $33$---that a certain rank was assigned) is listed in \reftable{Table:AtomicFeaturesRankCounts}.

\subsubsection{Discussion} \label{Subsubsec:DiscussionAtomic}

The boxplots of \reffigure{Fig:BenchmarkAtomicFeatureSets} consistently show, across all considered feature sets, that both variants of SPatterNet are the two most efficient implementations in terms of aggregate results such as the median slowdowns (or speedups), the most extreme outliers, etc. The advantage of the SPatterNet approaches over the other two approaches arguably becomes more pronounced as the size of the feature sets increases.

Interestingly, the Tree approach appears to be outperformed by the Naive approach, in particular on the smallest feature sets, even though the Tree approach is meant to be an optimisation based on generalisation relationships between feature instantiations. We remark that in atomic feature sets, there are generally relatively few features or feature instantiations that actually generalise others; by design the atomic features are meant to be mostly independent features with little overlap. This is particularly true for the smallest sets; in the larger atomic feature sets, which allow for longer walks, it is more likely that multiple different walks (including some turns) will have overlapping destinations. Therefore, in the smaller atomic feature sets, the Tree approach is slower than the Naive approach, because its generalisation-based optimisation is ineffective, but it still has a more complex implementation with additional overhead. In the larger atomic feature sets, it appears that this optimisation becomes closer to worthwhile. The SPatterNet approaches include similar generalisation-based optimisations (which are ineffective in small atomic feature sets), but also other optimisations that can already pay off for smaller feature sets.

The per-game comparisons summarised by \reftable{Table:AtomicFeaturesRankCounts} lead to similar conclusions. The two SPatterNet approaches already share the top 2 ranks in the majority of games even for the smallest atomic feature set, and in the largest feature sets they entirely dominate the top 2 ranks. Furthermore, neither of them has the worst performance (rank 4) in any case. Between the two SPatterNet approaches, the results in the table suggest that the standard variant may perform better for smaller feature sets, whereas the JIT variant may perform better for larger feature sets. However, considering the results depicted by \reffigure{Fig:BenchmarkAtomicFeatureSets}, such differences are unlikely to be large in magnitude.

\subsection{Trained Feature Sets} \label{Subsec:TrainedFeaturesExperiment}

In our next experiment, we move on to benchmarking larger, trained feature sets which include non-atomic composite features that have been discovered to be potentially useful in a training process. In each game, we start out with its Atomic-$2$-$4$ feature set---the largest of the atomic feature sets evaluated in the previous experiment. We interleave \cite{Utgoff_1998_Constructive,Soemers_2019_Biasing} policy training and feature discovery (the addition of new features) in a self-play training setup between MCTS agents that are guided by the trained policy \cite{Anthony_2017_ExIt,Silver_2017_AlphaGoZero}, where the policy uses the features as input. For every game, we run such a training process for up to $200$ episodes, or up to $24$ hours. After every self-play episode, we attempt to add one proactive and one reactive feature to the feature set, by combining two existing feature instantiations into a new composite feature \cite{Soemers_2019_Biasing}. This means that, in comparison to the Atomic-$2$-$4$ sets, each of our feature sets grows by up to $400$ features (of which $200$ proactive and $200$ reactive). More detailed information on this training setup is provided in \refappendix{Appendix:DetailsTraining}.

After such a training run for a game, we do not only have a new, larger feature set, but also a trained policy $\pi$ which can, for any given state-action pair $(s, a)$, provide a probability $0 \leq \pi(s, a) \leq 1$ with which the action $a$ should be selected in state $s$. For each of the large, trained feature sets, we benchmark the number of playouts per second for every game two times; once by still sampling actions uniformly (evaluating active features solely for the sake of benchmarking performance), and once by sampling actions according to the trained policy $\pi$.

\subsubsection{Results}

\begin{figure}
\centering
\includegraphics[width=\linewidth]{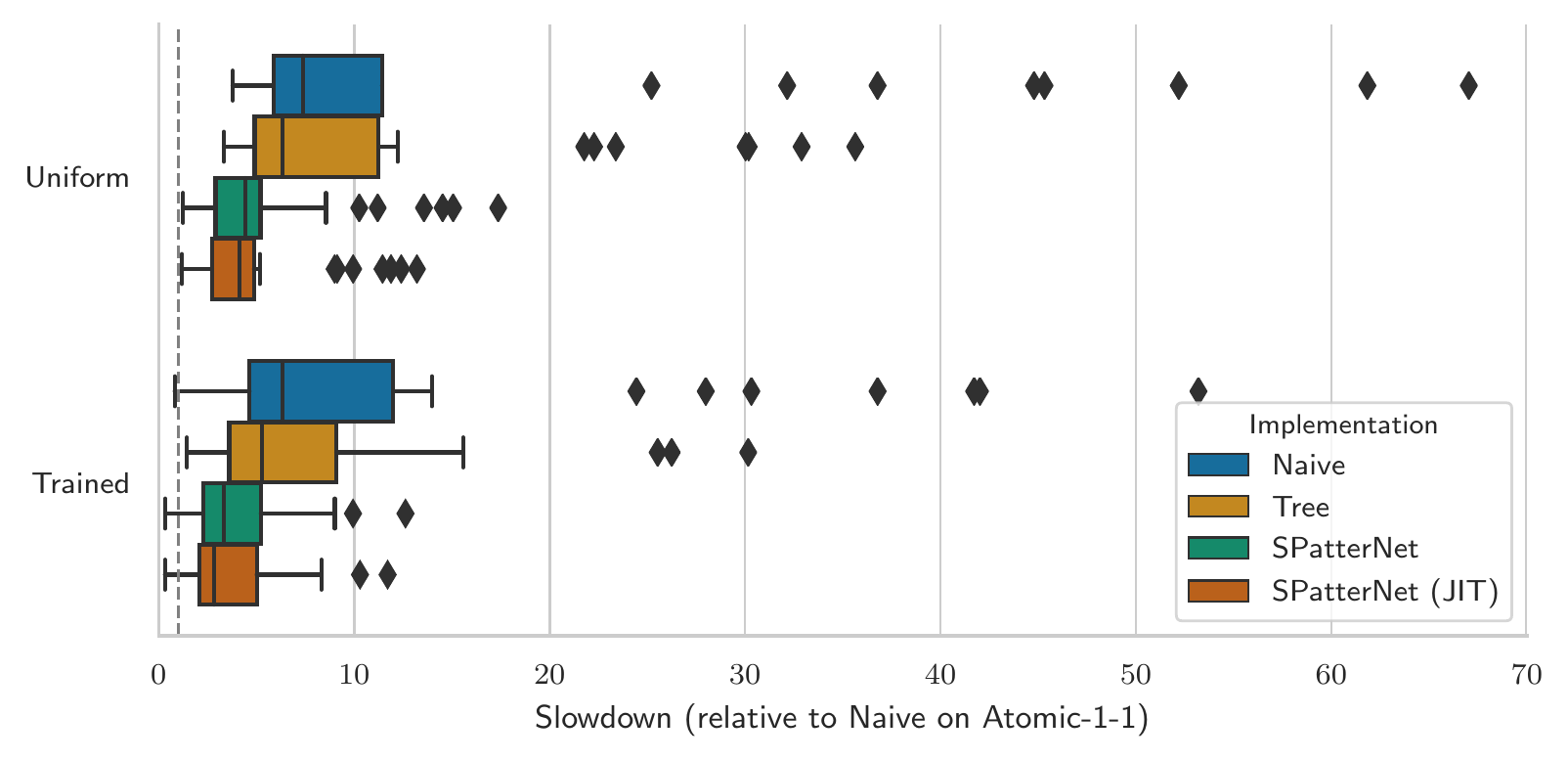}
\caption{Boxplots summarising slowdowns of several feature evaluation implementations on large, trained feature sets; once where actions are sampled uniformly at random, and once where actions are sampled according to a trained policy. Every boxplot summarises $33$ data points, from $33$ different games. Lower slowdown values are better (with values $< 1$ being speedups---the dashed vertical line indicates this point). }
\label{Fig:BenchmarkTrainedFeatureSets}
\end{figure}

\reffigure{Fig:BenchmarkTrainedFeatureSets} depicts boxplots that summarise the slowdowns resulting from the trained feature sets. Note that we still use the same baseline---the playouts per second from the Naive approach on the smallest atomic feature set---as in \reffigure{Fig:BenchmarkAtomicFeatureSets} to compute the relative slowdowns. \reftable{Table:TrainedFeaturesRankCounts} lists for each approach how often, out of $33$ games, it achieved each of the ranks ranging from 1 (best) to 4 (worst).

\begin{table}
\centering
\begin{tabular}{@{}lrrrr@{}}
\toprule
& \multicolumn{4}{c}{Frequency per Rank} \\
\cmidrule(lr){2-5}
& Rank $1$ & Rank $2$ & Rank $3$ & Rank $4$ \\
\midrule
\textbf{Uniform Action Sampling} & & & & \\
\hspace{3mm}Naive & 0 & 0 & 5 & 28 \\
\hspace{3mm}Tree & 0 & 0 & 28 & 5 \\
\hspace{3mm}SPatterNet & 3 & 30 & 0 & 0 \\
\hspace{3mm}SPatterNet (JIT) & 30 & 3 & 0 & 0 \\
\textbf{Trained Policy Action Sampling} & & & & \\
\hspace{3mm}Naive & 0 & 0 & 6 & 27 \\
\hspace{3mm}Tree & 0 & 0 & 27 & 6 \\
\hspace{3mm}SPatterNet & 9 & 24 & 0 & 0 \\
\hspace{3mm}SPatterNet (JIT) & 24 & 9 & 0 & 0 \\
\bottomrule
\end{tabular}
\caption{For each of the $33$ games, and for both uniform action sampling as well as action sampling from a trained policy, we rank the four feature evaluation approaches based on their performance. This table lists how often each rank was obtained by every feature evaluation approach, where a rank of 1 means best performance and a rank of 4 means worst performance in a given game.}
\label{Table:TrainedFeaturesRankCounts}
\end{table}

\subsubsection{Discussion}

Both \reffigure{Fig:BenchmarkTrainedFeatureSets} and \reftable{Table:TrainedFeaturesRankCounts} show continuations of the trends discussed in \refsubsubsection{Subsubsec:DiscussionAtomic}. The two variants of SPatterNets outperform the others in terms of aggregate results, as well as dominating the top $2$ ranks across all $33$ games. There also appears to be a more pronounced advantage for the JIT variant over the standard variant of SPatterNet in the trained feature sets. The Tree approach appears to more convincingly outperform the Naive approach (especially in \reftable{Table:TrainedFeaturesRankCounts}) than it does in the atomic feature sets. Its generalisation-based optimisation is significantly more effective in the trained feature sets, because every new feature that is added during the training process is a composite of---and therefore generalised by---at least two other feature instantiations.

We do not observe any major differences in these trends between the results gathered from uniformly random playouts, and playouts where actions are sampled from trained policies. On average playouts appear to run more quickly when actions are sampled from the trained policy, but there are also games where they are slower. These differences are observed regardless of which implementation is used for evaluating active features, and are due to overall more ``intelligent'' action selection---which in many games causes playouts to have a shorter duration in terms of the average number of actions per playout.

\subsection{Playing Strength of MCTS with Trained Policies}

The purpose of the final experiment for this paper is to evaluate the impact that the improved speed of SPatterNet over the baselines has in terms of the playing strength of MCTS agents that are guided by trained policies. Taking the same trained feature sets as used in \refsubsection{Subsec:TrainedFeaturesExperiment}, a separate MCTS agent is constructed for each of the four feature evaluation approaches under consideration (Naive, Tree, SPatterNet, and SPatterNet (JIT)). Except for using different approaches to evaluate features, these four agents are identical. The rest of the setup of the biased MCTS agents is as described in \refappendix{Appendix:DetailsTraining} for the agents used during self-play training, with the only exception being that, in evaluation games---as opposed to training games---the agents pick actions that maximise visit counts, rather than proportionally to visit counts. For each of the six possible pairings of four such agents (excluding mirror matchups, excluding pairings that are permutations of other pairings), 100 evaluation games were played between the two agents, in each of the 30 two-player games also used in previous experiments (excluding three games for more than two players). The three games for more than two players are excluded from this experiment, because they require a significantly larger amount of computation resource for appropriate statistical analyses. This is due to the large number of possible permutations of agent assignments to player roles in games with more than two roles.

\subsubsection{Results}

\begin{figure}[t]
\centering
\includegraphics[width=\linewidth]{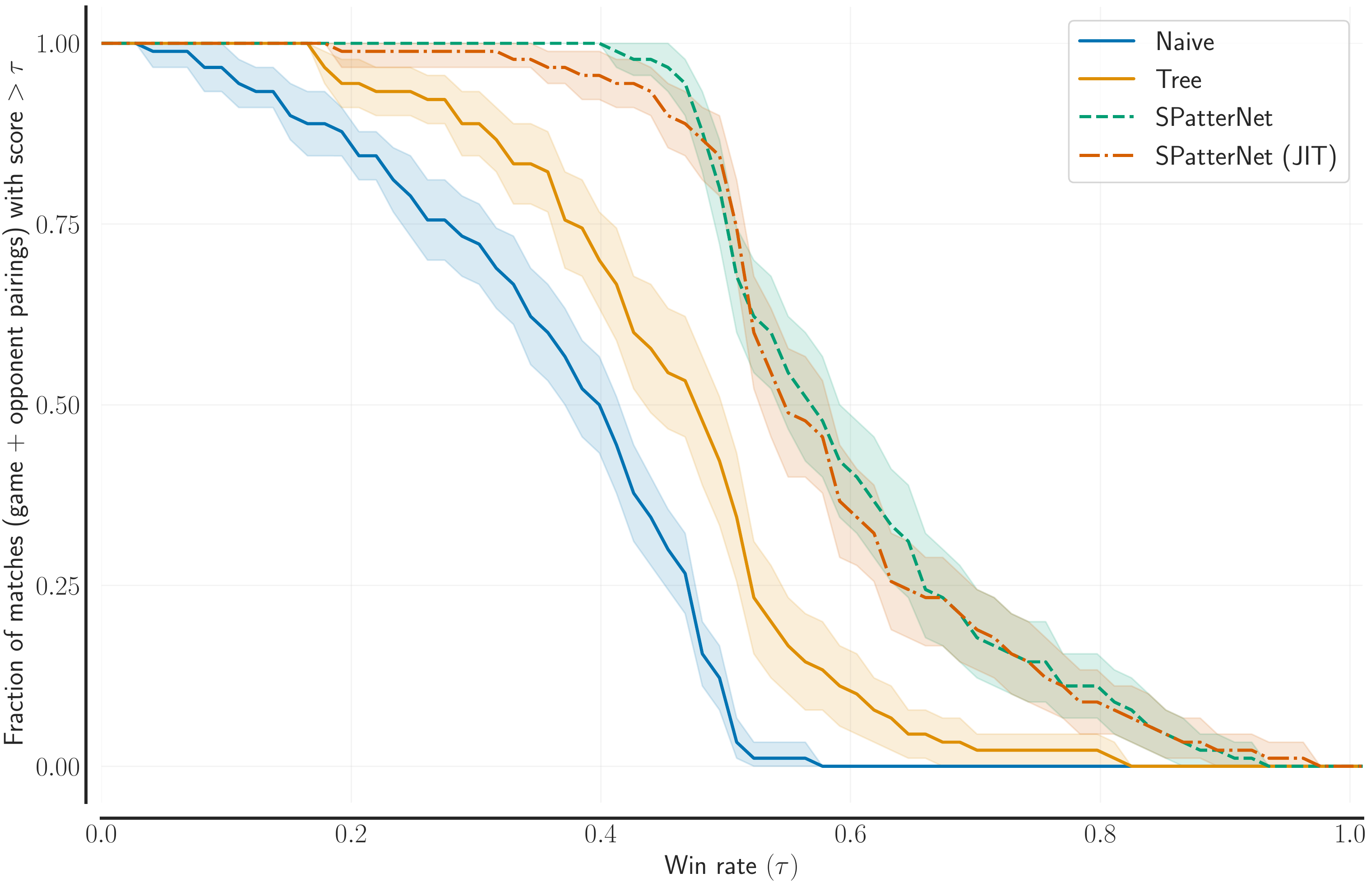}
\caption{Performance profiles \cite{Agarwal_2021_Deep} for MCTS agents biased by trained features, with different approaches for evaluating features.}
\label{Fig:PerformanceProfiles_FeatureSetImpls}
\end{figure}

\reffigure{Fig:PerformanceProfiles_FeatureSetImpls} depicts performance profiles \cite{Agarwal_2021_Deep} for the MCTS agents using different approaches for their feature evaluations. A datapoint at $(x, y)$ in the plot means that an agent achieved a winrate (averaged over the four possible opponents, including mirror matchups) greater than or equal to $x$ on a fraction equal to $y$ out of the thirty games. The shaded intervals depict 95\% bootstrap confidence intervals from 2000 replicates, indicating variability in results over the four different opponents for every agent.

\begin{figure}[t]
\centering
\includegraphics[width=\linewidth]{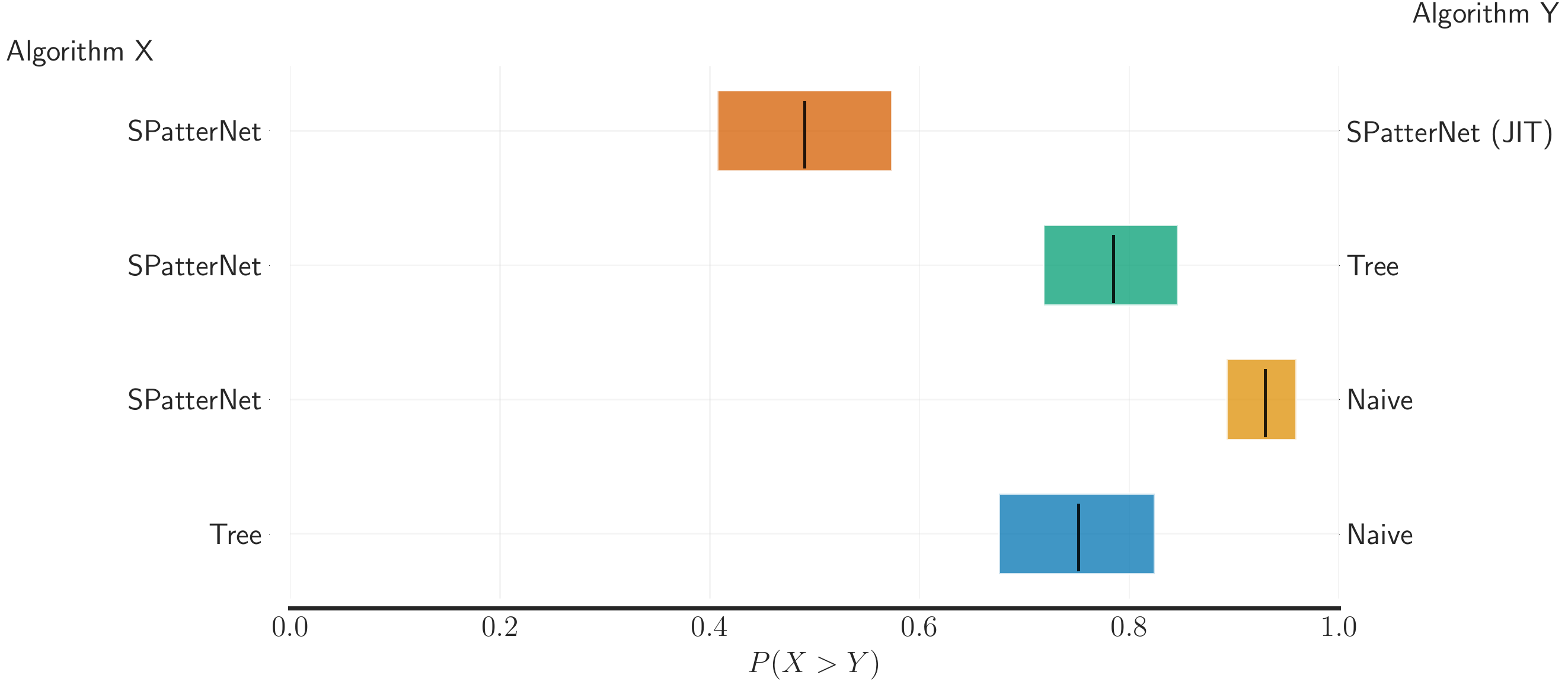}
\caption{Pairwise probabilities of improvement \cite{Agarwal_2021_Deep} for algorithms on the left-hand side to outperform algorithms on the right-hand side.}
\label{Fig:ProbabilitiesImprovementFeatureSetImpls}
\end{figure}

\reffigure{Fig:ProbabilitiesImprovementFeatureSetImpls} depicts probabilities of improvement for four different pairs of agents. For each of the pairs (rows), it provides the probability that the approach on the left-hand side improves upon the approach on the right-hand side on any given individual game from the set of thirty games, using the implementation from the \textit{rliable} framework \cite{Agarwal_2021_Deep}, which uses the Mann-Whitney U-statistic \cite{Mann_1947_MannWhitney}. Confidence intervals are based on 2000 bootstrap replications.

\reftable{Table:WinPercentageStats} lists 95\% confidence intervals for the median, interquartile mean (IQM), and mean win rates of the four evaluated algorithms, each against the three other algorithms over the 30 different games. Note that the point estimates for the means and medians correspond to the areas under the curves and the points of intersection with the horizontal $y = 0.5$ line, respectively, of the performance profiles in \reffigure{Fig:PerformanceProfiles_FeatureSetImpls}.

\begin{table}[t]
\centering
\begin{tabular}{@{}lrrr@{}}
\toprule
& \multicolumn{3}{c}{Win rate against all other algorithms} \\
\cmidrule(lr){2-4}
Algorithm & Median & IQM & Mean \\
\midrule
Naive & [0.35, 0.41] & [0.37, 0.41] & [0.35, 0.38] \\
Tree & [0.45, 0.49] & [0.44, 0.48] & [0.43, 0.48] \\
SPatterNet & [0.56, 0.61] & [0.55, 0.60] & [0.58, 0.62] \\
SPatterNet (JIT) & [0.54, 0.59] & [0.54, 0.58] & [0.57, 0.60] \\
\bottomrule
\end{tabular}
\caption{95\% bootstrap confidence intervals for median, interquartile mean (IQM), and mean win rates (stratified across 30 games) of MCTS agents biased by trained features, with different approaches for evaluating features.}
\label{Table:WinPercentageStats}
\end{table}

\subsubsection{Discussion}

The performance profiles of \reffigure{Fig:PerformanceProfiles_FeatureSetImpls} show both variants of SPatterNet (the regular and the JIT variant) \textit{stochastically dominating} \cite{Levy_1992_Stochastic,Dror_2019_DeepDominance} both of the baselines (with their plots not being below either of the baselines for any point $\tau$), which is a clear sign of them providing consistently higher levels of performance. \reffigure{Fig:ProbabilitiesImprovementFeatureSetImpls} similarly shows the SPatterNet approach outperforming both of the baselines by statistically significant margins (with the Tree baseline itself also significantly outperforming the Naive baseline). While the standard variant of the SpatterNet approach appears to possibly slightly outperform the JIT variant in the performance profiles, this is not by a significant margin, and also cannot be concluded from the pairwise comparison in \reffigure{Fig:ProbabilitiesImprovementFeatureSetImpls}. In the latter figure, with the upper thresholds of the confidence intervals exceeding $0.75$ for all three of the bottom rows, these results are also statistically meaningful following the threshold recommended by \citet{Bouthillier_2021_Accounting}. The results in \reftable{Table:WinPercentageStats} lead to similar conclusions. For both SPatterNet and SPatterNet (JIT), even the lower bounds of all three aggregate metrics exceed the corresponding upper bounds for the two baseline algorithms, demonstrating statistically significant levels of improvement.

\section{Applications Beyond Games} \label{Sec:Discussion}

The focus in this paper was on efficiently evaluating patterns of spatial state-action features for game playing as an application domain. However, the SPatterNet approach developed for this purpose applies to a highly general and abstract formulation of the problem, where we simply aim to optimise the order in which propositions of logical formulas in disjunctive normal form are evaluated. This problem formulation, and hence also the SPatterNet approach, may be applicable to other types of features than spatial ones, or other types of problems than game playing.

For example, potential other applications may include classification domains with costly features, such as medical, fraud detection, or computer security domains, where feature values can be costly to obtain \cite{Mannino_1999_Optimizing,Trapeznikov_2013_Supervised,Janisch_2020_Classification}. In such domains, sets of learned decision rules may be thought of as logical formulas in disjunctive normal form, where every conjunction corresponds to a single rule, and a disjunction is formed by a set of different rules that have the same output class \cite{Mannino_1999_Optimizing}. 

Aside from sets of decision rules, models such as Decision Trees or Random Forests also lend themselves naturally towards being translated into disjunctive normal form. Such models have also been used in settings with costly features \cite{Nan_2015_FeatureBudgeted,Nan_2016_Pruning}. In a Decision Tree, every path from the root to a leaf may be viewed as a conjunction, and sets of paths leading to different leaves with the same output class can be viewed as disjunctions of conjunctions. While the optimal order in which to evaluate propositions would likely closely correspond to the order in which they appear in a Decision Tree, it may deviate if there are implication relationships between propositions, and propositions that appear in multiple subtrees below a shared ancestor. For Random Forests with, for example, a majority voting rule, an efficient ordering for propositions may also differ from the order dictated by any single individual tree of the forest.

\section{Conclusion} \label{Sec:Conclusion}

In previous work, we proposed an initial design and formalisation of spatial features for general games \cite{Browne_2019_Strategic} in the Ludii general game system, and demonstrated that they can be used to train simple policies that effectively improve playing strength in a variety of different games from self-play using few resources \cite{Soemers_2019_Biasing,Soemers_2020_Manipulating}. However, whether or not such policies substantially improve playing strength also depends greatly on the amount of computational overhead associated with feature evaluations \cite{Soemers_2019_Biasing}. This paper has two core contributions. Firstly, we extend the design and formalisation of the spatial features from previous work, and provide significantly more detail on several implementation and design decisions. Combined with the publicly available source code, this should aid replicability. Secondly, we propose a new approach to significantly improve the efficiency of evaluating active features, reducing the computational overhead when using features to, for instance, guide a tree search. An empirical evaluation demonstrates the improvements in efficiency, and also shows that this significantly improves the playing strength of agents using the features to guide their search.

Within the domain of AI for games, one idea for future research is transfer learning of policies based on spatial state-action features between games. The features have been designed to facilitate this---for instance by formalising the walks used to define relative positions in such a way that their semantics remain similar when transferred to different board geometries---but this has yet to be evaluated. Another idea for future research would be a further extension of the feature formalisation, to include additional types of elements that may be important for categories of games that are not yet well-supported. For example, games with hidden information would likely benefit from an extended format where features can specify that certain positions should be hidden, and games where pieces can stack up on top of each other would benefit from adjustments to take such a third ``spatial dimension'' into account. These are some of the same categories of games that \citet{Soemers_2022_DeepLearning} also identified as having received little attention in research based on deep learning approaches.

While we focused on applications for game AI in this paper, we remark that the proposed SPatterNet approach for organising propositions and determining the order in which they should be evaluated is not necessarily restricted to game AI. A similar approach may more generally be applicable to any domain where sets of disjunctions of conjunctions need to be evaluated efficiently, in particular when there are generalisation relationships or implications (based on domain knowledge) between them that the SPatterNet approach can leverage.

\section*{Acknowledgements}
This research is funded by the European Research Council as part of the Digital Ludeme Project (ERC Consolidator Grant \#771292) led by Cameron Browne at Maastricht University's Department of Advanced Computing Sciences. We wish to thank Walter Crist, C{\'e}dric Piette, Chiara Sironi, and Mark Winands for helpful pointers to related work. This work was carried out on the Dutch national e-infrastructure with the support of SURF Cooperative (grant no. EINF-1133). This work used the Dutch national e-infrastructure with the support of the SURF cooperative using grant no. EINF-4028. This publication is part of the project ``Evaluation of Trained AIs for General Game Playing'' (with project number EINF-4028) of the research programme Computing Time on National Computer Facilities which is (partly) financed by the Dutch Research Council (NWO).

\appendix

\section{Games Used for Experiments} \label{Appendix:Games}
\begin{figure}
\centering
\begin{subfigure}{.19\textwidth}
  \centering
  \includegraphics[width=\linewidth]{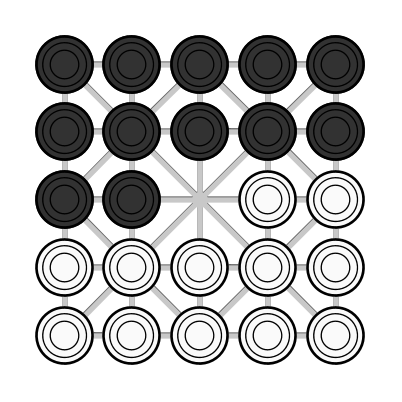}
\end{subfigure}
\begin{subfigure}{.19\textwidth}
  \centering
  \includegraphics[width=\linewidth]{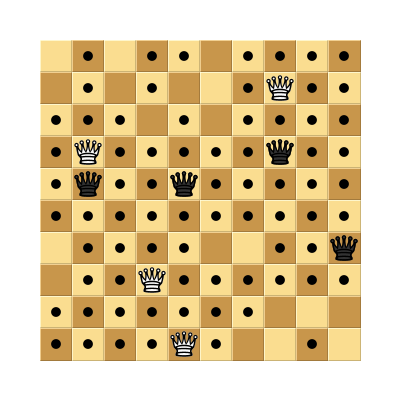}
\end{subfigure}
\begin{subfigure}{.19\linewidth}
  \centering
  \includegraphics[width=\linewidth]{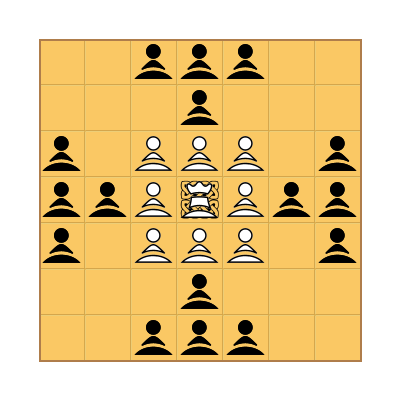}
\end{subfigure}
\begin{subfigure}{.19\textwidth}
  \centering
  \includegraphics[width=\linewidth]{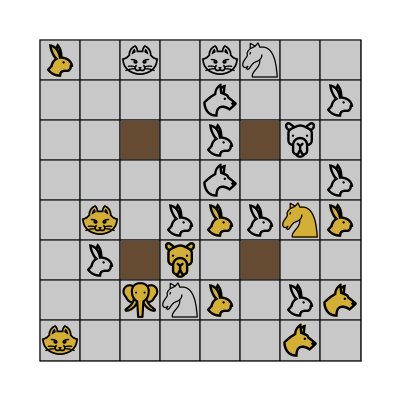}
\end{subfigure}
\begin{subfigure}{.19\textwidth}
  \centering
  \includegraphics[width=\linewidth]{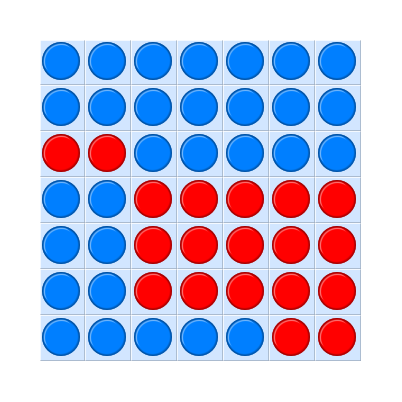}
\end{subfigure}
\begin{subfigure}{.19\textwidth}
  \centering
  \includegraphics[width=\linewidth]{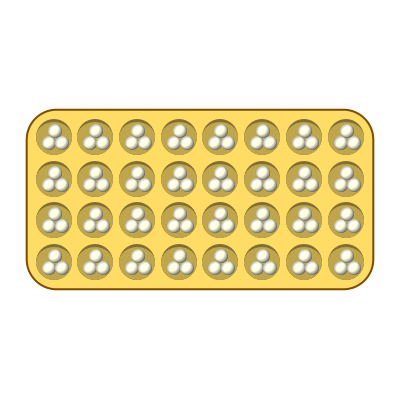}
\end{subfigure}
\begin{subfigure}{.19\textwidth}
  \centering
  \includegraphics[width=\linewidth]{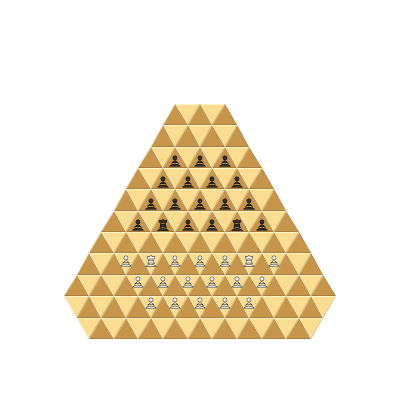}
\end{subfigure}
\begin{subfigure}{.19\textwidth}
  \centering
  \includegraphics[width=\linewidth]{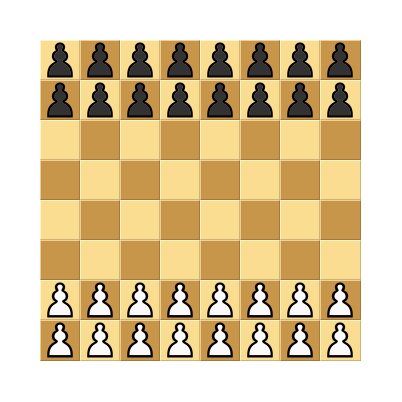}
\end{subfigure}
\begin{subfigure}{.19\textwidth}
  \centering
  \includegraphics[width=\linewidth]{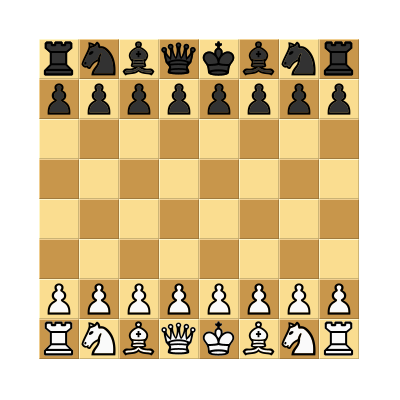}
\end{subfigure}
\begin{subfigure}{.19\textwidth}
  \centering
  \includegraphics[width=\linewidth]{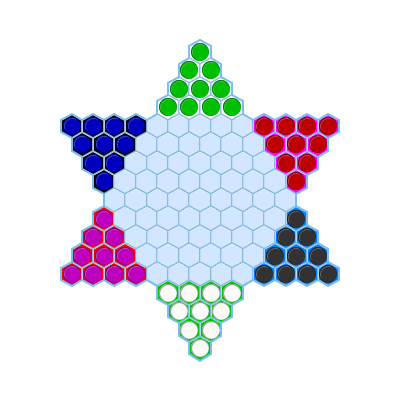}
\end{subfigure}
\begin{subfigure}{.19\textwidth}
  \centering
  \includegraphics[width=\linewidth]{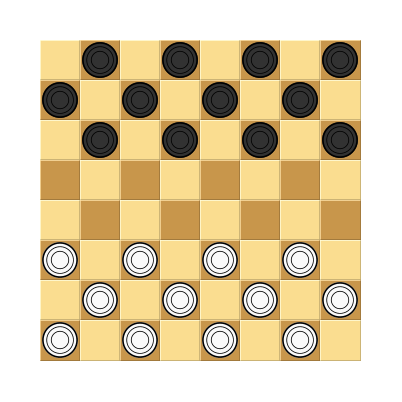}
\end{subfigure}
\begin{subfigure}{.19\textwidth}
  \centering
  \includegraphics[width=\linewidth]{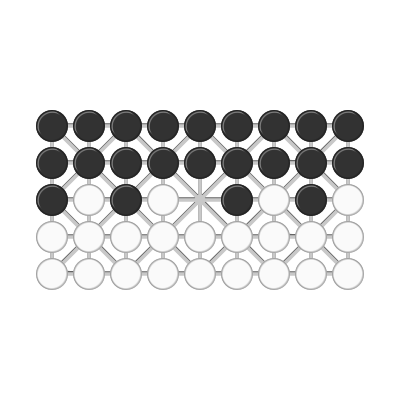}
\end{subfigure}
\begin{subfigure}{.19\textwidth}
  \centering
  \includegraphics[width=\linewidth]{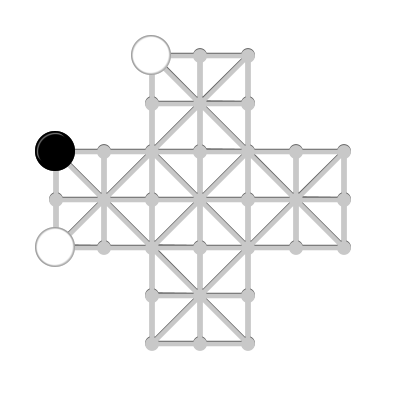}
\end{subfigure}
\begin{subfigure}{.19\textwidth}
  \centering
  \includegraphics[width=\linewidth]{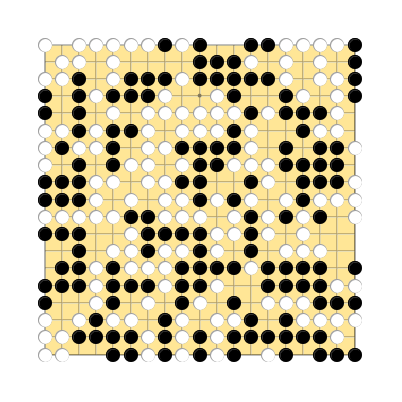}
\end{subfigure}
\begin{subfigure}{.19\textwidth}
  \centering
  \includegraphics[width=\linewidth]{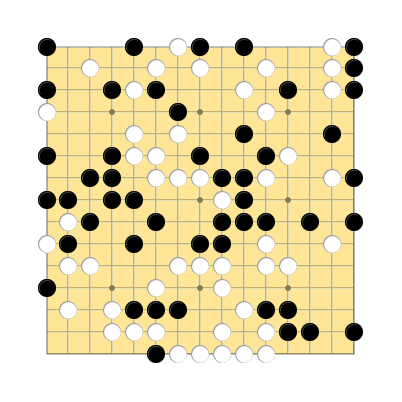}
\end{subfigure}
\begin{subfigure}{.19\textwidth}
  \centering
  \includegraphics[width=\linewidth]{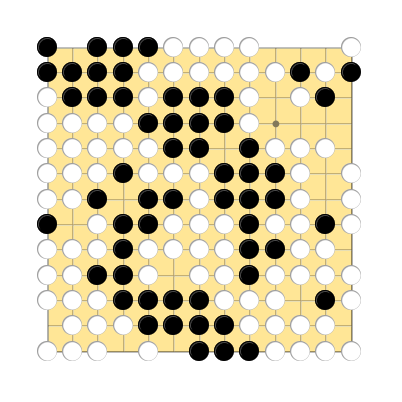}
\end{subfigure}
\begin{subfigure}{.19\textwidth}
  \centering
  \includegraphics[width=\linewidth]{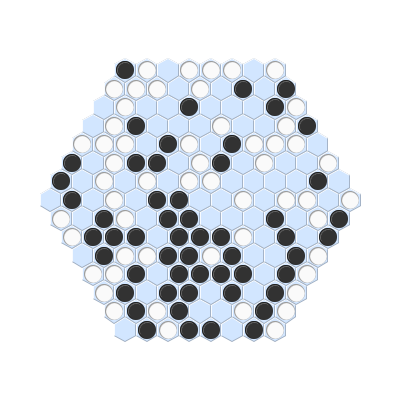}
\end{subfigure}
\begin{subfigure}{.19\textwidth}
  \centering
  \includegraphics[width=\linewidth]{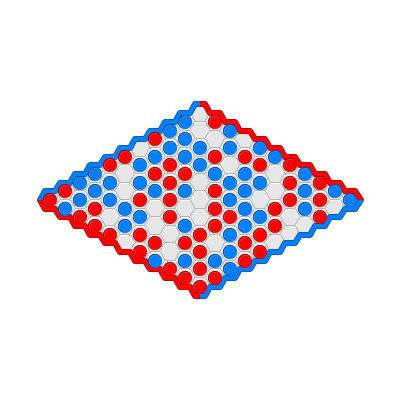}
\end{subfigure}
\begin{subfigure}{.19\textwidth}
  \centering
  \includegraphics[width=\linewidth]{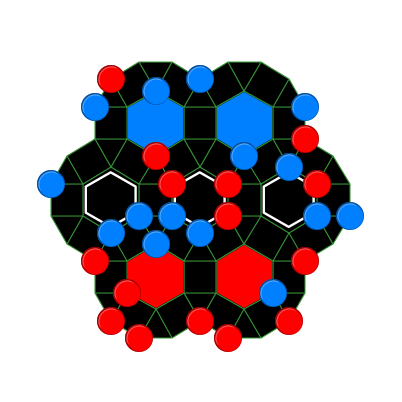}
\end{subfigure}
\begin{subfigure}{.19\textwidth}
  \centering
  \includegraphics[width=\linewidth]{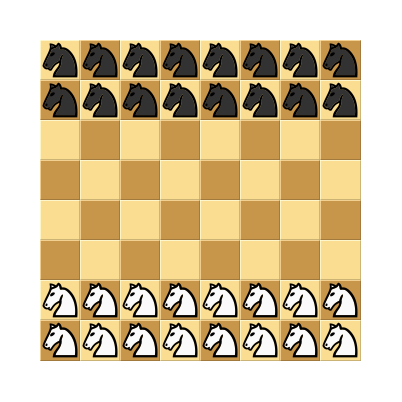}
\end{subfigure}
\begin{subfigure}{.19\textwidth}
  \centering
  \includegraphics[width=\linewidth]{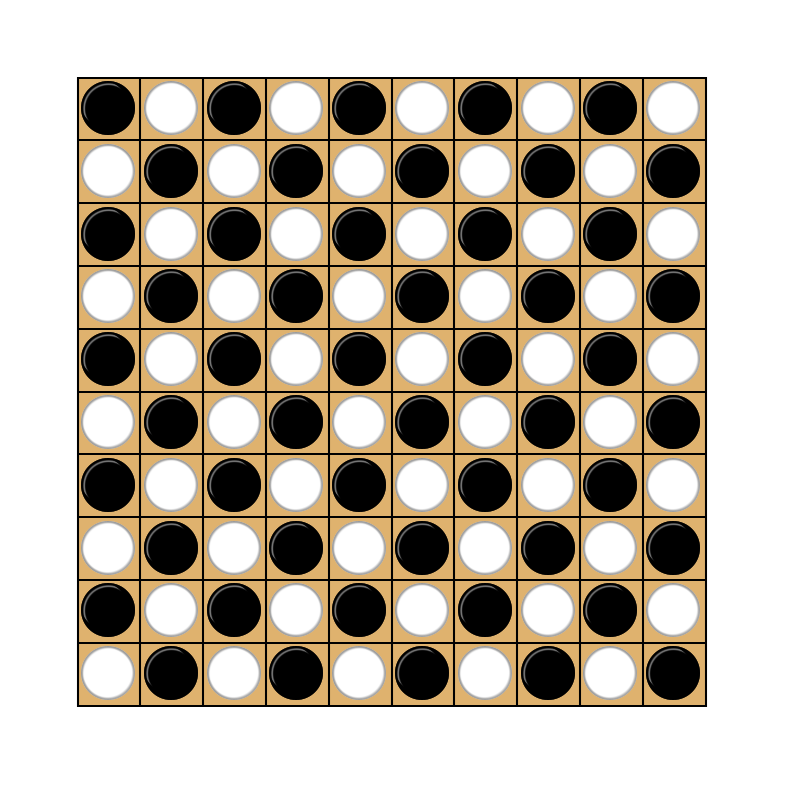}
\end{subfigure}
\begin{subfigure}{.19\textwidth}
  \centering
  \includegraphics[width=\linewidth]{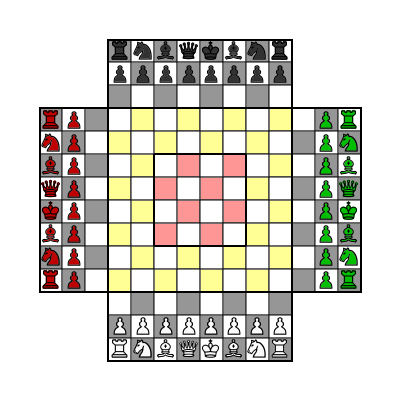}
\end{subfigure}
\begin{subfigure}{.19\textwidth}
  \centering
  \includegraphics[width=\linewidth]{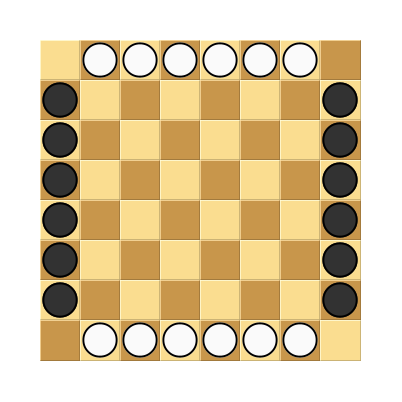}
\end{subfigure}
\begin{subfigure}{.19\textwidth}
  \centering
  \includegraphics[width=\linewidth]{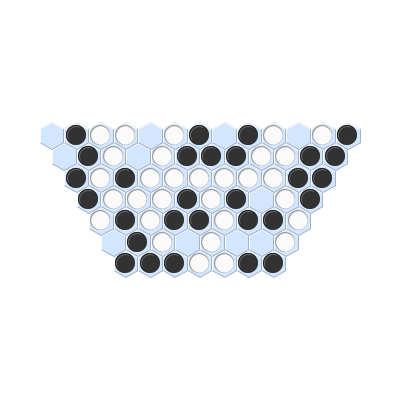}
\end{subfigure}
\begin{subfigure}{.19\textwidth}
  \centering
  \includegraphics[width=\linewidth]{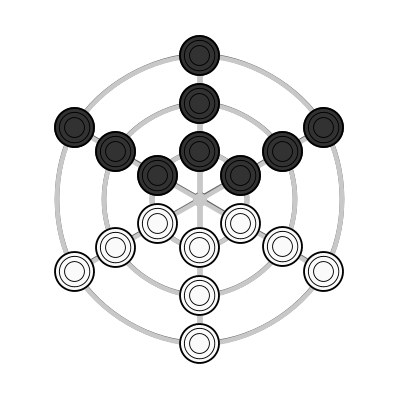}
\end{subfigure}
\begin{subfigure}{.19\textwidth}
  \centering
  \includegraphics[width=\linewidth]{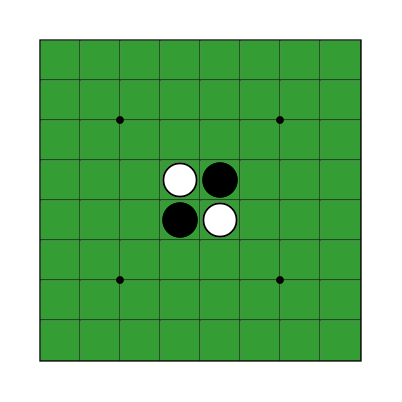}
\end{subfigure}
\begin{subfigure}{.19\textwidth}
  \centering
  \includegraphics[width=\linewidth]{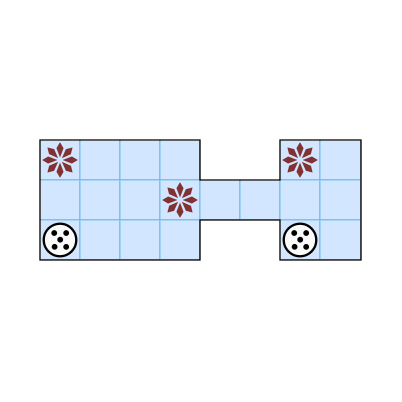}
\end{subfigure}
\begin{subfigure}{.19\textwidth}
  \centering
  \includegraphics[width=\linewidth]{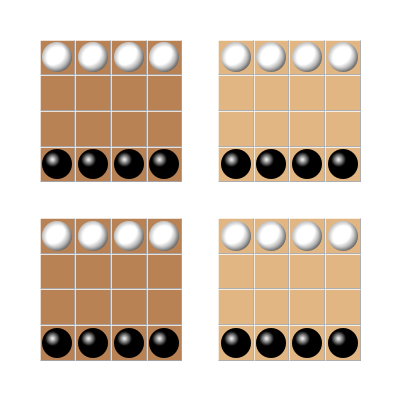}
\end{subfigure}
\begin{subfigure}{.19\textwidth}
  \centering
  \includegraphics[width=\linewidth]{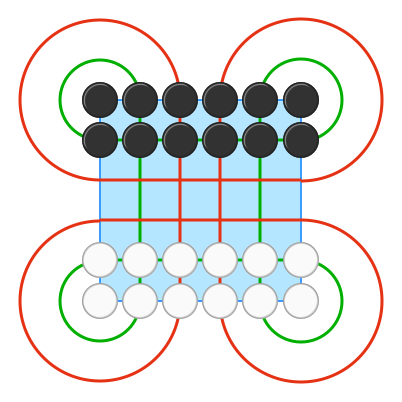}
\end{subfigure}
\begin{subfigure}{.19\textwidth}
  \centering
  \includegraphics[width=\linewidth]{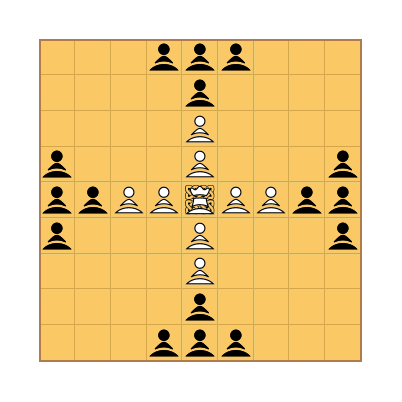}
\end{subfigure}
\begin{subfigure}{.19\textwidth}
  \centering
  \includegraphics[width=\linewidth]{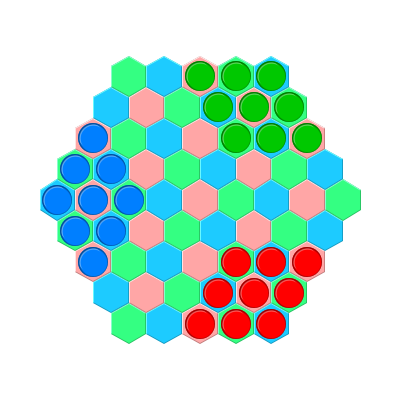}
\end{subfigure}
\begin{subfigure}{.19\textwidth}
  \centering
  \includegraphics[width=\linewidth]{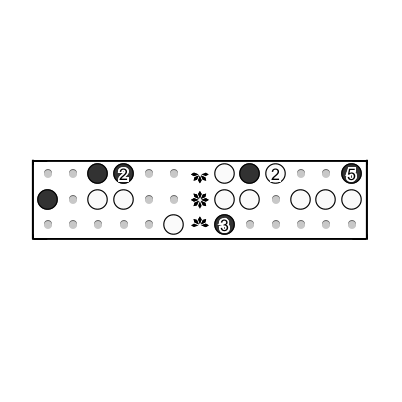}
\end{subfigure}
\begin{subfigure}{.19\textwidth}
  \centering
  \includegraphics[width=\linewidth]{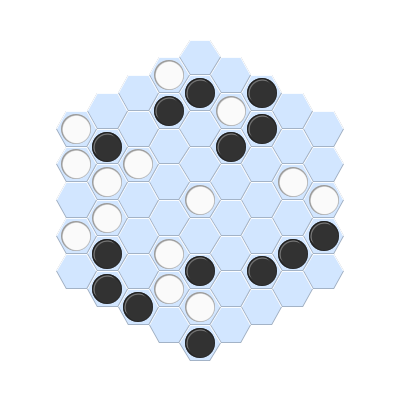}
\end{subfigure}
\caption{Screenshots of games used in experiments, generated by Ludii.}
\label{Fig:GameThumbnails}
\end{figure}

This appendix provides additional detail on all the games (a total of $33$ distinct games) used for experiments described throughout this paper. \reftable{Table:GameProperties} lists the names of all the selected games, as well as several properties. \reffigure{Fig:GameThumbnails} depicts thumbnails of all the games, which provides an impression of the variety in board shapes and connectivity structures included in the experiments. This selection of games includes:
\begin{itemize}
    \item Several games that have been commonly used as benchmarks for AI research, such as Arimaa \cite{Syed_2003_Arimaa}, Chess \cite{Campbell_2002_DeepBlue}, Go \cite{Muller_2002_Go,Silver_2016_AlphaGo}, and Hex \cite{Cazenave_2020_Polygames}.
    \item A mix of games where actions primarily involve movement from one site to another, and games where actions primarily involve placing new pieces on sites (see \reftable{Table:GameProperties})---these are significant differences for the spatial features discussed in this paper.
    \item Several games with a significant degree of asymmetry (in initial setup of pieces, piece types, victory conditions, etc.) between different players; ArdRi, Bizingo, Fox and Geese, and Tablut.
    \item Two stochastic games: Royal Game of Ur, and XII Scripta.
    \item Three games with more than two players: Chinese Checkers, Level Chess, and Triad.
    \item A high degree of variety in board shapes and connectivity structures between sites; see \reffigure{Fig:GameThumbnails}.
\end{itemize}

For all games, we used the default options (e.g., default board sizes) as implemented in the Ludii general game system. Detailed information on each of these games can be found on \url{https://ludii.games/library.php}.

\begin{table}
\centering
\begin{tabular}{@{}llccr@{}}
\toprule
Game & Primary Action Types & Asymmetric & Stochastic & $K$ \\
\midrule
Alquerque & Movement & $\times$ & $\times$ & $2$ \\
Amazons & Movement \& Placement & $\times$ & $\times$ & $2$ \\
ArdRi & Movement & \checkmark & $\times$ & $2$ \\
Arimaa & Movement & $\times$ & $\times$ & $2$ \\
Ataxx & Movement & $\times$ & $\times$ & $2$ \\
Bao Ki Arabu (Zanzibar 1) & Placement & $\times$ & $\times$ & $2$ \\
Bizingo & Movement & \checkmark & $\times$ & $2$ \\
Breakthrough & Movement & $\times$ & $\times$ & $2$ \\
Chess & Movement & $\times$ & $\times$ & $2$ \\
Chinese Checkers & Movement & $\times$ & $\times$ & $6$ \\
English Draughts & Movement & $\times$ & $\times$ & $2$ \\
Fanorona & Movement & $\times$ & $\times$ & $2$ \\
Fox and Geese & Movement & \checkmark & $\times$ & $2$ \\
Go & Placement & $\times$ & $\times$ & $2$ \\
Gomoku & Placement & $\times$ & $\times$ & $2$ \\
Gonnect & Placement & $\times$ & $\times$ & $2$ \\
Havannah & Placement & $\times$ & $\times$ & $2$ \\
Hex & Placement & $\times$ & $\times$ & $2$ \\
Kensington & Movement & $\times$ & $\times$ & $2$ \\
Knightthrough & Placement & $\times$ & $\times$ & $2$ \\
Konane & Movement & $\times$ & $\times$ & $2$ \\
Level Chess & Movement & $\times$ & $\times$ & $4$ \\
Lines of Action & Movement & $\times$ & $\times$ & $2$ \\
Pentalath & Placement & $\times$ & $\times$ & $2$ \\
Pretwa & Movement & $\times$ & $\times$ & $2$ \\
Reversi & Placement & $\times$ & $\times$ & $2$ \\
Royal Game of Ur & Movement & $\times$ & \checkmark & $2$ \\
Shobu & Movement & $\times$ & $\times$ & $2$ \\
Surakarta & Movement & $\times$ & $\times$ & $2$ \\
Tablut & Movement & \checkmark & $\times$ & $2$ \\
Triad & Movement \& Placement & $\times$ & $\times$ & $3$ \\
XII Scripta & Movement & $\times$ & \checkmark & $2$ \\
Yavalath & Placement & $\times$ & $\times$ & $2$ \\
\bottomrule
\end{tabular}
\caption{Various properties of the games used in experiments. In the second column, ``Movement'' is listed for games that involve actions with both source and destination positions (steps, slides, hops, etc.), and ``Placement'' is listed for games that involve actions with only a destination position (e.g., placing stones). The final column ($K$) lists the number of players per game.}
\label{Table:GameProperties}
\end{table}

\section{Details on Self-Play Training Setup} \label{Appendix:DetailsTraining}

Like the speed evaluations described in \refsection{Sec:Experiments}, every training run (one per game) ran as a separate process on two cores of a 2.6GHz Intel Xeon E5-2690 v3 CPU, with 4096MB of RAM allocated to the Java Virtual Machine (JVM), using Java version 8u261. Every training process ran for up to 200 episodes, or up to 24 hours of wall time. The training process is similar as described in our previous work \cite{Soemers_2019_Biasing,Soemers_2020_Manipulating}, which in turn is largely inspired by the AlphaGo Zero and Expert Iteration training processes \cite{Anthony_2017_ExIt,Silver_2017_AlphaGoZero}. Additional details are provided below.

Each training process starts by generating the Atomic-2-4 feature sets for each of $K$ players in a given $K$-player game. Self-play experience is generated by $K$ copies of MCTS agents, which use:
\begin{itemize}
    \item The same PUCT selection strategy as AlphaGo Zero \cite{Silver_2017_AlphaGoZero}, i.e. traversing the search tree by selecting actions $a^*$ according to 
    
    \begin{equation*}
        a^* = \argmax_a \hat{Q}(s, a) + C_{puct} \pi(s, a) \frac{\sqrt{\sum_{a'} N(s, a')}}{1 + N(s, a)},
    \end{equation*} 
    
    where $\hat{Q}(s, a)$ denotes the current estimated value (average backpropagated score) for action $a$ in state $s$, $C_{puct}$ denotes an exploration constant (set to $C_{puct} = 2.5$), $\pi(s, a)$ denotes the probability assigned to $a$ in $s$ by the trained policy, and $N(s, a)$ denotes the number of previous visits to the node that selecting $a$ in $s$ leads to in the search tree.
    
    \item An $\epsilon$-greedy playout strategy where actions during playouts are sampled uniformly at random with probability $\epsilon = 0.5$, and sampled according to the trained policy $\pi$ with probability $1 - \epsilon = 0.5$.
    
    \item A straightforward expansion strategy that consists of expanding the tree by exactly one node per MCTS iteration.
    
    \item Values in the range $[-1, 1]$ for backpropagation; losses correspond to a value of $-1$, wins to a value of $1$, draws to a value of $0$, second place in a $6$-player game to a value of $1 - \left( \left( 2 - 1 \right) \times \frac{2}{6 - 1} \right) = 0.6$, etc. Note that this range of values is twice as big as the $[0, 1]$ range commonly used by other programs. To compensate for this, we set the $C_{puct} = 2.5$ exploration constant to a value that is approximately twice as large as the value used by programs such as AlphaZero ($C_{puct}$ slowly shrinking from $1.25$) \cite{Silver_2018_AlphaZero} or ELF OpenGo ($C_{puct} = 1.5$) \cite{Tian_2019_ELF}.
    
    \item Value estimates $\hat{Q}(s, a)$ equal to the parent node's value estimate for nodes that have no visits (i.e., $N(s, a) = 0$).
    
    \item Tree reuse, meaning that relevant parts of the search tree from previous search processes (in previous turns) are preserved for subsequent searches by the same agent \cite{Pepels_2014_PacMan,Soemers_2016_GVGAI,Santos_2017_Hearthstone}.
    
    \item In stochastic games, an open-loop approach where nodes (other than the root node) do not store copies of game states, but only represent and collect statistics for the trajectories of actions leading up to them from the root node \cite{Perez_2015_OpenLoop}. States are re-generated by applying those sequences of actions to copies of the root state as required in different MCTS iterations.
\end{itemize}
Every move, the MCTS agent corresponding to the player to move uses $1$ second of thinking time, after which it selects a move proportionally to the distribution of visit counts among the children of the root node. 

After every full game of self-play, we store collected samples of experience in replay buffers, where separate replay buffers per player each contain samples corresponding to game states in which the corresponding player is the player to move. We use Prioritized Experience Replay (PER) \cite{Schaul_2016_PER,Soemers_2020_Manipulating}, with hyperparameters $\alpha = \beta = 0.5$. Each replay buffer has a maximum capacity of $2500$ tuples of experience.

After every action in self-play, we run, for every player, one step of gradient descent to minimise the cross-entropy loss function (given for a single state $s$, representing a tuple of experience)
\begin{equation*}
    \mathcal{L}(s) = \left( -\transpose{\boldsymbol{\pi}_{mcts}(s)} \log \boldsymbol{\pi}(s) \right),
\end{equation*}
where $\boldsymbol{\pi}_{mcts}(s)$ denotes the probability distribution over actions legal in $s$ proportional to the visit counts resulting from an MCTS search in $s$, and $\boldsymbol{\pi}(s)$ similarly denotes the distribution over actions in the state $s$ for the trained policy. We use a batch size of $N = 30$. Tuples of experience are additionally weighted based on the durations of the episodes from which they originate \cite{Soemers_2020_Manipulating}. Weighted importance sampling is used to correct for PER (partially) and the weighting according to episode durations. Gradient descent steps are taken using a centered variant of RMSProp \cite{Graves_2013_Generating}, with a base learning rate of $0.005$, a momentum of $0.9$, a discounting factor of $0.9$, and a constant of $10^{-8}$ added to the denominator for stability. Additionally, we regularise by applying weight decay with $\lambda = 10^{-6}$ after every step.

After every game of self-play, we grow every player's feature set by up to two features, which are constructed by re-combining two existing feature instances. We aim to construct new features such that the absolute correlation between its activity and the activity of either of its constituents is minimised, while the absolute correlation between its activity and the cross-entropy loss function is maximised \cite{Soemers_2019_Biasing}. Correlations for potential candidates are estimated from batches of size $N = 30$ sampled uniformly from the replay buffer. Every time, we generate at most one new proactive feature and one new reactive feature. Sometimes, either one or both of these may fail if all candidates turn out to be equivalent to a feature that already exists.

\bibliography{dlp-biblio-1}

\end{document}